\documentclass[journal,twocolumn]{IEEEtran}
\usepackage{graphicx}
\usepackage[hidelinks]{hyperref}
\usepackage{multirow}
\usepackage{blindtext}

\usepackage[english]{babel}
\usepackage[justification=centering]{caption}
\usepackage{amssymb}
\usepackage{pifont}
\usepackage[breakable, skins]{tcolorbox}
\usepackage{booktabs}

\definecolor{lb}{RGB}{44, 139, 183}

\newcommand\bcl[2]{\textcolor{#1}{{\fontseries{b}\selectfont #2}}}

\definecolor{lime}{HTML}{A6CE39}
\DeclareRobustCommand{\orcidicon}{%
	\begin{tikzpicture}
	\draw[lime, fill=lime] (0,0) 
	circle [radius=0.16] 
	node[white] {{\fontfamily{qag}\selectfont \tiny ID}};
	\draw[white, fill=white] (-0.0625,0.095) 
	circle [radius=0.007];
	\end{tikzpicture}
	\hspace{-2mm}
}

\foreach \x in {A, ..., Z}{%
	\expandafter\xdef\csname orcid\x\endcsname{\noexpand\href{https://orcid.org/\csname orcidauthor\x\endcsname}{\noexpand\orcidicon}}
}

\ifCLASSINFOpdf
\else
\fi

\begin{document}
\title{A Survey on Approximate Edge AI for Energy Efficient Autonomous Driving Services}

\author{Dewant~Katare\orcidA{},~\IEEEmembership{Student Member,~IEEE,}
       Diego~Perino\orcidB{}, 
       Jari~Nurmi\orcidC{},~\IEEEmembership{Senior Member,~IEEE,}\\
       Martijn~Warnier\orcidD{}, 
       Marijn~Janssen\orcidE{}, 
       and~Aaron~Yi~Ding\orcidF{},~\IEEEmembership{Member,~IEEE}
\thanks{Dewant Katare, Martijn Warnier, Marijn Janssen, Aaron Yi Ding are with the Delft University of Technology, The Netherlands.}
\thanks{Diego Perino is with Meta, USA.}
\thanks{Jari Nurmi is with Tampere University, Finland.}%
\thanks{Corresponding e-mails: d.katare@tudelft.nl, aaron.ding@tudelft.nl}}
\maketitle

\begin{abstract}


Autonomous driving services rely heavily on sensors such as cameras, LiDAR, radar, and communication modules. A common practice of processing the sensed data is using a high-performance computing unit placed inside the vehicle, which deploys AI models and algorithms to act as the brain or administrator of the vehicle. The vehicular data generated from average hours of driving can be up to 20 Terabytes depending on the data rate and specification of the sensors. Given the scale and fast growth of services for autonomous driving, it is essential to improve the overall energy and environmental efficiency, especially in the trend towards vehicular electrification (e.g., battery-powered). 
Although the areas have seen significant advancements in sensor technologies, wireless communications, computing and AI/ML algorithms, the challenge still exists in how to apply and integrate those technology innovations to achieve energy efficiency.
This survey reviews and compares the connected vehicular applications, vehicular communications, approximation and Edge AI techniques. The focus is on energy efficiency by covering newly proposed approximation and enabling frameworks. To the best of our knowledge, this survey is the first to review the latest approximate Edge AI frameworks and publicly available datasets in energy-efficient autonomous driving. The insights and vision from this survey can be beneficial for the collaborative driving service development on low-power and memory-constrained systems and also for the energy optimization of autonomous vehicles.

\end{abstract}
\IEEEpeerreviewmaketitle

\section{Introduction}
The use of sensors, advanced driver assistance systems (ADAS), and safety features in a vehicle shows a rising trend. The latest progression is towards integrating these sensors with the state-of-the-art deep learning architecture based on the sense, think, and act model, which can assist the driver or replace a driver by offering the highest level of autonomy \cite{r1}. The highest level of autonomy is described as the execution of driving processes that serve self-driving functionality from a source point to the destination point without any input or control to a vehicle from the human. Full automation can be achieved by integrating multiple sensors, such as camera, LiDAR, global navigation satellite system, radar, and communication modules with software-level solutions, thus providing the automotive driving features or the advanced driver assistance system \cite{adassensors2018kocic, adas2021antony}. The automotive industry is already using several simple and complex ADAS features for a long time, which has also improved the overall driver experience with the ultimate objective of providing better road safety \cite{safety-deep-learning, Deepfactors}. Braking assistance, lane departure warning, adaptive cruise control, and global positioning system (GPS) based navigation are some of the features that have been used since its introduction between 1990-2000 \cite{r6}. The current trend follows incorporating the deep learning and machine learning approaches within autonomous vehicles to provide maximum precision and human-level accuracy. The principle behind these statistically-based learning algorithms is to interpret the drivers surrounding when provided with impartial or neutral data. Based on the characteristics of the provided input, these algorithms classify or predict an output.

Some of the machine learning approaches have already replaced traditionally used algorithms in applications such as collision-warning systems \cite{lee2016real, CWS2018emb}, vision-based detection \cite{Pointpillars2019, Sharkawy2019Naecon}, path planning, lane change systems \cite{morris2011lane, hou2013modeling}, and recognition of multiple objects and further classification of them into traffic signs, cars, bicyclists, pedestrian to name few \cite{lv2014traffic, jia2016obstacle}. Although deep neural networks encounter the above-mentioned problems, the deployment on embedded and edge devices and related computational factors cannot be neglected. Therefore, this survey reviews the AI algorithms for connected vehicle applications; Edge AI approaches, and vehicular frameworks. From the above-mentioned topics, this survey focuses explicitly on energy-efficient mechanisms and approximate techniques. Figure~\ref{sections-survey} presents the taxonomy and topics covered in this paper. The outline of the sections in this survey paper is further divided as follows:

\begin{figure*}[!ht]
       \centering
  \includegraphics[width=.99\linewidth]{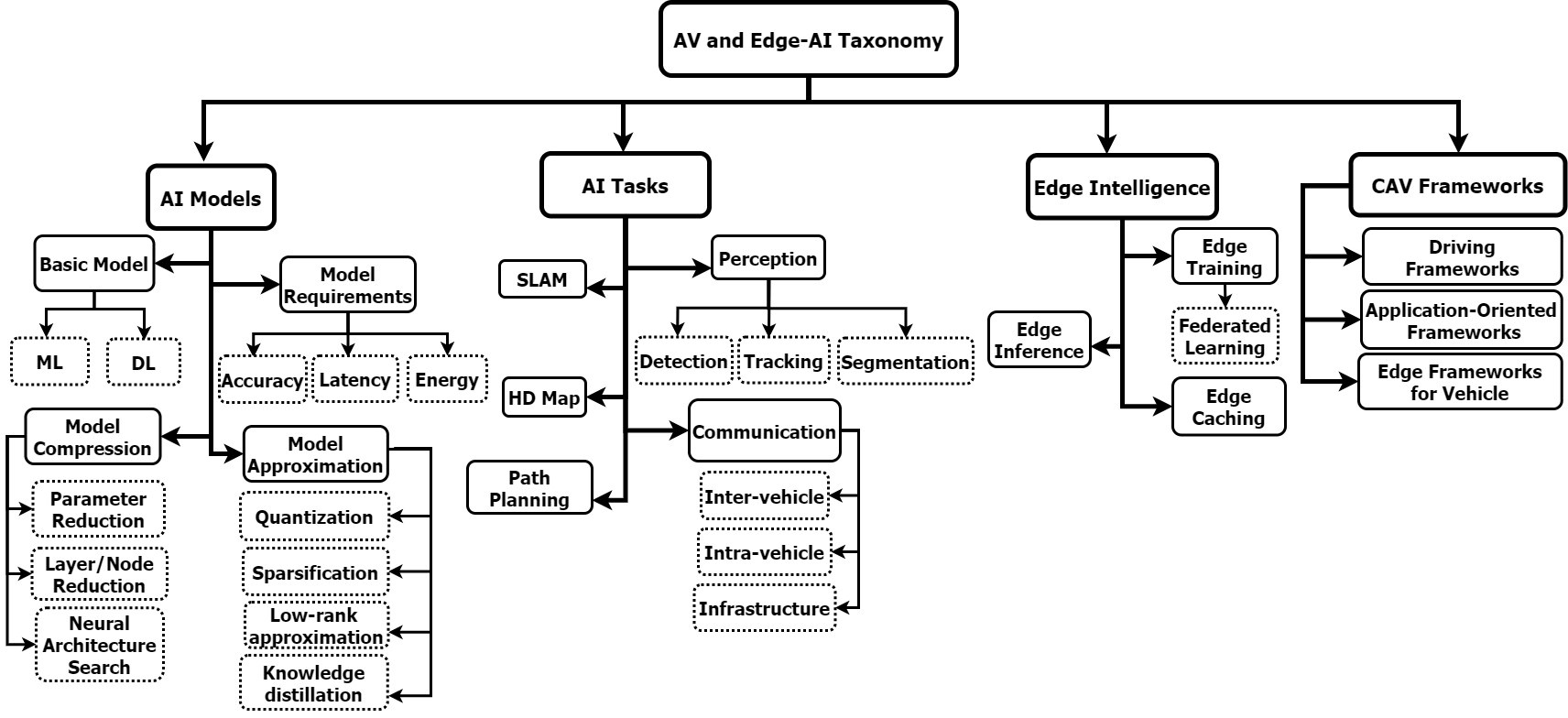}
  \caption{{Classification of Topics Covered in This Survey} \hspace{3cm}}
  \label{sections-survey}
\end{figure*}

\subsubsection{Motivation and Background}
This section illustrates the motivation and research questions targeted in this survey. It also discusses similar surveys and introduces background context for autonomous driving, approximate techniques, Edge AI, and vehicle communication.

\subsubsection{AI and Autonomous Driving}
In this section, the fundamentals such as Machine Learning \& Deep Learning approaches, are described. Autonomous driving services such as Perception, Localization, Path Planning, simultaneous localization and mapping (SLAM), and Vehicle-to-everything (V2X) are reviewed and compared based on state-of-the-art architecture and methodologies.

\subsubsection{Edge AI with Autonomous Driving}
This section discusses edge computing and the Edge Intelligence paradigm. This section reviews the articles published on cooperative driving, communication-efficient approaches, federated learning, Edge AI Inference, and Edge AI optimization methods.

\subsubsection{Enabling Frameworks}
This section covers the deep learning framework for autonomous driving and the Edge AI framework on computation, communication, and offloading capabilities, presented in the last few years. To the best of our knowledge, this survey is the first attempt to provide a review of the latest Edge AI frameworks for energy-efficient autonomous driving.

\subsubsection{Research Outlook and Open Problems}
This section summarises the survey by discussing open problems and potential challenges in deploying intelligent services within the vehicle-edge system. Further, this section contains information on approximation opportunities and enablers for edge intelligence approaches in autonomous driving services.

\begin{figure}[!ht]
  \centering
  \includegraphics[width=\linewidth]{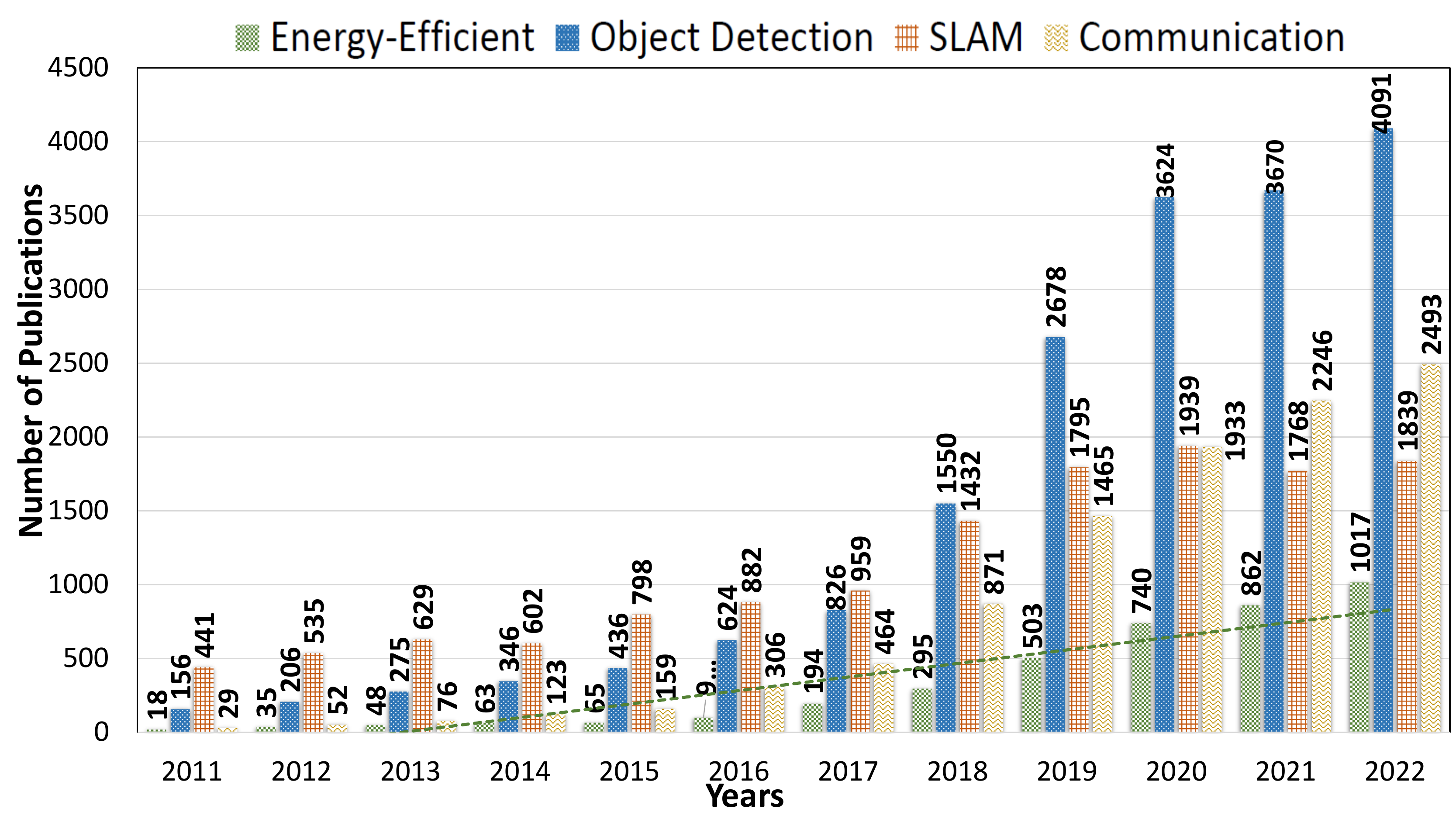}
  \caption{Publication trend in autonomous driving between 2011 and December 2022 (Source: ``scopus.com")}
  \label{num-paper-published}
\end{figure}

\section{Motivation and Background}

Autonomous vehicles and technologies have seen phenomenal growth. However, they are still far from being categorized as fully connected and autonomous systems. The current vehicular technologies need up-scaling and development in efficient communication, computation, reduced carbon emission, collaborative intelligence, and paramount safety. The primary focus and key research areas within the automotive domain were improving the performance parameters and developing baseline models and frameworks in object detection, SLAM, and vehicular communication, respectively.

To show the research trends in the autonomous driving domain, a graph is generated using data collected from the Scopus database. While collecting information from the Scopus database, the search is refined using popular keywords in the automotive domain, publication area (e.g., science, mathematics, information systems, and engineering), year range, and type of publication (e.g., conference paper, journal, books, chapters). The trend in the past decade, as shown in Figure~\ref{num-paper-published}, indicates that the primary focus was in the area of perception, specifically on object detection and segmentation, owing to the advancements in neural networks and datasets. SLAM and vehicular communication are becoming popular topics, with the latter catching up because of the recent development in 5G, next-generation cellular, and hybrid communication technologies. Energy-efficient approaches are showing a gradual increase, while the number of publications on energy-efficient methods is relatively less as compared to other subjects.

\textbf{Energy-Efficient keywords}: Energy-efficient Edge {\&} Vehicles, AI model compression {\&} approximation, TinyML, Energy-efficient Edge Framework, Vehicular communication compression {\&} Sparsification, Low-power Vehicular-Edge.

\textbf{Object Detection keywords}: Perception, 2D object detection, 3D object detection, adas classification, collaborative perception, cooperative perception, lane detection.

\textbf{SLAM keywords}: SLAM, EKF, KF, visual-slam, deep SLAM, pose estimation, graph SLAM, vehicular localization, vehicular mapping, Edge-SLAM, Deep-SLAM, Graph SLAM.

\textbf{Communication keywords}: V2X, V2V, V2I, C-V2X, 5G-V2X, DSRC, RSU, Vehicular communication, Inter-vehicular communication,  WiMax, Vehicular Networking.

The remaining of section covers the introduction, classification of topics, background of autonomous driving, software approximation approaches, Edge-Artificial Intelligence, and vehicular communication. The discussion is followed by requirements and needs to address the energy-efficient approximation for connected vehicular services.

\begin{table}[!ht]
\centering
\caption{List of acronyms used in this paper}
\label{acr-paper}
\begin{tabular}{@{}cc@{}}
\toprule
\textbf{Acronym} & \textbf{Definition}                             \\ \midrule
3GPP             & 3rd Generation Partnership Project              \\
4G               & Fourth Generation Technology                    \\
5G               & Fifth Generation Technology                     \\
AM               & Amplitude Modulation                            \\
ACC              & Adaptive Cruise Control                         \\
AEB              & Anti-Emergency Braking                          \\
AECC             & Automotive Edge Computing Consortium
    \\
ANN              & Artificial Neural Network                       \\
BLE              & Bluetooth Low Energy                            \\
BPSK             & Binary Phase-shift Keying                       \\
CAN              & Controller Area Network                         \\
CAV              & Connected Autonomous Vehicle                    \\
CCK              & Complementary Code Keying                       \\
CNN              & Convolutional Neural Network                    \\
COFDM            & Coded Orthogonal Frequency-division Multiplexing \\
CPU              & Central Processing Unit                         \\
C-V2X            & Cellular Vehicle-to-Everything                  \\
DAB              & Digital Audio Broadcasting                      \\
DNN              & Deep Neural Network                             \\
DSRC             & Dedicated Short Range Communication             \\
EKF              & Extended Kalman Filter                          \\
ETSI             & European Telecommunications Standards Institute \\
FDMA             & Frequency-Division Multiple Access              \\
FCC              & Federal Communications Commission               \\
FCW              & Forward Collision Warning                       \\
FL               & Federated Learning                              \\
FM               & Frequency Modulation                            \\
GFSK             & Gaussian Frequency Shift Keying                 \\
GNSS             & Global Navigation Satellite System              \\
GPS              & Global Positioning System                       \\
GPU              & Graphical Processing Unit                       \\
HD Map           & High-definition Map                             \\
IMU              & Inertial Measurement Unit                       \\
ITS              & Intelligent Transport Systems                    \\
KF               & Kalman Filter                                   \\
LTE              & Long Term Evaluation                            \\
M-QAM            & M-ary Quadrature Amplitude Modulation           \\
MANO             & Management and Orchestration                    \\
MFG              & Mean-Field Game                                 \\
MIMO             & Multiple-Input Multiple Output                  \\
ML               & Machine Learning                                \\
NR               & New Radio                                       \\
NX               & Next Generation                                 \\
NRF              & Neural Radiance Field                           \\
O-QPSK           & Offset Quadrature Phase Shift Keying            \\
OBU              & On-board Unit                                   \\
OFDM             & Orthogonal Frequency Division Multiplexing      \\
QPSK             & Quadrature Phase Shift Keying                   \\
RNN              & Recurrent Neural Network                        \\
ROS              & Robot Operating System                          \\
RSU              & Road Side Unit                                  \\
SGD              & Stochastic Gradient Descent                     \\
SLAM             & Simultaneous Localization and Mapping           \\
TPU              & Tensor Processing Unit                          \\
UWB              & Ultra Wideband                                  \\
V2G              & Vehicle-to-Grid                                 \\
V2I              & Vehicle-to-Infrastructure                       \\
V2N              & Vehicle-to-Network                              \\
V2P              & Vehicle-to-Pedestrian                           \\
V2V              & Vehicle-to-Vehicle                              \\
V2X              & Vehicle-to-Everything                           \\
WiFi             & Wireless Fidelity                              \\
WiMAX            & Worldwide Interoperability for Microwave Access \\ \bottomrule
\end{tabular}
\end{table}

\subsection{Autonomous Driving}
\begin{figure}[!ht]
    \centering
    \includegraphics[width=\linewidth]{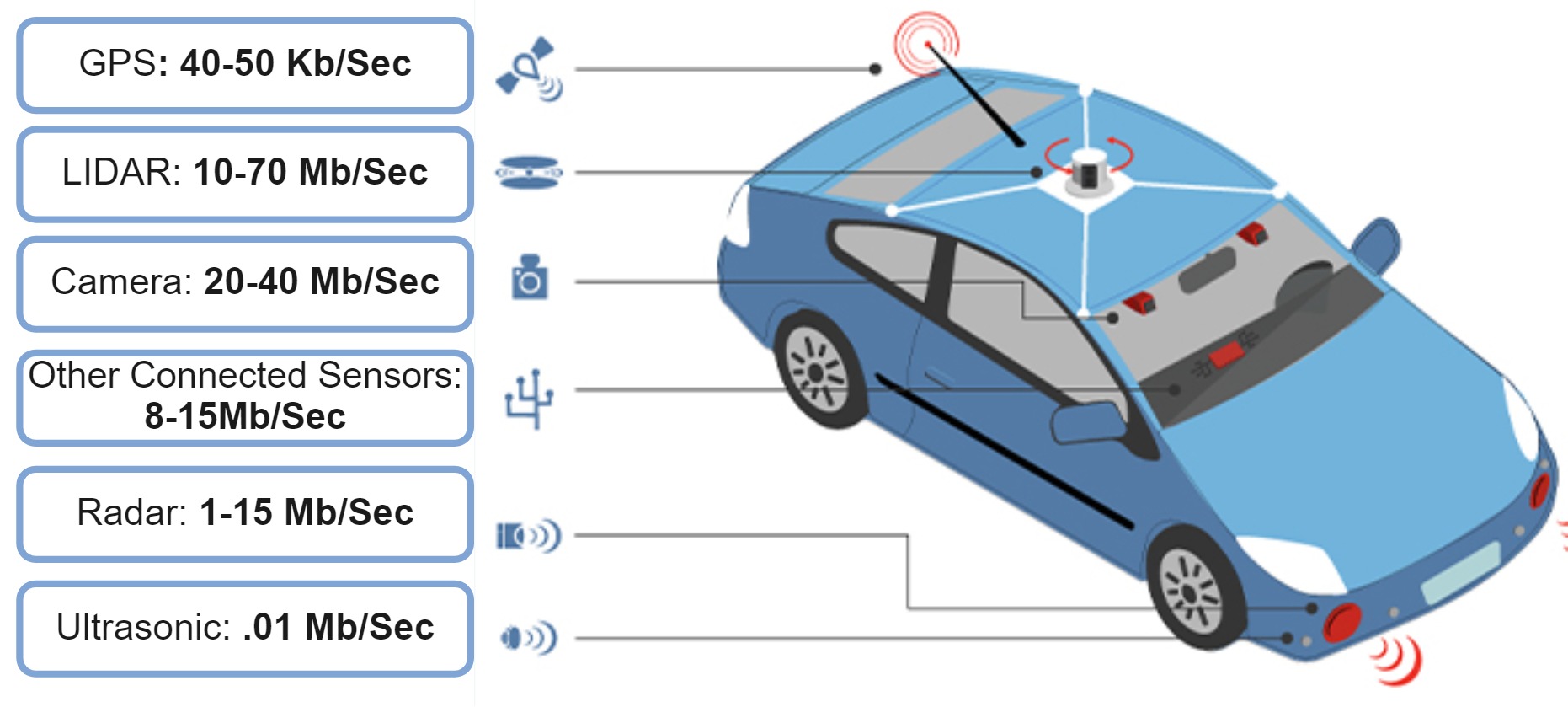}
    \caption{Data generated by the automotive sensors}
    \label{data-generated}
\end{figure}

Autonomous vehicles and on-board sensors generate a large amount of raw data that needs to be processed on-board by the vehicle computing unit, using DNN architectures and intelligent algorithms to enable driving services and applications. Figure ~\ref{data-generated} shows an approximate data rate from individual sensors in an autonomous car. The data rate may vary based on the sensor's specification (e.g., generation, bit-rate) and the data quality. Examples of vehicular applications using AI models are adaptive cruise control, object classification and obstacle detection, and SLAM. Studies from \cite{taiebat2019forecasting, krail2019energie} suggest that energy consumption from fully connected autonomous vehicles can be separated into three categories: 1) Consumption by an autonomous car (on-board sensors and Computing devices). 2) Energy consumption caused due to Infrastructure sensors involving Vehicular communication and Networking. 3) Energy consumption at the backend, for example, Edge servers, the central server maintaining legacy data, and the global DNN model. Studies \cite{liu2020computing} show that on-board energy consumption is higher than 1000's watts, and overall energy consumption from a single conditional automated diving vehicle combining all three categories could be around 2500 Wh per 100 km of driving \cite{krail2019energie}. High on-board energy consumption is due to the usage of compute-intensive algorithm and the processing devices such as graphics processors, which are essential for perception and visual applications.

Advanced driving assistance systems (ADAS) and features have been prevalent in the past decade. As shown in Figure~\ref{num-paper-published}, the research trend in the past decade has primarily been in the area of perception, specifically on object detection and segmentation, due tothe advancement in convolutional neural networks and the releases of autonomous driving datasets. SLAM and vehicular communication are also popular topics, with the latter catching up because of the recent development in 5G, next-generation cellular, and hybrid communication technologies. Energy-efficient approaches are also showing an increase in research trends. However, the number of publications on energy-efficient methods is relatively less as compared to other directions. 


\begin{table}[!th]
\caption{Number of sensors present in an autonomous car and an approximate count of sensors for level 4 and level.}
\label{T:Sensors_Count}
\begin{tabular}{|c|c|c|c|c|c|}
\hline
\multicolumn{6}{|c|}{\textbf{Sensors Count Approximately Present in an Autonomous Car}}                               \\ \hline
\textbf{Sensor}        & \textbf{Level 1} & \textbf{Level 2} & \textbf{Level 3} & \textbf{Level 4} & \textbf{Level 5} \\ \hline
\textbf{Control Units} & 1                & 1                & 2                & 3                & 3                \\ \hline
\textbf{Ultrasonic}    & 5                & 5                & 9                & 9                & 9                \\ \hline
\textbf{Radar}         & 2                & 4                & 4                & 8                & 8                \\ \hline
\textbf{Camera}        & 0                & 2                & 5                & 5                & 5                \\ \hline
\textbf{LiDAR}         & 0                & 0                & 1                & 2                & 2                \\ \hline
\textbf{GPS/GNSS}      & 1                & 1                & 1                & 1                & 1                \\ \hline
\textbf{DSRC}          & 0                & 1                & 1                & 1                & 1                \\ \hline
\textbf{V2X Module}    & 0                & 1                & 1                & 1                & 1                \\ \hline
\end{tabular}
\end{table}

The on-board computation approaches leading to power consumption \cite{buzatu_2019} demand the design of applications and energy-efficient Edge AI systems for automated driving services. Therefore, this survey paper focuses on identifying currently practiced AI algorithms and computation approaches that lead to high energy consumption. Further, it comprises of review from, design, and implementation of edge computing approaches for the autonomous driving services (for example, Perception, HD Map, SLAM), datasets, edge-assisted techniques, and vehicle-edge frameworks. Lastly, based on the gathered requirement and research gaps, an Edge AI processing pipeline is proposed, which contains the higher-level abstraction of components involved in service implementation across vehicle-edge settings. In this survey, the levels of autonomy is referred from the International Society of Automotive Engineers (SAE), consisting of six levels of automation in driving, which are as follows:

\begin{enumerate}
\item \textbf{Level 0 - No Automation}: All driving tasks are carried out by driver.

\item \textbf{Level 1 - Driver Assistance}: Driving tasks are carried by driver with little input from the vehicle sensors, this level introduces driving assist features.

\item \textbf{Level 2 - Partial Automation}: Driving tasks can be carried by computing unit placed in car with sensed input from the vehicle surrounding, the features include adaptive cruise control, autonomous emergency braking, however this level still requires the driver to maintain control of driving tasks and regularly monitor the vehicle surrounding.

\item \textbf{Level 3 - Conditional Automation}: Some tasks (sensing, actuation and control) are carried out by the sensors and the computing unit placed in the car, however the driver must be able to take control of the vehicle based on demand and situation.

\item \textbf{Level 4 - High Automation}: Vehicle is capable of performing all driving tasks by initiating communication with other vehicles under certain conditions, but the driver has the option to take control of vehicle.

\item \textbf{Level 5 - Full Automation}: Vehicle is capable of performing all driving tasks by communicating with other vehicles and infrastructure sensors under all conditions, but the driver may have the option to control the vehicle.
\end{enumerate}

\subsection{Approximate Techniques}
DNN applications such as 3D object detection and classification or SLAM are usually computationally intensive, memory-consuming, and energy-consuming tasks. The computing complexity increases for real-time applications when these larger-weight DNN are implemented on embedded systems with limited memory and computing power. For example, the currently deployed level 3 autonomous vehicles \cite{krok2020tesla, waymo_2020, apolloauto2021} primarily depend on vision sensors and systems and consume significant resources in terms of memory and energy. The scalability of these embedded systems with fully connected cooperative autonomous vehicles is yet to be known, incorporating full ADAS features. When these features are integrated into the resource \cite{RC1, RC2, RC3, RC4} and energy-constrained \cite{EE10, zeng2021energy, liang2021energy, EE7} real-time autonomous systems, the following challenges will be encountered: A) When the large volume of sensor data is processed over the DNN algorithms for autonomous driving services, it will directly affect the computing efficiency of the embedded systems with limited memory, which makes it essential to implement compression technique \cite{RC5, RC6, RC7, stahl2019fully} thus approximating the algorithms and embedded device usage and simultaneously optimizing it for better energy-efficiency. B) The computing complexity and low latency for applications such as SLAM makes it necessary to process the sensed data at the on-board computing unit rather than processing it at the Edge-server. Software and architectural approximation techniques such as data aggregation, and early-exit neural networks can help improve the on-board low latency and fast inference. This is comprehensively covered in section IV.

\subsection{Edge AI}
Edge AI or Edge Intelligence can be described as the combination of Edge Computing, and Artificial Intelligence \cite{r11}. It has emerged due to the requirements from the connected ecosystems, developed for the applications that require the processing of algorithms locally on the device or in the nearest available data center or server. The algorithms \cite{communication-efficient2020} utilize the data generated by the devices and make independent decisions for real-time applications without needing to connect to the centralized server or cloud for the decision-making process. A fully connected Level 5 autonomous car will result from collaboration between edge-sever communications and computing systems.

The current level 1 - level 3 autonomous vehicle highly relies on the Graphics Processing Unit (GPU chip) for their applications, and the GPU alone can consume up to 300-350Wh \cite{buzatu_2019, apolloauto2021, krail2019energie} of energy per 100 km of driving depending upon the data rate and quality of the sensors. As shown in Table~\ref{T:Sensors_Count}, the number of sensors increases for the fully-connected autonomous vehicle compared to the current scenario. The information shown in the table is an approximate estimate. Based on OEM and fleets \cite{ackerman2021full, apolloauto2021, krok2020tesla}, sensor distribution may vary according to the sensor kit and software technologies used in conjunction with it. The estimated power consumption of each vehicle can be from  100's to 1000's watts depending on the type of operation the vehicle is involved in. As per the reports \cite{toyota-data} the amount of data transmitted between the vehicle and the cloud can reach 10 exabytes in the future, which is excessive with respect to current practices, and the present cloud and server are not capable of handling and processing this real-time data quickly. Therefore, AI at the Edge can be implemented to process the data for time-sensitive tasks in the autonomous driving ecosystem and offload it.

For low-latency tolerable applications, such as HD map updates or traffic incidents sharing, Edge AI approaches can process the data locally at the edge-server, and later transmit the model or analysis of the result to the cloud or remote-server, by following energy-efficient mechanisms. In a fully automated driving environment, the Edge AI implementation can help in achieving better end-to-end accuracy by bringing down the current power and energy consumption. The Edge AI deployment process involves sensing, re-training, decision making, and collaborative learning through communication with other edge devices, and servers in the environment.

\subsection{Communications in Autonomous Vehicles}
\begin{figure}[!ht]
  \centering
  \includegraphics[width=\linewidth]{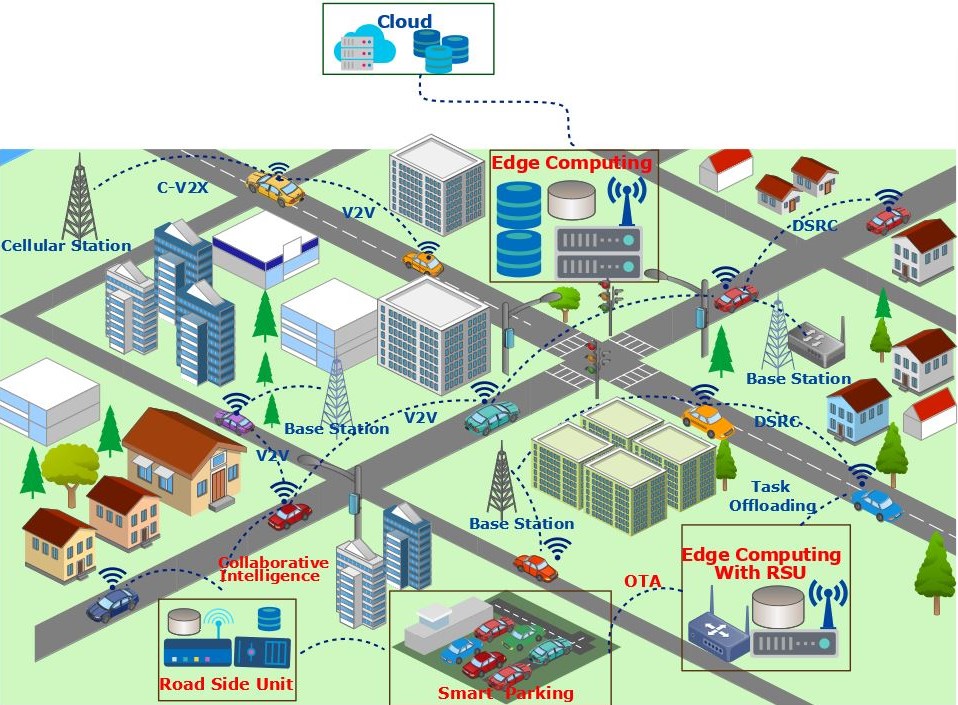}
  \caption{Communications in vehicular ecosystem across vehicles, infrastructure, and road-side networks.}
  \label{veh-comm}
\end{figure}

Communication in the vehicular ecosystem is a key to deploying cooperative and collaborative autonomous driving applications. An example of connected vehicles, base stations, road-side units, edge-servers, infrastructure and remote cloud is shown in Figure~\ref{veh-comm}. Several use-cases presented within the context of vehicle communication \cite{vc21, vc22, vc23, vc24, vc25, vc26}, discuss directly benefiting the perception, planning, and control related use-cases and subsequently impacting the energy consumed by the vehicle. Communications in vehicle can be further categorized as: Inter-Vehicle Communication \cite {vc11, vc12, vc13} \& Intra-Vehicle Communications \cite{ vc27, vc28}.

Intra-vehicle communication helps understand the vehicle's current state by exchanging information and signals between the sensors, actuators, and other electronic devices and components present within the vehicle. This communication is a combination of wired and wireless technologies. Commonly used wired technologies include Controller Area Networks (CAN), Digital Data Bus (D2B), Ethernet, FlexRay, Media Oriented System Transport (MOST), Low Voltage Differential Signaling (LVDS), Power Line Communication (PLC), Time-Triggered Fieldbus (TTP). Conversely, the wireless communication methods for Intra-vehicle communication include WIFI, BLE, Zigbee, and Ultra Wideband. Amongst the mentioned wireless technologies, BLE is one of the most commonly used by automotive manufacturers as it is a significantly proven technology and is relatively cheap compared to WiFi. It can transmit media relatively faster than Zigbee and comprises a good security layer. A comparison of these communication technologies is also shown in Table~\ref{t:longshortrange}.

The most important factor for the high use of BLE technology is relatively low power consumption \cite {PE1, PE7} and it has a large installed base and a guaranteed latency, as well as a stable specification. Automobile components and modules, normally connected by electrical signal wires, are increasingly being replaced by wireless signals. A reduction of 50\% in the number of signal wires is the goal of the automotive industry. Typically, an automobile contains about five kilometers of wiring, so there would be many wireless signals. A hybrid practice that uses both, wired clusters of automobile components and wireless inter-cluster connections is becoming more common. The infotainment panel at the vehicle dashboard is such an example. For Inter-Vehicle communication, the present human-driven or semi-autonomous vehicles are equipped with communication and radio modules, which receive information and signals mostly related to infotainment. The communication technology has evolved from AM, FM, DAB to HD Radio in which transmission method, media size, and quality of service have significantly improved. Since fully connected autonomous driving has wider communication and real-time processing requirements as the high-performance computing unit takes the decisions, researchers have proposed relevant technologies such as DSRC, V2V/V2I, WiMax, 5G-NR-V2X or C-V2X for local and long-range communication.

\subsection{Taxonomy of Edge AI Technologies for CAV}
This subsection introduces the taxonomy used in the remaining of this survey paper. First, AI methods used in autonomous driving are described. Second, Edge AI computing applications for autonomous vehicles are explained. Third, the approximation approaches and compression strategies are defined. Finally,  energy-efficient mechanisms and requirements in the vehicular ecosystem are discussed. For reference, the topics can be seen in Figure~\ref{sections-survey}.

\begin{enumerate}

    \item AI Models {\&} Autonomous Vehicles: An autonomous vehicle is an independent system capable of routing from source to destination by perceiving its surroundings using sensors and processing the sensed data on intelligent algorithms. Advancements in autonomous vehicles and related technology can be associated with the progress of vehicle sensors suite and intelligent algorithms/models. These models have enhanced connectivity, infotainment systems, electrification, and automation. Perception sensors (camera, LiDAR, radar), positioning sensors (GPS, GNSS), and communication modules are used to replace or imitate human driving behaviour using AI models. 
    
    \begin{enumerate}
        \item Basic Model: AI models proposed to automate/assist driving tasks can be divided as follows:
        
        \begin{enumerate}
        \item Machine Learning: Supervised, unsupervised, and reinforcement are the popular techniques explored within autonomous driving.
        \item Deep Learning: It is a subset of machine learning that consists of several types of neural networks trained on datasets to learn complex features from unstructured or structured data.
        \end{enumerate}

        \item Model Requirements: AI models have specific requirements and guidelines depending on the driving tasks. For e.g., localization, emergency braking, and detecting an obstacle/traffic sign should be highly accurate. Within the scope of this survey paper, the discussed model requirements are:
        
        \begin{enumerate}
        \item Accuracy: The principle behind using AI models is to eliminate human error while driving and achieve an expected level of accuracy for the driving tasks. It is measured as a score of correct predictions/estimation with respect to the total predictions by a model.
        \item Latency: Each driving tasks have varied execution requirement. For e.g., detection and localization have strict requirements of a few milliseconds(ms). For AI models, latency (in time) is used to characterize the performance of a model for a specific application.
        \item Energy: Desiring the highest level of accuracy for an AI model and fulfilling strict latency requirements for specific tasks generally leads to the use of high-performance computing units, which leads to energy consumption. Energy (Joules) can be estimated by capturing AI models' power consumption (Watts).
        \end{enumerate}

        \item AI models compression: These techniques enable processing large volume data or AI models, such as dense and deep neural networks on resource-constrained devices with limited computation resources. For vehicular applications, lossless and lossy compression has been explored for models and data. Popular compression approach includes:
    
\begin{itemize}
        \item Parameter reduction: Reducing parameters from the AI model can compress the model, which may lead to deployment on resource-constrained devices. For e.g., the pruning of non-contributing weights/layers results in parameter reduction, which generally leads to model compression. 

        \item {Layer/Node reduction:} To reduce the compute and memory requirements of neural networks layer/node reduction approach is adopted.  Generally practised as a structured approach to reduce computational demand, while balancing the model accuracy. Minimal matrix operations and parameter-sharing are some examples.
        
        \item {Neural Architecture Search:} This compression category can be seen as optimizing the parameters/hyper-parameter of neural networks with a search dimension. Model downsizing, and balancing high communication bandwidth demand within the vehicular environment can be such search dimensions.
\end{itemize}
    
    \item Approximate Techniques: These techniques are unconventional approaches from the area of mathematics which has several use-cases in the field of science and engineering (e.g., numerical precision, circuit design and tools). In approximation, a balanced mechanism is used to trade-off metrics and parameters quantitatively for achieving fast computation (on-board latency) by trading-off computing performance (precision) \cite{AC6, AC5, AC4}. Software and model compression approaches proposed for framework and AI models in connected autonomous vehicles can be categorized as approximate model or approximation techniques. However, this generalization do not address energy-efficiency (one of the three dimensions in approximate computing) from the viewpoint of computation and communication.
        
        \begin{enumerate}
        \item Quantization: Vehicular applications are dependent on intelligent algorithms, which generally use 32-bit floating point precision for training the model and gradient estimate. The elements can be approximated using quantization to fewer bits, reducing the model size and decreasing the bandwidth load. The approach is inspired by the human nervous system, where information is stored in discrete form \cite{tee2019quantized}.  
        
        \item Sparsification: In this approach, a vector is represented by its approximate form where the non-zero components are equal to the corresponding original vector. It is a compression technique often implemented in collaborative and distributed learning approaches such as federated learning which requires frequent communication between the devices, in this case, between the vehicle and edge or cloud. 
        
        \item Low-rank approximation: Another technique to implement reduced computation for AI models in low-rank approximation. Tucker or Canonical polyadic decomposition has been well used for CNN and DNN. The technique successfully reduces the model size, but it also results in a significant loss in the model's accuracy. 
        
        \item Knowledge Distillation: It is an approach to approximately represent a larger DNN model in compressed/reduced form. Although the technique allows the development of approximate versions of AI models, complexity and open challenges remain in knowledge transfer.
   \end{enumerate}
\end{enumerate}

\item AI Tasks: Driving tasks implemented using AI models can be categorized as perception, SLAM, HD map, path/motion planning, and communication. The AI model and respective driving tasks can be further differentiated on the basis of data processing, feature extraction mechanisms, and hardware used.
    
    \begin{enumerate}
    \item Perception applications provide scene understanding and are performed using vision sensors such as a camera or LiDar, at the vehicle's on-board computing unit or at the sensor units present within the ecosystem (e.g., CCTV cameras with the computing unit). These applications are performed using CNN or DNN models deployed on the GPU. As the models largely consist of dense layers, the computational demand and energy cost for deployment are relatively high.
    \item SLAM application enables vehicles to localize in their surrounding using sensor data. AI models enabling SLAM applications are also memory and compute-intensive. The complexity further increases because of the low inference requirement/processing of these algorithms.  
    \item HD map sometimes also referred to as 3D map is an evolving driving service/feature, which provides visuals of the vehicle surrounding replacing the currently used 2D maps. It is expected to be used with detection and localization tasks.
    \item Communication in the vehicular environment is dynamic and heterogeneous. It exists in three forms; in the vehicle, between vehicles and within the infrastructure. With the evolution, vehicular communication depends on the generation of hardware/software and sensory technologies. High-level autonomy is highly dependent on connected vehicles and smart infrastructure sharing raw data, weights, and algorithms. Similar to on-board computation, the complexity in vehicular communication arises due to the large volume of data and additional load on bandwidth.
    \item Path/Motion Planning is a crucial AI task that enables the vehicle to navigate from source to destination by avoiding obstacles. A traditionally used algorithm is A-star. However, recent approaches involve using AI models with vision sensors, thus combining motion planning and path prediction by avoiding obstacles.
\end{enumerate}
    
    \item Edge AI and CAV: Initially, cloud computing was proposed to facilitate computation, and decision-making for the connected vehicles \cite{AVE, PI-Edge, ibn2021edge}. However, the cloud computing approach had several challenges in transmitting high volume or flood of data from the vehicle to the cloud, data privacy and leakage, adversarial and poisoning attacks on the ground truth data, and algorithms present in the cloud \cite{ibn2021edge}. Therefore, an approach to bring computation near the data source to tackle surplus data transmission to the cloud has been proposed in the form of edge computing. This technique has been further enhanced by proposing Edge-Intelligence, which allows the deployment of AI applications on Edge devices to facilitate inference near the data source. Edge AI improves data privacy and security and shows promising aspects in tackling the distributed computation and communication challenges for the connected vehicular ecosystem, which consists of services such as driver's assistance, infotainment, decision-making, and safety-critical applications. It is further divided into Edge training, Inference and Caching. 
    
    \begin{enumerate}
        \item Edge Training: As future vehicular applications will be carried out in dynamically distributed and connected environments, edge training can enable and facilitate collaborative/joint learning within participating devices using federated learning. It also allows re-training and updating models.

        \item Edge Inference: Edge inference enables deployment of AI model in resource-constrained devices. Considering the complexity of deploying fully connected autonomous vehicles and the severity, the following concerns should be addressed:
    \begin{enumerate}
        \item Latency: The vehicular environment is complex, and many applications have strict latency requirements. In fully connected vehicles involving AI applications, latency includes sensor data processing, data fusion, algorithm processing or computation, and communication between devices.
        \item Real-time Inference: Deploying real-time applications is essential for connected autonomous vehicles. The adjacency of computing to the data source tackles the low-latency and time-sensitive requirements. However, high computational and relative energy costs should be considered in such deployment cases.
        \item Offloading: For resource-constrained devices (low compute and battery powered), offloading data and computation to the nearest edge servers can facilitate local deployment, which also reduces the traffic amount from the vehicles/edge devices to the cloud. 
        \item Heterogeneity: In a vehicle-edge-cloud ecosystem, heterogeneity exists in the sensed data, computing capabilities, communication devices, and protocols. This property poses significant challenges for deployment and resource management strategies.
        \item Reliability: Possibility of deploying low-latency and real-time applications makes Edge AI reliable for vehicular applications. Also, it prevents sharing of sensitive and safety-critical data. However, rural or highway driving, communication, congestion, packet delay, and bandwidth requirement are concerns.
    \end{enumerate}
    
        \item Edge Caching: As training/re-training, updating the weights, and model in a distributed environment require frequent data exchange, caching becomes an essential and important function, which deals with the collection, storing, processing, and real-time labelling of data. 
        
    \end{enumerate}
    \item CAV Frameworks: Advancements in sensory technologies, AI models, driving tasks and on-board processors/computers have resulted in the development of autonomous driving frameworks. These driving frameworks can be currently categorized as driving task/assist oriented, independent application/service oriented or as compute-communication oriented frameworks.

\end{enumerate}

\begin{table*}[!ht]
\caption{Coverage and Comparison of previously published Survey}
\label{t:survey-comparison}
\begin{center}
\begin{tabular}{| c | c | c | c | c | c | c | c |}\hline
\textbf{Previous Work} & \multicolumn{7}{ c |}{\textbf{Topics Covered}}  \\ 
\cline{2-8}
& \textbf{Perception} & \textbf{SLAM} & \textbf{Comm} & \textbf{HD Map} &  \textbf{Dataset} & \textbf{Edge AI} & \textbf{Energy Efficient}\\
\hline
{\color{red}This Survey} &{\color{red}Y}  &{\color{red}Y}  &{\color{red}Y}  &{\color{red}Y}  &{\color{red}Y}  &{\color{red}Y} &{\color{red}Y}\\ 
\hline
2018 - Autonomous Driving Cars \cite{sv1} &Y  &Y  &Y  &N  &N  &Y &N\\ \hline
2019 - Edge Computing System \cite{sv2} &N  &N  &Y  &N  &N  &Y &N\\ \hline
2019 - Edge Computing For AD \cite{sv3} &Y  &Y  &Y  &Y  &N  &Y &N\\ \hline
2019 - Edge Intelligence for IoV \cite{sv4} &Y  &Y  &Y  &Y  &N  &Y &N\\ \hline
2020 - AD: Common Practices \cite{sv5} &Y  &Y  &Y  &N  &Y  &N &N\\ \hline
2020 - Deep Learning for AD \cite{sv6} &Y  &Y  &N  &N  &Y  &N &N\\ \hline
2020 - Energy Aware \cite{sv7}  &N  &N  &Y  &N  &Y  &Y &Y\\ \hline
2020 - Communication-Efficient \cite{communication-efficient2020} &Y  &N  &Y  &N  &N  &Y &N\\ \hline
2021 - Edge Computing \cite{sv9} &N  &N  &Y  &N  &Y  &Y &N\\ \hline
2021 - Edge-Benchmarking \cite{sv10}  &N  &N  &Y  &N  &N  &Y &Y\\ \hline

\end{tabular}
\end{center}
\end{table*}

\subsection{Motivation and Methodology of Choosing Literature}

In past years, detailed survey in emerging autonomous driving technologies \cite{sv1, sv3}, common practices \cite{sv5}, deep learning techniques\cite{sv6}, and communication-efficient \cite{communication-efficient2020} approaches has been published. However, little to no attention has been given to energy-efficient approaches and related software approximation techniques for connected autonomous vehicles. In \cite{sv1}, an overview of current and emerging autonomous driving technologies by following the case-study approach is presented. While discussing emerging technologies, the authors also briefly described the future research opportunities in connected autonomous vehicles. A comprehensive study of edge computing systems and edge computing opportunities for autonomous driving is presented in \cite{sv2, sv9} and \cite{sv3} respectively. The review paper gives attention to computing architecture, software framework, privacy, and security in vehicular communication. In a similar context, \cite{sv4} presented a review of mobile edge intelligence techniques for vehicles and discussed edge-assisted perception, mapping, and open issues. Articles \cite{sv5, sv6} covered recent autonomous driving state-of-art AI models and techniques in detail. Key discussed topics were machine/deep learning models, driving safety features, system components, and architecture. The review conducted in \cite{sv7} covers energy-aware approaches for hardware and software layers in the edge computing domain, focusing on the framework layer. By focusing on key communication challenges, authors in \cite{communication-efficient2020} presented a comprehensive review of communication-efficient techniques for edge computing systems. In \cite{sv10}, authors reviewed cloud-edge computing, and popular frameworks by focusing on application and optimization techniques and benchmark runtime.

\begin{figure}[!ht]
  \centering
  \includegraphics[width=\linewidth]{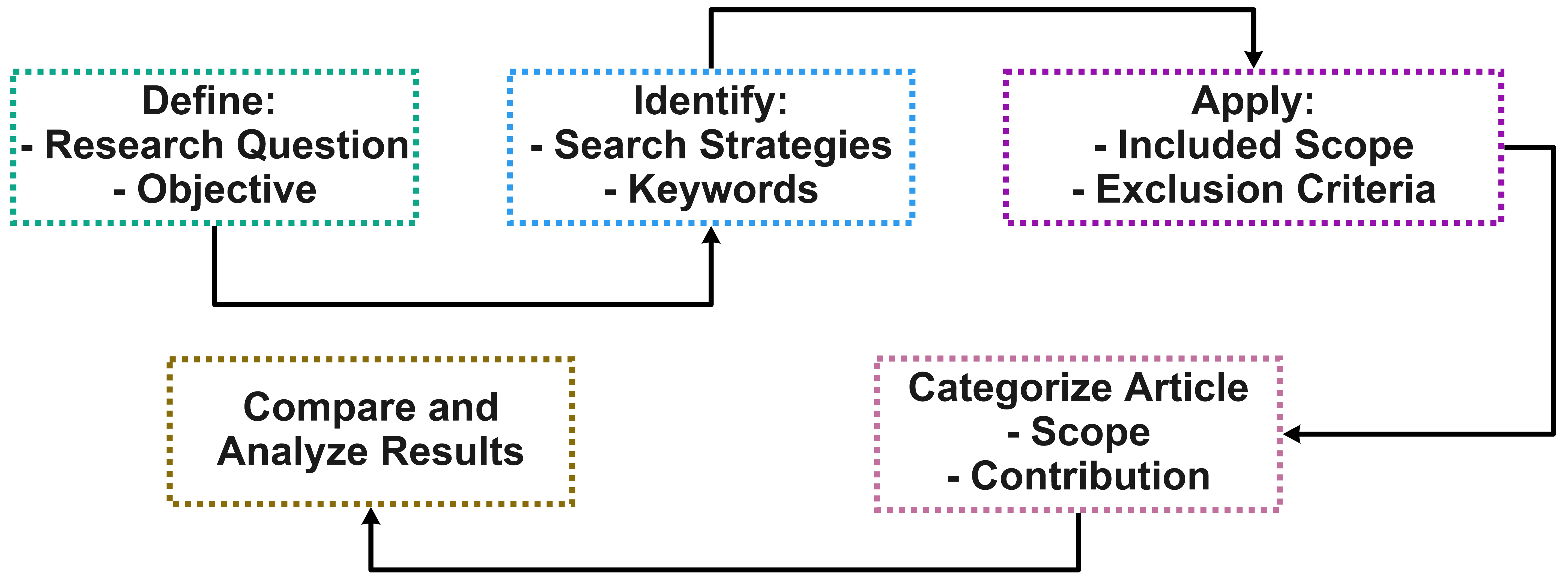}
  \caption{Approach for systematic literature review adapted from Kitchenham and Charters \cite{kitchenham-charters}}
  \label{kitchenham}
\end{figure}

To highlight the value of this survey, a comparison with related surveys is shown in Table~\ref{t:survey-comparison}. This comparison table is based on coverage of topics: deep learning practices (perception), data \& compute-intensive tasks (SLAM, Communication, High-definition Maps), datasets, applications of Edge Intelligence, and related energy-efficient approaches. The review procedure utilized in preparing this literature survey is based on SLR approach adapted from Kitchenham and Charters \cite{kitchenham-charters}, also shown in Figure~\ref{kitchenham}. This approach demands defining of research questions and objectives initially, followed by identifying the search strategies. While searching the relevant and related content, a connected paper search approach is followed, the inclusion and exclusion criteria will be applied with the keywords and terms to refine the article based on the scope and objectives. In the last two stages of the SLR approach, the collected articles are categorically divided based on the article's contribution toward approximation techniques, autonomous driving applications, and Edge Intelligence. Some approximation techniques overlap in multiple research questions. Therefore, a combined approach is used for review. 

\begin{tcolorbox}[title=\textbf{Key Questions Addressed}, breakable, skin=enhanced jigsaw]
\begin{enumerate}

\item What are the current AI model development and deployment strategies for connected vehicular tasks/applications such as perception, SLAM, vehicular communications, and HD maps?


\item What are the recent communication-efficient approaches that are proposed in a vehicle-edge ecosystem for CAV?

\item Which approximation strategies are proposed as software-level solutions for communication and computation in vehicle-edge environments?


\item What are the vehicle-edge framework development strategies that can enable vehicular services using joint inference, by using efficient computation and communication techniques?
\end{enumerate}
\end{tcolorbox}

\section{AI \& Autonomous Driving}
AI/Machine learning approaches and techniques have been widely used for autonomous driving tasks and services. Commonly used techniques are supervised learning, unsupervised learning and semi-supervised learning \cite{indyk2019learning, DB48}. In supervised learning a machine learning model is trained with labelled dataset, while in unsupervised learning a machine learning model is trained with unlabeled dataset, with the common purpose of prediction or classification. In semi-supervised learning a machine learning model is trained with both labeled and unlabeled datasets. This approach is proposed to save training time and computational resources \cite{jeong2020federated, diao2021semifl}.

\subsection{Perception}
Autonomous vehicles driven using sensory technologies and AI algorithms can be seen in the form of taxies from Waymo, Zoox, Cruise etc.\cite{ackerman2021full, gibbs2017google}. These vehicles are mostly dependent on Perception related tasks: segmentation, Object classification-detection and localization. These three tasks are currently considered as crucial element for the enablement of autonomous driving. The object detection task can be further divided into 2D or 3D detection, which are mainly reliable on the line-of-sight sensors such as High-Definition Camera \cite{IntelRealsense2021, mobileye} and LiDAR \cite{LiDar2020}. 2D object detection task is generally carried using convolutional neural network and recurrent neural network architecture which involves feature detection and estimation of rectangle or square shaped bounding boxes \textit{(x, y)} around the detected objects in an image or video frame, whereas the 3D detection involves estimating a cube shaped, three dimensional bounding box in an object, by estimating the position of the object in the 3D plane \textit{(x, y, z)}. 

Deep learning has been widely accepted as attractive or prominent technique for image and vision related applications because of development of the state-of-the-art neural network architectures \cite{googlenet, Imagenet, RCNN-Liang-2015}, and their delivered accuracy's. The object detector are classified into one-stage and two-stage detectors depending upon the backbone of training and inference method used. Table~\ref{t:detectors} covers popular and recently published object \& lane detection approaches for autonomous driving. Table~\ref{t:detectors} is formulated on AI model performance over the popular driving datasets (covered in Table~\ref{t:Datasets}), hardware implementation, detection methods, and speed (FPS) which is crucial for real-time deployment. For the 3D detection the initial approach and technique involves pre-processing of the 3D point clouds data and adopting them into the data structure required for the existing deep learning algorithms, thus providing an output based on the algorithm. Recent researches have proposed to process the LiDAR point clouds directly on deep neural network without converting them to any representations. For example \cite{qi2017pointnet, 3d10} proposed different form of deep neural net architectures, called as Pointnets and Frustum Pointnets respectively. These deep learning architectures have shown higher performance and have proved as benchmark for 3D perception based detection such as object classification and semantic segmentation. Pointnets++ architecture \cite{qi2017pointnet++} proposed by Qi et al. is capable of both classification and semantic segmentation of 3D point clouds by learning the local and global feature vector from the raw point clouds. Zhou et al. presented VoxelNet \cite{voxelnet}, a deep learning architecture detecting 3D bounding boxes  based on reading of LiDAR Point clouds, here the LiDAR point clouds were divided into 3D voxel spaced equally. The architecture successfully detects and gives high performance for the car, cyclist and pedestrians. The most prominent 3D object detector Frustum-Pointnet \cite{3d10} is presented by Qi et al., which predicts the bounding box on an object based on instance segmentation and the bounding box estimation. A similar method Pointfusion \cite{3d4} is proposed by Xu et al. which utilizes the Pointnet \cite{qi2017pointnet} and ResNet \cite{ResNet} architecture for estimating the 3D frustum and object classification.

\begin{figure}[!ht]
  \centering
  \includegraphics[width=\linewidth]{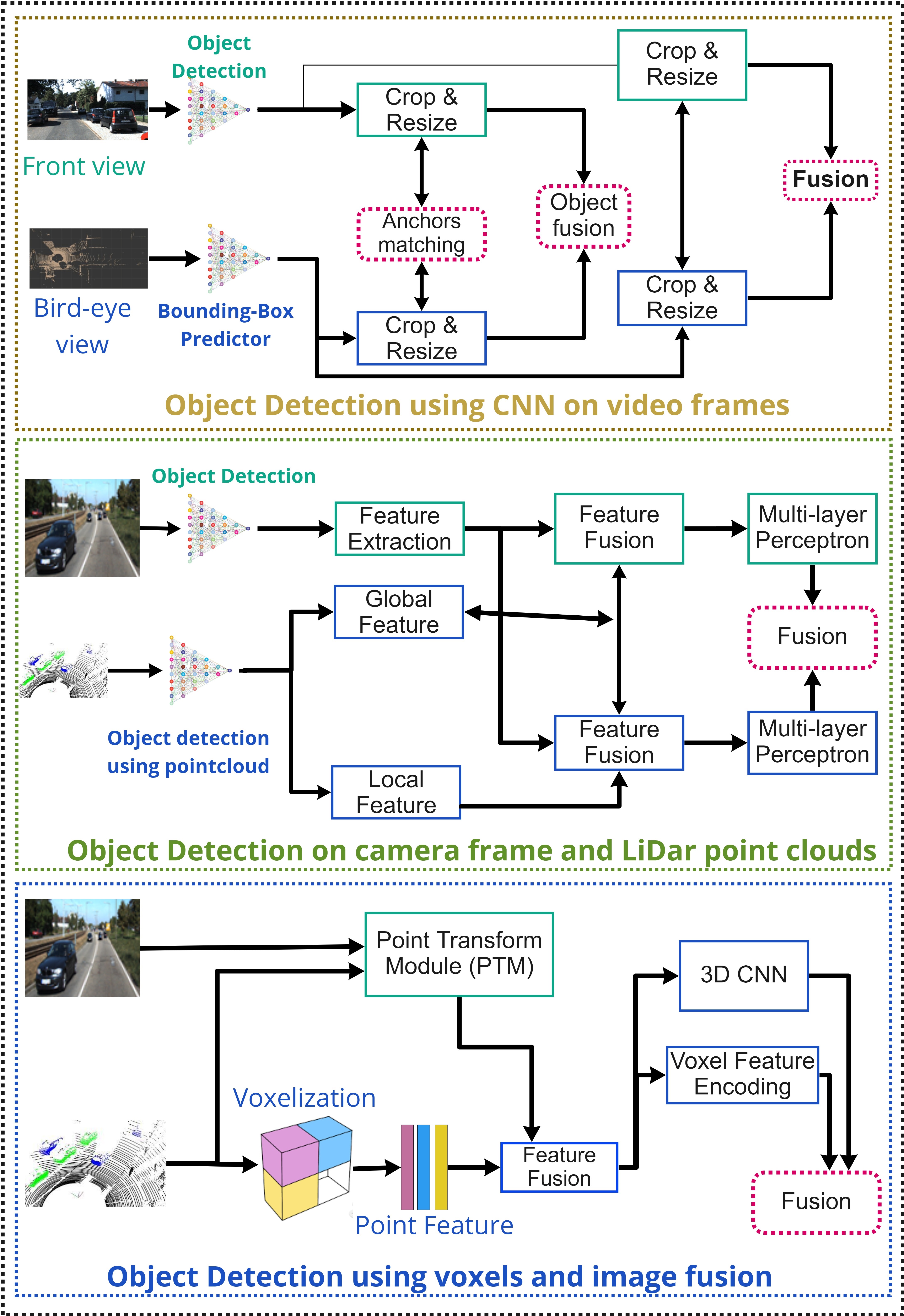}
  \caption{DNN pipeline to show 3D object detection using video frames, bird-eye-view, and LiDAR point clouds.}
  \label{3d-det}
\end{figure}

\subsubsection{2D object Detection}
2D object detection in an autonomous vehicles are primarily based on the single or multiple cameras connected to sense the environment or surrounding of the car. The 2D object detection architecture or algorithm requires the raw image as an input, and outputs the bounding box with the class or label of the detected object. In 2D object detection the bounding box is an axis-aligned rectangle, which is precisely estimated on the position of the multiple objects or classes in that image, here the bounding box can be parameterized as (x\textsubscript{min}, x\textsubscript{max}, y\textsubscript{min}, y\textsubscript{max}) where (x\textsubscript{min}, y\textsubscript{min}) are the pixel coordinates of the bottom-left bounding box corner, and (x\textsubscript{max}, y\textsubscript{max}) are the pixel coordinates of the top-right corner. An example of the un-annotated captured image and point cloud from the KITTI dataset \cite{DB4} is shown in Figure~\ref{data-sample}, the image shows the front camera view and the generated LiDAR point cloud. 

Some of the bench-marked 2D object detectors for real-time applications from the camera frames are \cite{squeezenet, SqueezeDet, yolov4, SSD2016}. These architectures are based on the approach where the image is processed on filters and layers of the convolution neural network, extracting the feature map of the entire image. The selected object regions are passed onto these extracted feature map, and mapped onto the region feature vector, which on the basis of the class scores predicts the type of object and proposes the bounding box onto it. The selected object regions are passed onto these extracted feature map, and mapped onto the region feature vector, which on the basis of the class scores predicts the type of object and proposes the bounding box.

\subsubsection{3D Object Detection}
\begin{figure}[!ht]
  \centering
  \includegraphics[width=\linewidth]{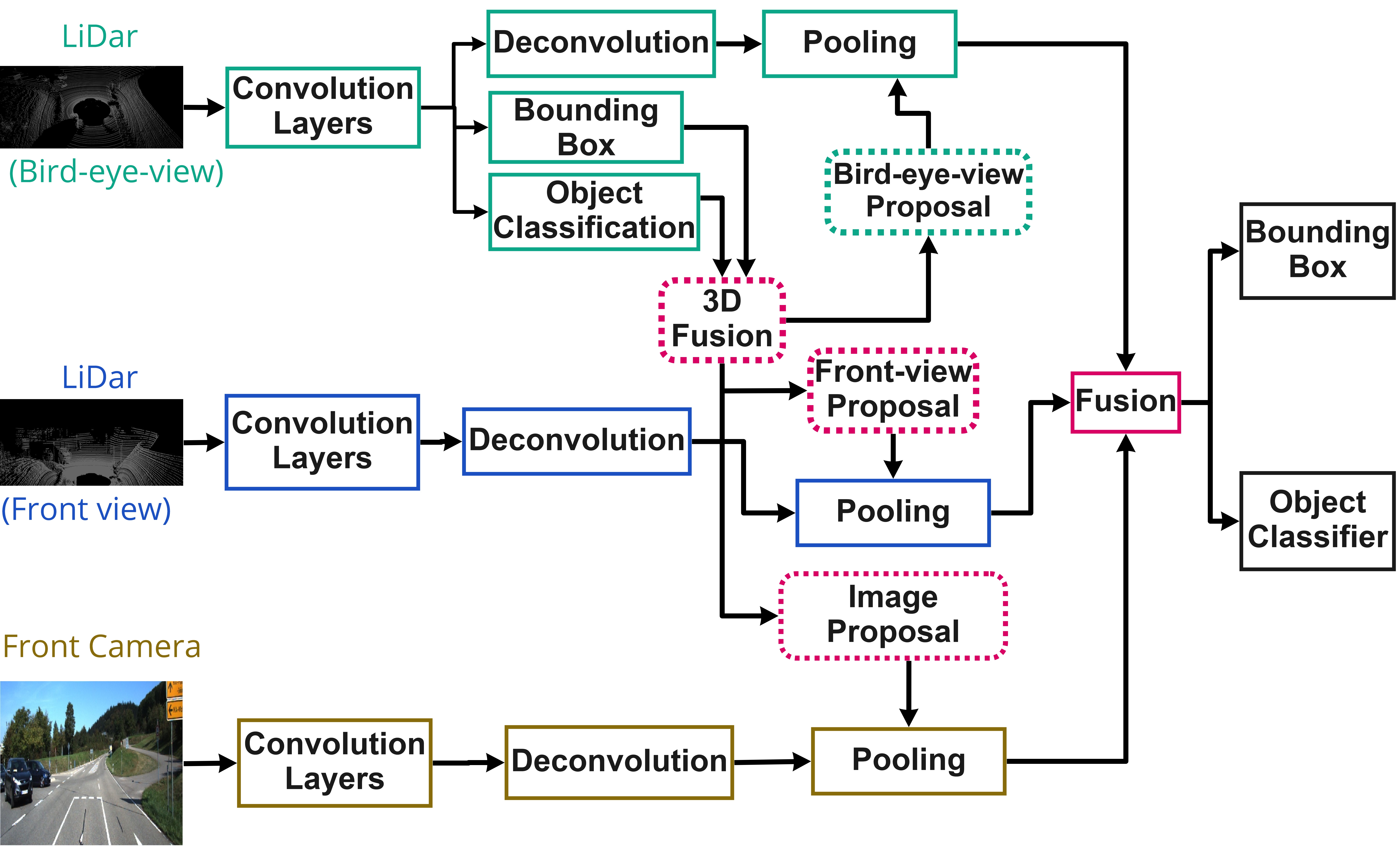}
  \caption{Pipeline for the fusion of feature maps. This approach has been proved essential for LiDAR and Image based 3D-object detection.}
  \label{3d-det-feature-fusion}
\end{figure}

3D object detection is dependent upon sensors such as RGBD camera, 3D radars, LiDAR or combined sensed values, as they can represent the vehicle surrounding in 3D setting. For inference the raw sensed values are processed using the deep learning algorithm, which requires the image with length, width and depth information or the LiDAR point cloud in sparse or dense format as an input. The output from these deep learning algorithm are as follows: At first it detects and classifies the object present in the scene and secondly it predicts a 3D bounding box for the detected objects in the line of sight. In the 3D object detection pipeline, the backbone of the architecture uses neural network with convolutional layers. The convolutional layers are responsible for feature extraction method from the scenes in the local feature map and the global feature map. The next stage comprises of deconvolution layer. The parameters weights obtained after the deconvolution layer are used for two process, in first it is fused together using probabilistic approach to generate and aggregate a score for the detected feature and they secondly they are processed on the pooling layer to fuse them further to obtain the detected object and the predicted bounding box. 3D bounding box can be parameterized as \textit{(x, y, z, l, w, h, $\theta$)}. Here the \textit{(x, y, z)} is the 3D coordinates of the bounding box center, the \textit{(l, w, h)} is length, width and height, respectively of the bounding box, and $\theta$ is the yaw angle of the bounding box. Two different approaches of 3D object detection based on image and LiDAR point clouds is shown in Figure~\ref{3d-det} and Figure~\ref{3d-det-feature-fusion}, where the object detection is used using fusion from the LiDAR point cloud and the respective camera image. 

Most of the statistical or deep learning related algorithms for near real-time 3D object detection and semantic segmentation \cite{katare2019autonomous, Sharkawy2019Naecon} are based on PointNet \cite{qi2017pointnet}, the models proposed here are trained and evaluated on the KITTI dataset, which contains images and LiDAR point clouds collected from the forward facing stereo camera and velodyne LiDAR. Recent point-cloud based architectures such as \cite{Pointpillars2019, MV3D, CenterPoint, geo-cnn, Pv-rcnn}  have made it easier to directly use the raw point cloud for efficient detection on hardware. As reviewed in this section, research in perception category have mainly focused on improving accuracy of the DNN model, multi-object detection and tracking, and implementation on embedded devices, the challenges and opportunities for energy efficient addressed from this sections are: high computational demand, data fusion, collaborative learning models.

\begin{tcolorbox}[title=\textbf{Takeaways}, breakable, skin=enhanced jigsaw]
\begin{enumerate}
    
    \item Computational efficiency: Existing models consist of sequential convolution and fully connected layers with a primary objective of achieving high accuracy on a driving dataset. Deployment of such models is strictly dependent on high-performance devices which increases the onboard, computing and energy costs. Processing such neural networks on a resource-constrained embedded device by maintaining benchmark accuracy remains an open challenge.  

   \item Data fusion: Camera and LiDAR sensors data is used as independent or in combination to detect an object from the vehicular surroundings. However, the current  practices remain to process data on individual pipelines and perform a fusion at the last stage. This leads to excessive use of computation resources for the same operation. 
    
    \item Domain gap: Remarkable progress in object detection can be credited to intelligent algorithms trained on automotive datasets. The sensory technologies used for data collection frequently change in generations (e.g., LiDAR and improvement in resolution). However, little attention has been given to domain adaptation of these algorithms for the next generation of datasets.
\end{enumerate}
\end{tcolorbox}


\begin{table*}[!ht]
\begin{center}
\captionsetup{justification=centering}
\caption{State-of-the-art DNN architectures benchmarked over KITTI and COCO datasets. The table is arranged according to the timeline, data and method used for computation, and on-board inference speed.}
\label{t:detectors}
\begin{tabular}{|c|c|c|c|c|c|c|}
\hline
\textbf{Detection Type}  & Ref  & Year & Data   & Method  & Speed (fps) &Analysis           \\ \hline
\multirow{11}{*}{\textbf{2-D Object}}   
& Faster R-CNN \cite{od1} &2016 &Camera   &2Stage &17 (V100)             &  \multirow{11}{*}{} \\ \cline{2-6}
& SSD \cite{SSD2016}                 & 2016 & Camera   & 1 Stage &22 (Titan X)             &  Dependent on the single or multiple cameras                 \\ \cline{2-6}
& Yolo \cite{od3, yolov4, od5} & 2016 & Camera   & 1 Stage &54 (V100)             & Connected to Sense the Environment                   \\ \cline{2-6}
& SqueezeNet \cite{squeezenet}          & 2017 & Camera   & 2 Stage &17 (Titan X)            &                 \\ \cline{2-6}
& SqueezeDet \cite{SqueezeDet}          & 2017 & Camera   & 2 Stage &30             &  Models are Initially trained on Powerful GPU                  \\ \cline{2-6}
& CornerNet \cite{od8}           & 2018 & Camera   & 2 Stage &33 (Titan X)           & and Later deployed on Embedded device                    \\ \cline{2-6}
& FSAF \cite{od9}                & 2019 & Camera   & 2 Stage &38            &                    \\ \cline{2-6}
& CenterNet \cite{od10}          & 2019 & Camera   & 1 Stage &28 (Titan Xp)            & Real-time inference and SW Acceleration                    \\ \cline{2-6}
& Bottom-up \cite{od11}         & 2019 & Camera   & 1 Stage &             43 (Titan X)& depends on DL frameworks                    \\ \cline{2-6}
& Foveabox \cite{od12}           & 2020 & Camera   & 1 Stage &35 (V100)            &                    \\ \cline{2-6}
& IntPred \cite{od13}           & 2020 & Camera   & 1 Stage &42.8 (GTX 1080)            &                    \\ \hline
\multirow{21}{*}{\textbf{3-D Object}}   
& Baidu \cite{3d1}               & 2016 & LiDAR   & 2 Stage &             & \multirow{21}{*}{} \\ \cline{2-6}
& Vote3deep \cite{3d2}          & 2017 & LiDAR   & 2 Stage &28.6             &                    \\ \cline{2-6}
& MV3D \cite{MV3D}                & 2017 & Ca + Li  & 2 Stage &2.8            & Previous Approach was to transform point                     \\ \cline{2-6}
& PointFusion \cite{3d4}         & 2018 & Ca + Li  & 2 Stage &5             & clouds into images and later use them                   \\ \cline{2-6}
& VoxelNet \cite{voxelnet}            & 2018 & Ca + Li  & 2 Stage &2            & on cnn architecture                     \\ \cline{2-6}
& Deep 3D \cite{3d6}             & 2018 & Ca + Li  & 2 Stage & -            &                    \\ \cline{2-6}
& IPOD \cite{3d7}                & 2018 & Ca + Li  & 2 Stage &37             &                    \\ \cline{2-6}
& PIXOR \cite{3d8}               & 2018 & Ca + Li  & 2 Stage &28.6             & Frustum based approaches improved direct                    \\ \cline{2-6}
& Hdnet \cite{3d9}               & 2018 & Ca + Li  & 2 Stage &20             & use of raw-point cloud on DNN however                       \\ \cline{2-6}
& Frustum PointNets \cite{3d10}  & 2018 & Ca + Li  & Fusion  &2.9             &lacked processing speed for real-time                   \\ \cline{2-6}
& Second \cite{3d11}             & 2018 & Ca + Li & 2 Stage &40             &embedded deployment \& Applications                   \\ \cline{2-6}
& Squeezeseg \cite{3d12}         & 2018 & Ca + Li  & Fusion  &50             &                     \\ \cline{2-6}
& Pointpilllars \cite{Pointpillars2019}      & 2019 & Ca + Li  & 1 Stage &25 (GTX 1080 ti)             &                   \\ \cline{2-6}
& PointRCNN \cite{3d14}          & 2019 & Ca + Li  & 1 Stage &10             & Data Fusion pipelines improved the                   \\ \cline{2-6}
& Lasernet \cite{3d15}        & 2019 & Ca + Li  & 1 Stage &83             &segmentation application on point clouds                    \\ \cline{2-6}
& Class-Balanced \cite{3d16}      & 2019 & Ca + Li  & 1 Stage &42             &                    \\ \cline{2-6}
& Sparse-to-dense \cite{3d17}    & 2019 & Ca + Li  & Fusion  &10             &                    \\ \cline{2-6}
& Mono3d++ \cite{3d18}           & 2019 & Ca + Li  & 1 Stage &20            & Approaches such as machine-learned pillar                     \\ \cline{2-6}
& Pointpainting \cite{3d19}      & 2020 & Ca + Li  & 1 Stage &2.5             & encoders are learned in an end-to-end                    \\ \cline{2-6}
& SA-SSD \cite{3d201}              & 2020 & Ca + Li  & 1 Stage &25             & manner as part of LiDAR 3d object detection                  \\ \cline{2-6}
& 3dSSD \cite{3d20}              & 2020 & Ca + Li  & 1 Stage &25             &  networks heavily relying on labeled               \\ \cline{2-6}
& SE-SSD \cite{3d211}              & 2021 & Ca + Li  & 1 Stage &32             &  training data                   \\ \cline{2-6}
& SPG \cite{3d212}              & 2021 & Ca + Li  & 1 Stage &41.56             &                    \\ \cline{2-6}
& Voxel-Transformer \cite{3d213}              & 2021 & Ca + Li  & 1 Stage &43             &                    \\ \cline{2-6}
& Pyramid-RCNN \cite{3d214}              & 2021 & Ca + Li  & 1 Stage &-             & Grid based methods converts the point-               \\ \cline{2-6} 
& Channel-wise \cite{3d215}              & 2021 & Ca + Li  & 1 Stage &39             & cloud unstructured data to pixel \& voxel                   \\ \cline{2-6}
& Voxel-To-Point \cite{3d216}              & 2021 & Li  & 2 Stage &41             & for 2D and 3D convolution processing                    \\ \cline{2-6}
& Voxel-RCNN \cite{3d217}              & 2021 &Ca + Li  & 1 Stage &40.8             &                    \\ \cline{2-6} 
& Multi-View to H-3D \cite{3d218}              & 2021 & Ca + Li  & 1 Stage &             & Recent Approaches involves using encoders                    \\ \cline{2-6}   
& SA-Det3D \cite{3d219}              & 2021 & Ca + Li  & 1 Stage &36             & for detection refinement of far and distant                   \\ \cline{2-6}
& X-View \cite{X-view}              & 2021 & Ca + Li  & 1 Stage &47             & objects, these decoders enhances the point                    \\ \cline{2-6}
& CenterPoint \cite{CenterPoint}        & 2021 & Ca + Li  & Fusion  &16             & feature through hierarchical aggregation.                   \\ \hline
\multirow{5}{*}{\textbf{Lane}} 
& Vpgnet \cite{3d22}            & 2017 & Ca       & 2 Stage &20             & \multirow{5}{*}{}  \\ \cline{2-6}
& LaneNet \cite{3d23}              & 2018 & Ca       & 2 Stage &50             & Most DNN model uses RGB Images for input                    \\ \cline{2-6}
& E2E Lane Det \cite{3d24}        & 2018 & Ca   & 1 Stage &-             & which is challenging in real-world situation                    \\ \cline{2-6}
& Spatial as Deep \cite{3d25}      & 2019 & Ca + Li  & 1 Stage &-             & as per changed weather \& Light Condition                    \\ \cline{2-6}
& 3DLaneNet \cite{3d26}           & 2019 & Ca + Li  & 1 Stage &53   &   \\ \cline{2-6}
& Gen-LaneNet \cite{3d27}           & 2020 & Ca + Li  & 1 Stage &60             &3D lane detection improves constraints such                 \\ \cline{2-6}
& Real-time Lane-det \cite{3d28}    & 2021 & Ca  & 1 Stage &48             &as making turns or merging to another lane               \\ \cline{2-6}
& Low-light Lane \cite{3d29}           & 2021 & Ca   & 1 Stage &-            & with inclusion of sensors: radar, LiDAR                \\\hline
\end{tabular}
\end{center}
\end{table*}

\subsection{HD Map}
High-definition map in an autonomous vehicle can provide dynamic and static conditions, such as semantic information, topology, and geometric information, from the vehicle surrounding using cameras and LiDAR sensors \cite{HD4}. One of the key requirements in autonomous driving is to accurately localize itself with respect to its surroundings and the infrastructure, and gathered information from an HD map can be used to support this function including vehicle motion control, motion planning, and perception \cite{HD16, HD18}. Therefore, maps are essential components for level 4 and beyond autonomous driving. Previously maps were used as a driver assistance feature \cite{HD2} to guide in navigation from source to destination. Google and Apple were the first of the few organization to collect street, city, and highway data which later enabled the flexible transportation and mobility by using GPS devices or map based applications on the regular smartphones. With the advancement in technology and algorithms the 3D maps of cities such as New York, Washington were created. HD maps for autonomous driving is the result of advancements in sensor and driving use-cases \cite{HD2, HD4}. 

\begin{figure}[!ht]
\centering
\includegraphics[width=\linewidth]{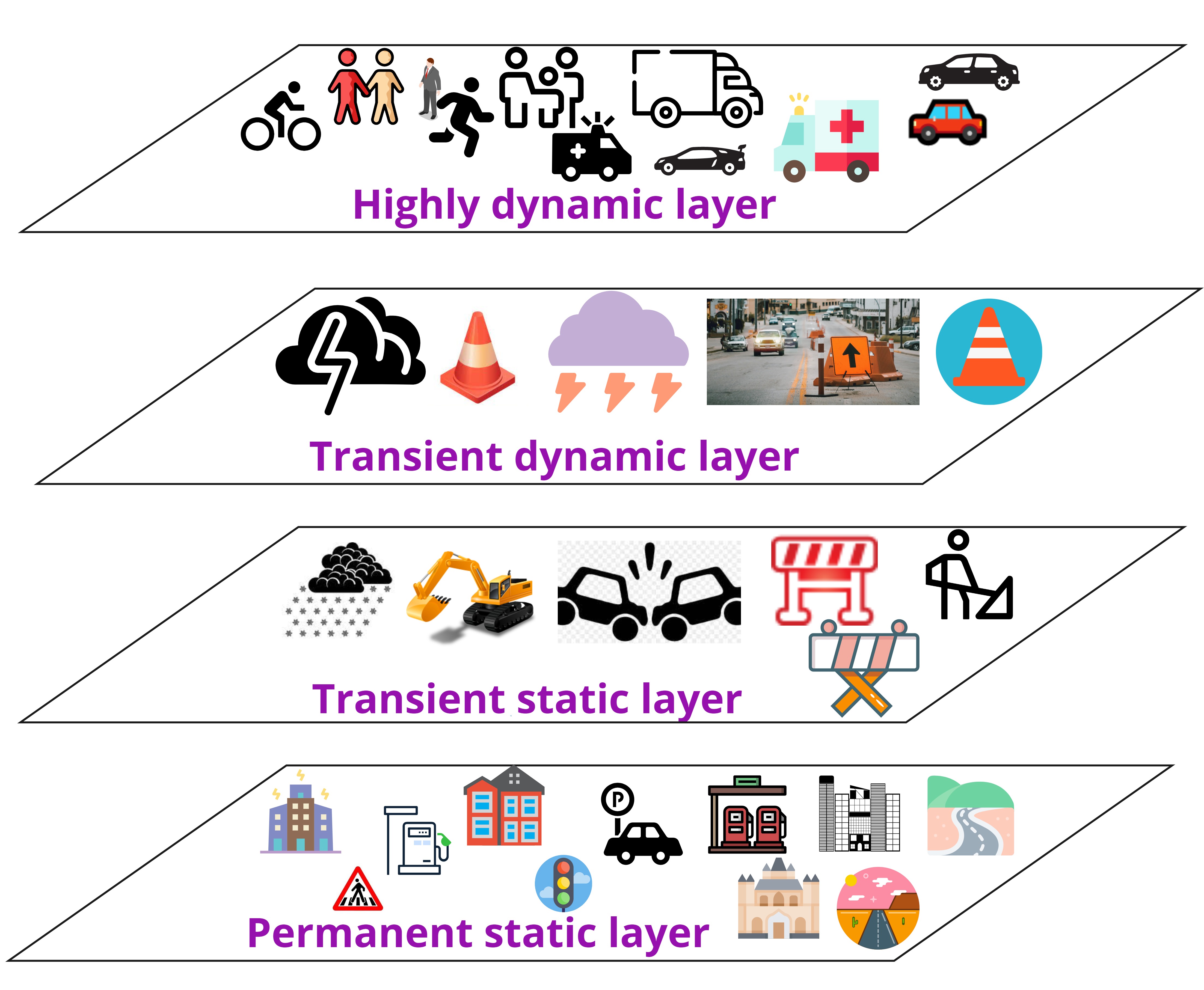}
\caption{HD-map layers representation in ecosystem}
\label{fig-hd}
\end{figure}

Current HD maps lack specifications about the data type or standard guidelines, such as annotated information that should be stored while creating them. The automotive edge computing consortium (AECC) has proposed a version of an HD map consisting of four layers. This map version is based on Local Dynamic Map initially proposed by the European Telecommunications Standards Institute (ETSI) \cite{buvcko2021smart}. The layer includes two static and dynamic layers, which are further classified based on timelines and changes expected within the vehicular ecosystem (Figure~\ref{fig-hd}). Current use-cases, includes creating an HD map from the raw sensor data and updating an existing map using crowdsourced data from the vehicles and infrastructure sensors in the vehicle-edge-cloud setting. The four layers proposed in the AECC version are as follows:

\begin{itemize}
    \item Permanent static layer serves as the foundation by providing a static map of the surroundings. This layer consists of road maps, buildings, and roadside infrastructures. This layer consists of map data and information that does not change frequently.
    \item Transient static layer contains information about scenarios that may be subject to change over a few days to a few hours. As shown in the figure~\ref{fig-hd} this layer may contain information on the change to static layer for, e.g., snowfall, road construction, maintenance and accidents.
    \item Transient dynamic layer contains information on surrounding that frequently changes. Here, change can occur in a few minutes and last a few hours. It may contain information on road obstacles, heavy rainfall and storms.
    \item Highly dynamic layer frequently changes; in a few seconds to a few minutes. Thus contains information about moving objects such as other vehicles, pedestrians and motorcyclists. This section has not included information requiring frequent updates that may be less than a second interval in an HD map.

\end{itemize}

Relevant work in HD maps in using deep neural networks includes: Hdnet, Vectornet, Exploiting sparse semantic HD maps \cite{HD4, HD16, HD17, HD18}. Machine learning based approach and workflow for creation of high definition semantic map is presented in \cite{HD4}. In this paper author discussed the steps from data capture using sensors, annotations, and map generation. Use-case such as pose estimation, traffic sign and line mapping, lane/road marking were also discussed. In the similar context a complete HD map framework for autonomous driving is presented in \cite{HD18}. The authors comprehensively presented the HD map application by describing the pre-built maps, storage in cloud, locally built maps and update in the global map based on change in static semantic conditions. In this paper, the framework is distributed into on-vehicle mapping, user-end localization, and on-cloud mapping. 

For on-vehicle mapping traditional semantic method, pose estimation, perspective transformation and local mapping have been used \cite{HD18, CNN-SLAM, ORB-SLAM}. On-cloud mapping is responsible to merge and aggregate map data from multiple vehicles. Functions are used to merge local data timely such that the global map is up-to-date. As the size of data and volume is not fixed, a function to compress the map data is also implemented at the On-cloud mapping. Lastly, the user-end localization are vehicles requesting map information from the cloud. When the vehicle receives the map, an algorithm to decompress map data is implemented and data is further processed through a semantic localization pipeline.

Researchers have also predicted that around 10{\%} of the roads or static conditions changes every year because of the construction and related scenarios. Therefore, crowd-sourcing based HD map update have been proposed to update the global map using individual vehicles \cite{HD6, HD7, HD15, HD17, HD12}. In \cite{HD15}, authors proposed to use sensors, such as GNSS, IMU and camera, to detect the change in the HD map using BiseNet architecture as semantic baseline and visual SLAM for localization and mapping. For experiment authors used arrow sign as an example from the surrounding and by using vectorization and matching approach detected the change in existing map data. Similar approach to update HD map using edge-servers is proposed in \cite{HD17}. In this paper authors discussed the issue of diminishing marginal utility and premature convergence of map data from individual vehicles. To this end, task distribution mechanism which uses adaptive time period division mechanism is proposed. In the experiments using edge devices and computing unit the effectiveness is verifies using coverage, cost and efficiency. 

A crowd-sourcing based approach to create HD map using graph-SLAM \cite{surfel-based} is proposed in \cite{HD6}. The authors used GNSS, odometry, point cloud data, and land marking to be processed using a graph-SLAM algorithm. The authors used pose estimation, smoothing filter, trajectory alignment for the landmarks. Road model inference and lane geometry is used to create the functions for lane boundary lines, connections and point observations. To evaluate the approach, an experiment with the ground-truth data was implemented. Deep learning methods using crowd-source based HD map update is proposed in \cite{HD7, HD12}. In \cite{HD7}, authors proposed a change detection algorithm using boosted particle filter. The particle filters are applied during the localization along with a classification algorithm. In \cite{HD12}, authors proposed a framework that maps the sensed image/frame from camera to probabilities of HD map change. As the HD map data consist of geometric information and lane marking, deep learning metric is used to reduce the domain gap. In experiments authors implemented object detector with a pixel-level change detection from the input/sensed image, evaluated on city-scale dataset.

Other interesting techniques that can be explored for HD map creation and development are neural radiance field \cite{NeRF, block-nerf}, and mean-field game \cite{huang2019game, huang2021dynamic}. Instead of using three coordinate system (x, y, z), in neural radiance field \cite{NeRF} a five coordinate system including (x, y, z, $\alpha$, $\phi$) are used, where the last two are viewing direction. Authors used fully connected neural network to generate 3D scenes and frames based on the trained 2D images. For comparative study, performing techniques such as neural volumes, scene representation networks and local light field fusion is used to directly predict a multi-plane image for the input. The approach is very useful for 3d models of object captured from camera. Similar approach is proposed in block-nerf\cite{block-nerf} to represent surrounding in large scale view. In \cite{block-nerf}, architecture layers are modified using pose refinement, generative latent optimization, to adapt image appearance embedding as different images could be captured in different environment conditions. For experiments and evaluation, authors reconstructed 3D scenes using 2.8 million images captured from camera. Interesting work using mean-field game is proposed in \cite{huang2019game, huang2021dynamic}. In \cite{huang2019game}, authors proposed a computational framework by categorizing the scenario into microscopic and macroscopic perspective to control velocity for vehicles, and further develop traffic flow for autonomous vehicles. A comprehensive study is presented to characterize equilibrium solutions in both continuous MFGs and discrete differential games, a similar approach can be implemented in HD map creation and update, which requires strategic interaction between connected autonomous vehicles.

The challenges and opportunities in energy efficient approaches with HD map applications are as follows:
\begin{enumerate}
    \item Data collection and Processing: An hour of driving approximately corresponds to 1.5TB data from a car. Processing and interpretation of collected data requires efficient algorithms and high-end computational resources.

    \item Map storage and sharing: One of the primary challenge is the design of common energy-efficient framework for edge servers which can store and share the HD map to the autonomous vehicle through local wireless (802.11p),  cellular or hybrid communication approach.
    
    \item HD Map update: Approximately 10-15\% of surrounding or street scenes are expected to change because of the development in infrastructure. Therefore an energy-efficient approach and scheme to update the existing HD map, rather updating the database in periodic manner. 

    \item Intelligent driving: The amount of information perceived by sensors in city and highway driving is different, intelligent algorithms developed for Edge server assisted HD map update can help  to identify the sensory information needed to map and update.
\end{enumerate}

\begin{tcolorbox}[title=\textbf{Takeaways}, breakable, skin=enhanced jigsaw]
HD map is essential and an emerging technique in autonomous driving. Present HD maps are available from the semantic and geometric perspective. HD maps can be created locally every-time using vehicular computing unit, but this tends to be compute intensive. NRF, MFG and deep learning techniques can be explored for data generation, map creation and global HD map update. Crowd-sourced map update is promising approach, however data merge, schedule and aggregation approaches should be regularly optimized.
\end{tcolorbox}

\subsection{SLAM}
Simultaneous Localization and Mapping often abbreviated as SLAM has been widely researched in robotics, and autonomous systems, including indoor applications focusing on warehouses and manufacturing units. In an autonomous vehicle, SLAM is a process utilizing algorithms to estimate the real-time position of the vehicle by continuously perceiving and sensing the environment using embodied sensors. The goal of using SLAM is to create a virtual environment for the vehicle by identifying the obstacles, and infrastructure, thus assisting in creating a path for safe navigation. 

In \cite{Autoware, apolloscape}, authors have proposed maps \cite{HD7, HD8, HD9}, also referred to as 3D maps, in combination with SLAM for efficient and precise localization. SLAM techniques are mostly dependent on algorithmic approaches such as probabilistic roadmap (PRM), rapidly-exploring random graph (RRG), rapidly-exploring random tree (RRT), and parti-game directed RRTs (PDRRTs). These algorithms are designed to accurately search the subset of euclidean space over the high-dimensional geometry by randomly building a space-filling tree (RRT). SLAM application demands low latency (5ms or less) and high computational resources, thus consuming a significant amount of energy from on-board computing unit. Recent SLAM approaches have been proposed without the use of a Global Positioning System (GPS), and can be separated into two categories: Filter-based techniques and Optimization-based techniques. The filter-based category is primarily built on the Bayes theorem, thus utilizing Probabilistic estimation using Bayesian filters. 

Some of the commonly used approaches are: Kalman Filter, Extended Kalman Filter (EKF), Unscented Kalman Filter (UKF). In the same category other used techniques are particle-filters such as FastSLAM, Rao-Blackwellized Particle filters and Monte Carlo filters, commonly practised as learning algorithms for dynamic Bayesian networks. Table~\ref{t:slam} shows a list of popular slam approaches that are based on line of sight sensors, radar, and their fusion. Recently visual or 3D SLAM approaches have been a popular method to localize the vehicle within the environment. The table categorizes the type of SLAM techniques such as 2D SLAM (Camera) or 3D SLAM (RGBD camera and LiDAR). Depending on the input data, a grid, voxel, or point cloud map is used for projection or visualization of SLAM methods. The Optimization-based category for SLAM is primarily based on Graph SLAM, which is also motivated by the Bayesian theorem and is primarily a graphical representation of it by utilizing the matrix form and thus relating the state of the vehicle within the environment. The matrix consists of values or information related to vehicle pose, which can be used to solve the localization problem. 

The techniques utilizing Graph SLAM are: Oriented fast and Rotated Briefs-SLAM (ORB SLAM), Large-Scale Direct Monocular SLAM (LSD-SLAM). Other commonly used techniques are based on deep learning practices such as: CNN-SLAM, DeepFusion, Deepfactors, Structured-SLAM, DRM-SLAM. These practices are promising bases on their evaluation and performance on driving datasets such as KITTI, however, they still pose a challenge based on efficient and faster computation scenarios required in non-identical practical driving situations. 

Compared to SLAM approaches involving point clouds, visual SLAM is a more preferred approach in terms of cost which uses significantly less expensive cameras compared to LiDARs. However, visual SLAM may not be precise and as accurate as point clouds based SLAM approaches, but it is significantly faster on standard computing devices \cite{DV-Loam}. Another disadvantage of visual SLAM is being sensitive to the changes in the scenes, illumination and appearance. The accuracy and precision of proposed SLAM approaches could perform differently in dynamic or bad weather conditions. In terms of advantage, visual SLAM has better graphic coverage than point-clouds unless multiple LiDAR are used. Deployment of SLAM in Edge AI environment bring several challenges and opportunities, key points can be highlighted as:

\begin{tcolorbox}[title=\textbf{Takeaways}, breakable, skin=enhanced jigsaw]
\begin{enumerate}
    \item Computation: In general the SLAM application demands high computation cost for smaller maps, several problems with respect to processing and accuracy can be encountered with respect to non-ideal conditions and size of data captured for processing. At present powerful GPU devices are required for processing, which brings the overall cost of vehicles high.

    \item Latency time: For real-time execution, latency must be lower than 5 ms if incorporated using Edge or Cloud Computing.
    
    \item Algorithm: DNN approaches used for SLAM makes it suitable to operate in familiar environment. However, change in location, weather and daylight conditions can bring additional complexities as the sensed output will be inconsistent and DNN model will not be able to process it.
    
    \item Execution: Future connected vehicles are expected to execute services in distributed manner (at the vehicle, edge-server or cloud). With the current DNN algorithms, computational, latency and network bandwidth requirement, it is more realistic to process and execute SLAM at the vehicles on-board computing unit.
\end{enumerate}
\end{tcolorbox}

\begin{table*}[!ht]
\captionsetup{justification=centering}
\caption{The table shows deep learning models proposed for vehicular SLAM application. It also includes approaches proposed within the indoor environment, which are scalable for the outdoor scenes.}
\label{t:slam}
\begin{center}
\begin{tabular}{|c|c|c|c|c|c|c|c|}
\hline
\multicolumn{8}{|c|}{\textbf{Comparison of SLAM techniques for Autonomous Driving Services}}                                                                                                                                           \\ \hline
\textbf{Reference}      & \textbf{Type}             & \textbf{Method} & \textbf{Projection} & \textbf{Localization} & \textbf{Real-time} & \textbf{\begin{tabular}[c]{@{}c@{}}Compute \\ Power\end{tabular}} & \textbf{Environment} \\ \hline
Real-time Loop \cite{real-timeloop}        & 2D SLAM       & EKF             & Grid Map            & Good                  & Yes                & Low                                                               & Indoor                \\ \hline
Duality-based \cite{duality}         & 2D SLAM       & Graph           & Grid Map            & Medium                & Yes                & Medium                                                            & Indoor                \\ \hline
Particle Grid-mapping\cite{particle-grid} & 3D SLAM       & Particle        & Grid Map            & Good                  & No                 & -                                                                 & Outdoor               \\ \hline
Tiny SLAM \cite{TinySLAM}             & 3D SLAM       & Particle        & Point Cloud Map     & Good                  & Yes                & Low                                                               & Indoor                \\ \hline
Rotating 3D SLAM\cite{fang2017real}      & 3D SLAM       & Particle        & Point Cloud Map     & Good                  & Yes                & High                                                              & Indoor                \\ \hline
Surfel-Based \cite{surfel-based}          & 3D SLAM       & Graph           & Point Cloud Map     & Medium                & Yes                & High                                                              & -                     \\ \hline
CPFG-SLAM \cite{CPFG-SLAM}           & 2D SLAM       & Probabilistic   & Grid Map            & Good                  & Yes                & High                                                              & Indoor                \\ \hline
IMLS-SLAM \cite{IMLS-SLAM}           & 3D SLAM       & Least-Square    & Point cloud         & Excellent             & No                 & Low                                                               & -                     \\ \hline
MC2-SLAM \cite{Elastic}             & 3D SLAM       & Scan-Map        & Point Cloud map     & Medium                & Yes                & High                                                              & -                     \\ \hline
LIMO \cite{LIMO}                  & 3D SLAM       & Probabilistic   & Point Cloud Map     & Good                  & Yes                & -                                                                 & -                     \\ \hline
STEAM-L \cite{STEAM-L}               & 3D SLAM       & Scan-Map        & Point Cloud Map     & Medium                & Yes                & -                                                                 & -                     \\ \hline
M3RSM \cite{M3RSM}                 & 3D SLAM       & Scan-Scan        & Point Cloud         & Good                  & Yes                & Low                                                               & Indoor + Outdoor      \\ \hline
LOAM \cite{LOAM}                & 3D SLAM       & Particle        & Point Cloud         & Excellent             & -                  & Low                                                               & Indoor                \\ \hline
V-LOAM \cite{V-LOAM}               & 3D SLAM       & Particle        & Point Cloud         & Good                  & Yes                & Low                                                               & Indoor + Outdoor      \\ \hline
ORB-SLAM \cite{ORB-SLAM}             & 3D SLAM       & Graph           & Point Cloud         & Excellent             & Yes                & High                                                              & Indoor + Outdoor      \\ \hline
Deepfactors \cite{Deepfactors}           & 3D SLAM       & Probabilistic   & Depth Map           & Good                  & Yes                & High                                                              & Indoor                \\ \hline
CodeSLAM \cite{CodeSLAM}              & 2D SLAM       & Keyframe        & Map                 & Good                  & Yes                & Low                                                               & Indoor                \\ \hline
Structured-SLAM \cite{Structured-SLAM}      & 2D SLAM       & Graph           & Plane Segmentation  & Good                  & Yes                & High                                                              & Indoor                \\ \hline
CNN-SLAM \cite{CNN-SLAM}              & 3D SLAM       & Graph           & Semantic            & Excellent             & Yes                & High                                                              & Indoor                \\ \hline
LOAM Livox \cite{LOAM-Livox}            & 3D SLAM       & Graph           & Point Cloud         & Good                  & Yes                & High                                                              & Outdoor                \\ \hline
F-LOAM \cite{F-Loam}               & 3D SLAM       & Map-matching    & Voxel               & Excellent             & Yes                & Low                                                               & Indoor + Outdoor      \\ \hline
DV-Loam \cite{DV-Loam}              & 3D SLAM       & Frame-Frame     & Point Cloud         & Excellent             & Yes                & High                                                              & Outdoor               \\ \hline
\end{tabular}
\end{center}
\end{table*}

\begin{table*}[!ht]
\begin{center}
\caption{Long Range Communication Technologies for Autonomous Driving}
\label{t:longshortrange}
\begin{tabular}{@{}|cccccccc|@{}}
\toprule
\multicolumn{8}{|c|}{\textbf{Long Range Communication Technologies}}                                                                                                                                                                                                                                                      \\ \midrule
\multicolumn{1}{|c|}{\textbf{Technology}} & \multicolumn{1}{c|}{\textbf{Standard}} & \multicolumn{1}{c|}{\textbf{Spectrum}} & \multicolumn{1}{c|}{\textbf{Range}} & \multicolumn{1}{c|}{\textbf{Modulation}} & \multicolumn{1}{c|}{\textbf{Latency (ms)}} & \multicolumn{1}{c|}{\textbf{Security}} & \textbf{Field Trial} \\ \midrule
\multicolumn{1}{|c|}{DSRC}                & \multicolumn{1}{c|}{802.11p}           & \multicolumn{1}{c|}{5.8 - 5.9 GHz}     & \multicolumn{1}{c|}{1 Km}           & \multicolumn{1}{c|}{OFDM}                & \multicolumn{1}{c|}{100}                   & \multicolumn{1}{c|}{B}                 & Yes                  \\ \midrule
\multicolumn{1}{|c|}{C-V2X}               & \multicolumn{1}{c|}{3GPP}              & \multicolumn{1}{c|}{800/1800 MHz}      & \multicolumn{1}{c|}{5 Km}           & \multicolumn{1}{c|}{SC-FDMA}             & \multicolumn{1}{c|}{10}                    & \multicolumn{1}{c|}{B}                 & Yes                  \\ \midrule
\multicolumn{1}{|c|}{WiMax}               & \multicolumn{1}{c|}{802.16}            & \multicolumn{1}{c|}{2.5 GHz}           & \multicolumn{1}{c|}{50 Km}          & \multicolumn{1}{c|}{MIMO, OFDM}          & \multicolumn{1}{c|}{10}                    & \multicolumn{1}{c|}{B}                 & Yes                  \\ \midrule
\multicolumn{1}{|c|}{5G NR V2X}           & \multicolumn{1}{c|}{3GPP}              & \multicolumn{1}{c|}{24 - 86 GHz}       & \multicolumn{1}{c|}{5 Km}           & \multicolumn{1}{c|}{OFDM}                & \multicolumn{1}{c|}{1}                     & \multicolumn{1}{c|}{A}                 & Yes                  \\ \midrule
\multicolumn{8}{|c|}{\textbf{Short Range Communication within Vehicles}}                                                                                                                                                                                                                                                  \\ \midrule
\multicolumn{1}{|c|}{\textbf{Technology}} & \multicolumn{1}{c|}{\textbf{Standard}} & \multicolumn{1}{c|}{\textbf{Spectrum}} & \multicolumn{1}{c|}{\textbf{Range}} & \multicolumn{1}{c|}{\textbf{Modulation}} & \multicolumn{1}{c|}{\textbf{Latency (ms)}} & \multicolumn{1}{c|}{\textbf{Security}} & \textbf{Bit rate}    \\ \midrule
\multicolumn{1}{|c|}{WiFi}                & \multicolumn{1}{c|}{802.11 ac}         & \multicolumn{1}{c|}{5 GHz}             & \multicolumn{1}{c|}{100 m}          & \multicolumn{1}{c|}{1, 2, 3, 5, 7}       & \multicolumn{1}{c|}{NA}                    & \multicolumn{1}{c|}{24-bit CRC}        & 1 Gb/s               \\ \midrule
\multicolumn{1}{|c|}{BLE}                 & \multicolumn{1}{c|}{802.15.1}          & \multicolumn{1}{c|}{2.4 GHz}           & \multicolumn{1}{c|}{30 - 50 m}      & \multicolumn{1}{c|}{4}                   & \multicolumn{1}{c|}{4 - 6}                 & \multicolumn{1}{c|}{24-bit CRC}        & 1 - 24 Mb/s          \\ \midrule
\multicolumn{1}{|c|}{ZigBee}              & \multicolumn{1}{c|}{802.15.4}          & \multicolumn{1}{c|}{2.4 GHz}           & \multicolumn{1}{c|}{75 - 100 m}     & \multicolumn{1}{c|}{1, 6}                & \multicolumn{1}{c|}{30}                    & \multicolumn{1}{c|}{16-bit CRC}        & 20 - 250 Kb/s        \\ \midrule
\multicolumn{1}{|c|}{UWB}                 & \multicolumn{1}{c|}{802.15.3}          & \multicolumn{1}{c|}{3.1 - 10.6 GHz}    & \multicolumn{1}{c|}{75 m}           & \multicolumn{1}{c|}{1, 7}                & \multicolumn{1}{c|}{NA}                    & \multicolumn{1}{c|}{32-bit CRC}        & 10 Mb/s              \\ \midrule
\multicolumn{8}{|c|}{Modulation Type (Short Range Communication) - ``BPSK = 1, CCK = 2, COFDM=3, GFSK = 4, M-QAM = 5, O-QPSK = 6, QPSK = 7"}                                                                                                                                                                                                           \\ \bottomrule
\end{tabular}
\end{center}
\end{table*}

\subsection{Vehicular Communication}
Communication within vehicular environment plays a key role in self-driving functionality \cite{vc1}. V2X or vehicle to everything communication is another key factor in the self-driving vehicle ecosystem that allows and enables the communication between vehicles to any relevant entity in the environment for example pedestrians, traffic lights, data centres. V2X comprises of several sub-components and standards such as V2V (Vehicle to Vehicle Communication), V2I (Vehicle to infrastructure), V2P (Vehicle to Pedestrian), V2N (Vehicle to Network), and V2G (Vehicle to Grid) has also been included considering the electric vehicles, charging stations and their involvement in the infrastructure. The Ideal system in V2X communications comprises of pair of radio devices often called as On-Board units (OBU), and Road-side units (RSU). OBU's are placed in the car, sharing car-related information to the RSU and receiving the traffic or surrounding related information from it. Some of the popular modules include \cite{commsignia2017, cohda2019} which has already been released in the past 4 years. Also hybrid communication approaches combined with cellular technology (CV2X) \cite{vc5}, Dedicated Short-range communication modules (DSRC) \cite{vc2, vc3, vc4}, also with the LTE based systems and 5G  \cite {vc14, vc15, vc17, vc18, vc21} has been proposed. In \cite{ismael2021reliable} authors explored reliable connected-vehicle services using wireless local area network, ad-hoc network or hybrid communication architectures using cellular connectivity. To estimate the time duration for connection establishment probabilistic model implementing single-hop communication link in vehicular networks \cite{katsaros2017conceptual} is explored. To further ensure the reliability of communication in vehicular ecosystem a reliable emergency message dissemination scheme (REMD) \cite{benrhaiem2019reliable}, has been presented by authors. Results from REMD scheme shows high reliability which is around 99{\%} in each hop with low overhead, delivering the message for time-critical applications meeting the low-latency requirements for sensitive applications. The authors also employ the zero-correlated unipolar orthogonal codes to combat the hidden terminal problem. In the approach the periodic beacons are exploited, to precisely estimate the reception quality of 802.11p wireless link in each cell; then, uses this information to determine the optimal number of broadcast repetitions in each hop. In addition, to ensure reliability in multi-hop, cooperative communication within the network is also enabled, The simulation results show that REMD outperforms the existing well-known schemes for reliable communication.

The initial vehicular communication was developed considering the local wireless networks such as dedicated short-range communication or Wi-Fi (802.11p) which is an updated version of 802.11b to enable wireless access in a vehicular environment. However based on the scalability some other versions such as C-V2X\cite{vc5} were proposed which operates in both the 5.9GHz spectrum and also in the cellular spectrum thus providing channels for long-range communication between vehicles and the surroundings, Table~\ref{t:longshortrange} shows some of the popular long-range communication technologies. The solutions consisting of proposed combinations can provide low-latency, high reliability and throughput demand \cite{vc3}. Also to overcome these challenges another approach such as next-generation V2X (NG V2X) or New radio technology (NR V2X) \cite{vc21} has been proposed, as per the results, these approaches overcome the challenges and have better network performance and parameters. Key communication technologies proposed for vehicular communication are discussed below.

\textbf{DSRC:} One of the initial technology proposed for medium-range vehicular communication is dedicated short-range communication (DSRC). This technology can be used in autonomous vehicles to deploy applications within a transmission range of 25-100 meters. It is a sub-protocol within vehicle-to-everything (V2X) that can enable communication between vehicle-to-vehicle (V2V). V2V supports automated message propagation and exchange of vehicle information (e.g., velocity, acceleration, separation distance, the direction of travel) with nearby vehicles. The purpose of exchanging these messages and vehicle information is to improve traffic conditions and to implement safety applications, such as collision avoidance and safe overtaking \cite{CEAD15}. With the increase in message transmission capability, recently proposed methods also include cooperative perception using V2V communication \cite{CEAD6, CEAD17}. Potential driving and safety-critical applications developed and tested with DSRC are collision warning systems and emergency braking \cite{vc2, vc3, vc7}. However, with the evolution of next-generation vehicular communication technologies and use-cases requiring high-volume data transmission, the technology has not been widely adopted by automotive manufacturers and communication providers \cite{FCC19-129}.

\textbf{C-V2X:} Cellular-V2X is based on the sidelink LTE radio interface enabling point-to-point communication with nearby vehicles and devices. As described in 3GPP, C-V2X generally operates in two channels i.e., 10 MHz or 20 MHz, and includes LTE-V2X and 5G-V2X \cite{vc5}. C-V2X utilizes a time-frequency resource structure, where the time is divided into 1ms sub-frames, and the frequency channel is divided into 180 kHz wide resource blocks. These resource block exists in the same sub-frame and can be further clustered into sub-channels \cite{vc14}. Resource allocation schemes and optimization techniques were proposed in \cite{vc14, vc5} to improve network latency performance. Network performance measurements and scenario-in-loop field-testing method for 5G-V2X were presented in \cite{vc17}, where applications for testing involved braking, obstacle detection, and tracking. A shortcoming in C-V2X technology, in comparison to DSRC is that the vehicles cannot process and exchange messages directly, as it is dependent on the LTE. Another flaw in the current approach is the inability to work in remote or geo-locations with poor cellular/network coverage.

\textbf{NR V2X:} New Radio (NR) V2X is designed to complement the applications that are not fully supported in C-V2X because of varied latency, bandwidth and throughput requirements \cite{vc21}. NR V2X use-cases comprises of efficient and reliable delivery of aperiodic messages, which was not very well supported in C-V2X \cite{vc22, vc23}. As compared to V2X, NR V2X also supports groupcast and broadcast transmission methods which are specifically required for applications such as vehicle platooning \cite{vc21, vc23, vc26}. The development in this category will bring several opportunities for urban and highway driving services, such as platooning, predictive planning, and real-time edge analytics involving traffic flow management and forecasting. Several challenges exist in vehicular communication in terms of latency, privacy, and reliability.

\begin{tcolorbox}[title=\textbf{Lessons Learned}, breakable, skin=enhanced jigsaw]
\begin{enumerate}
    \item Latency: In an urban driving scenario, multiple vehicles could be in the same location and will be communicating with the local edge server. This situation brings a challenge for real-time low latency applications such as SLAM, which requires transmission of huge data from vehicle sensor to edge server and vice versa.
    
    \item Privacy: In vehicular communication, some sensitive information such as vehicle registration number, vehicle health, real-time status along with sensors data, and statistic models is shared. Sharing this information exposes a threat of data poisoning, model weights manipulation and adversarial attack on the system.

    \item Collaborative application: As mentioned a local edge server will be communicating with multiple vehicles, and the vehicle is also communicating with a peer vehicle for the applications implementing collaborative driving. The collaborative driving applications require data aggregation methods and processing practices at the edge server to combine similar data from multiple sensors sources and have a common prediction.
\end{enumerate}
\end{tcolorbox}

\subsection{Energy Efficient Approaches in Autonomous Driving}
Autonomous systems such as robots, unmanned aerial vehicle are mostly powered by fixed battery source. The same assumption can be made for the future vehicles depending upon the availability of fuels and planning of the future sustainable transportation systems. For the current deployed autonomous vehicle, It is important to consider the energy required and used by sensors, automotive embedded processors and embedded devices, such as GPU, TPU and CPU while sensing the surrounding data and processing of algorithms. The energy consumed from the processor and devices can be derived by sampling the power consumption at the training of deep neural network model or architecture \cite{garcia2019estimation}. Another brute force method could be to use power measurement devices with the embedded devices during the inference, and log the power consumption over the processing of algorithm. However these approaches are not very much effective as the autonomous driving ecosystem consists of heterogeneous types of devices, in which some might not be equipped with TPU or GPU, therefore it is important to consider a neutral method to calculate the power usage, in which power consumption from each of these devices or nodes is categorically calculated \cite{seewald2021coarse} based on the type of processor. To further estimate the total power consumption for heterogeneous devices in distributed learning setting, summation of the total training time on each of them can be used. However this approach might not work for the federated learning implementation, as the training time between participating devices can significantly vary and the fundamental of federated learning is based on the communication rounds between the devices and the ultimate convergence rate. 

Based on the approach such as resource or computational ability, only certain available devices are chosen for training during each communication round, as based on the specification the participating devices, they might not offer the equal computational capability\cite{CEAD14}. Also another factor in case of distributed training is the total time needed to train the model as it highly depends upon the communication efficiency between the participating devices and the server. It is important to note that in addition to the on-board energy consumption, these approaches also brings into account the energy consumption caused due to communication between devices, network stations and server \cite{krail2019energie}.

\begin{figure}[!ht]
  \centering
  \includegraphics[width=\linewidth]{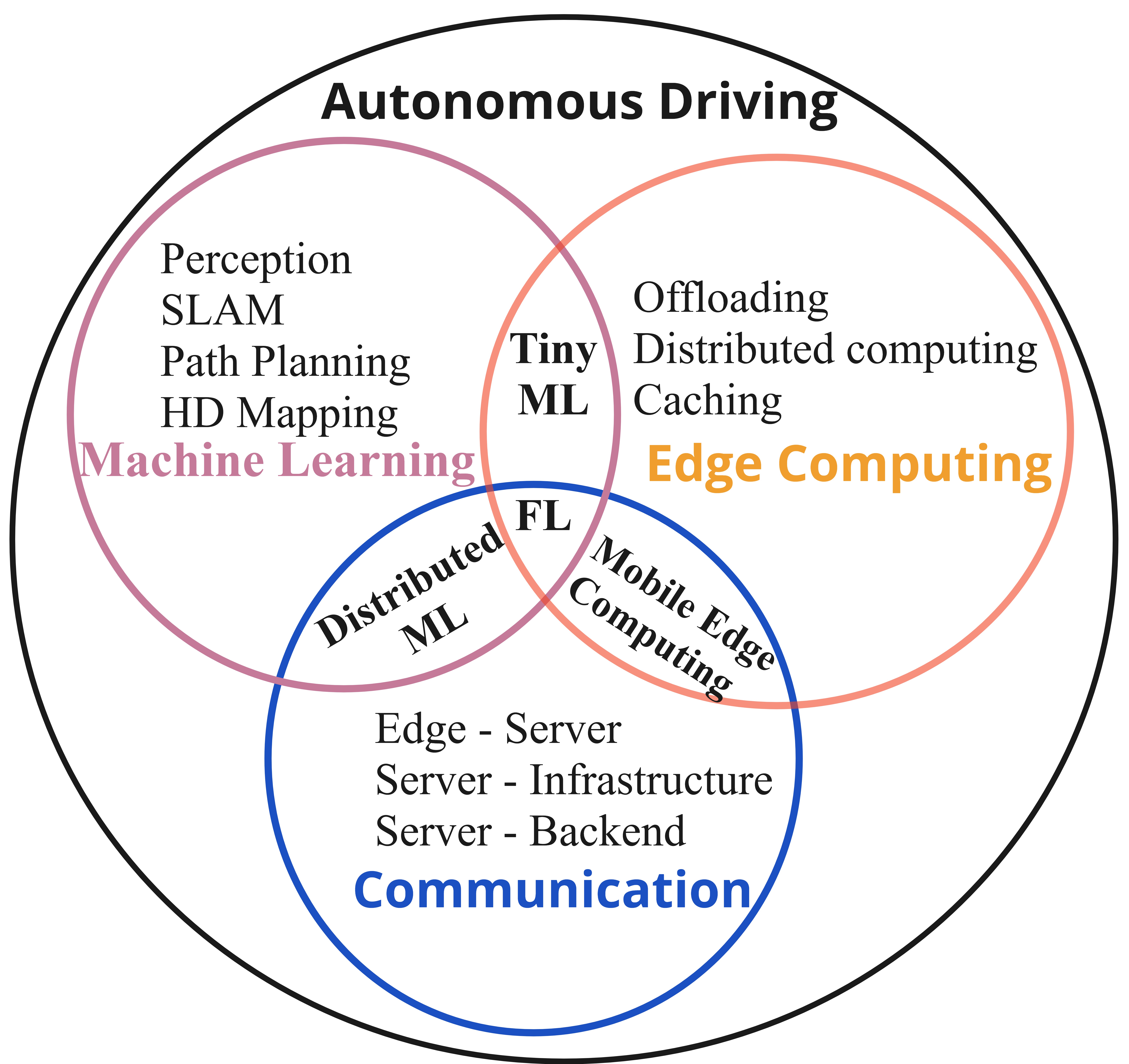}
  \caption{Overlap of ML-Driving Services, Communication and Edge Computing}
  \label{overlap}
\end{figure}

Figure~\ref{overlap} is shown based on compare and contrast approach, to merge the content and show an overlap of energy-efficient methods covered in this survey paper. As shown the topics are divided into machine learning based application for autonomous driving services, Edge computing based methods for autonomous driving and the vehicular communication. As these approaches have varied system demands, based on latency, memory and computational requirement, an attempt to show the overlapped area where software approximation can be applied has been made. The emerging areas are Tiny ML (promotes deep learning in compressed form in embedded processors), Distributed Machine Learning \& FL which implements collaborative training and inference among several embedded and edge devices. Mobile Edge computing has also emerged as a popular topic which allows processing of data and decision making process close to the Edge thus overcoming latency and memory drawbacks. Rest of this section discusses computing-efficiency and compression methods. 

\subsubsection{Computing Efficiency}
DNN based vision oriented systems such as object classification, 3D object detection and SLAM are usually computational intensive, high resource and energy consuming tasks. The computing complexity relatively increases for real-time applications when these larger weight DNN are implemented on the embedded systems with limited memory \cite{PE10}. For example the currently deployed level 3 autonomous vehicles \cite{krok2020tesla, waymo_2020} are mostly dependent on vision sensors systems and consumes significant resources in terms of memory and energy. The scalability of these applications on embedded systems with fully connected cooperative autonomous vehicles is yet to be known incorporating full ADAS features. With the implementation of fully connected autonomous driving, the common assumption is the complex calculation and usage of deep/dense neural network will increase the calculation time, thus making some real-time applications difficult to process within the required latency, and on the other hand, the large weights of the neural network will also bring challenges to some embedded systems with limited memory \cite{AC4, AC5, AC6, AC7, AC8}. Therefore, there is a need to implement and develop low-weight and compressed neural network for efficient and low-latency calculations.

\subsubsection{Compression}
Compression is an approximation technique which can be implemented for the model and the data to allow the real-time inference on resource constrained devices. Some of the popular compression technique in deep learning involves pruning, low-rank approximation, quantization, knowledge-distillation, sketching. Deep Compression \cite{han2015deep} proposed by Han et. al, implements combination of pruning, quantization and Huffman coding on the state-of-art deep neural network such as Alexnet, VGG-16 by maintaining the architecture accuracy. In federated learning practices along with the deep learning approximation technique, the compression is also implemented in communication algorithms using sparsification of gradients. In this section this survey paper discusses these compression approaches by also mentioning some popular inference methods for resource constrained embedded devices.

\textbf{Low-rank Approximation:}
A direct mathematical approach to compress a dense neural network is low-rank approximation. As traditional neural network are developed on filters and layer comprising of several matrix, factorization \cite{swaminathan2020sparse, Sainath2013} and decomposition \cite{denton2014exploiting, jaderberg2014speeding, GUO2018, WenXWWCL17, alvarez2017compression, xu2020trp, indyk2019learning, lee2019learning} of these matrix has helped in reducing the parameters from the neural network, the popular approaches involves singular value decomposition \cite{indyk2019learning,GUO2018}, tucker decomposition \cite{kim2015compression} and canonical polyadic decomposition \cite{astrid2017cp}. For decomposition the approach can be targeted to reduce the parameter for overall dimension reduction or targeting a channel through decomposing the relevant filter. In \cite{GUO2018} authors proposed a method in which convolutional filter with low rank are decomposed into several depth-wise and point-wise filter. With this approach the large scale model size is compressed and could be easily deployed on mobile and edge devices, however accuracy loss for the network is higher as few high ranked filter could still be decomposed in this approach based on the assumption from a neighbor low-ranked filter. Another approach to prevent accuracy loss is implementing sparse regularization \cite{mitsuno2020hierarchical, alvarez2017compression} in an hierarchical manner as this approach can enhance network learning by grouping the filter which can be decomposed based on magnitude. Other techniques \cite{indyk2019learning, lee2019learning} involves finding kernel or filter with low magnitude during training to enhance the model learning (Accuracy) and later applying a singular value decomposition to achieve a better compression ratio.

\textbf{Pruning:}
Pruning is originally a technique applied in agriculture or horticulture to remove certain parts of tree or plant (branch, leaves, stubs) which are not effectively contributing. Inspired from this idea, researcher has applied and implemented pruning in convolutional or deep neural network to compress and reduce the overall parameter of these neural networks and to enable deployment an easy process on resource constrained embedded device for real-time application which also requires smaller models with fast computation process. In current practice there are two popular approach for pruning, removal of weights \cite{RC5, Jiang2019Pruning, Tprune, Yang2018, Yang2019, Yang2020Pruning, shaohui2019} and removal of neurons \cite{tan2020dropnet, Ruichi2018, li2016pruning, luo2017thinet, li2019single, liu2021pruning, nakadai2021investigation} respectively. Removal of weights from neural network does not affect the accuracy of model as only those weights are removed which have a magnitude close to zero. Since the implementation of weights removal is based on sparse matrix computation, in some cases it requires dedicated processors to apply this method in neural network \cite{Tprune, RC5}. For these methods authors have also proposed Structured Sparsity Learning (SSL) framework designs for hardware (e.g. mobile computing, FPGA framework) \cite{RC5}. In \cite{Yang2018, Yang2020Pruning} the approach covers pruning the soft-filter where filters are pruned while training a DNN model in iterative manner after the model has been trained for an epoch, based on the magnitude or score. The methodology used for scoring the filter is based on ($l_1$ or $l_2$) normalization. Once the model is pruned, there are changes in the hyper-parameter and dimension of the network, therefore it is important to adjust them by reconstructing the pruned filter using forward and backward propagation. The second approach which involves removal of neurons is based on heuristic methods and directly impacts the accuracy and overall performance of the neural network however the model performance can be optimized with the fine-tuning \cite{Dongkuan2021, xiao2019autoprune} or model retraining practices.

\textbf{Quantization:}
Uniform and non-uniform quantization techniques are popular methods to compress an AI model. In the uniform quantization technique \cite{fangxin2020ausn, choi2018pact, zhou2017balanced}, a linear approach is used to distribute the quantized values over the space uniformly. While in non-uniform quantization, the logarithmic or exponential approach is used to distribute the quantized values non-uniformly. Methods to quantize deep neural network non-uniformly is presented in \cite{jung2019learning, yang2019quantization, liao2020sparse, jeon2020biqgemm}, which is based on quantization interval learning. Here the quantization intervals are parameterized over the intervals, and the obtained function is applied over the weights and activation of the deep neural network to achieve model compression. Quantization has also helped reduce CNN's overall weight and size, which consists of many convolutional layers. Quantization for layers has been proposed in \cite{aghasi2017net, zhu2018adaptive, gluska2020exploring, panda2020quanos} by using the statistical parameter or scaling factor for the layer. This granular based approach can significantly reduce the model size, however it also results in relative loss of model accuracy as a kernel or filter containing important feature will loose its weights because of another kernel or filter with no feature present in the same layer. A better approach to counter this problem is quantization in group \cite{yuan2021towards, shen2020q, EVOQ}, where kernel or filter with no feature or weights can be grouped together and removed. This approach maintains the architecture accuracy but requires additional scaling parameter for each layer. Recent used approach in granular quantization is with channels \cite{huang2021codenet}, in this approach the length of activation and weights are scaled for each channel to reduce the overall weight \cite{li2019fully, zhong2020channel}  for each convolution filter during training. The scaling factor is applied on input feature maps and output feature maps of the channels as they have different lengths, which results in parameter reduction without loss in accuracy. Some applications require to modify or rearrange the parameter of convolution or deep neural network after the model is trained, this approach is often termed as quantization aware training and post-training quantization. Quantization aware training process includes retraining the model with methods such as: straight through estimator \cite{fan2020training, zhuang2018towards, yin2019understanding}, target propagation \cite{nokland2016direct, lee2015difference, demidovskij2020effective}, regularization \cite{nakata2021adaptive, sasaki2019post}. 

\textbf{Knowledge-Distillation:}
Another efficient approach of deploying large sized neural network to edge devices is Knowledge distillation. This technique \cite{ahn2019variational, tung2019similarity, yim2017gift, park2019relational, mishra2017apprentice, jin2021kdlsq, choi2020data, sarfraz2021knowledge, liu2021exploring} consists of two processes, in first part the large model is trained over a complete set of dataset on high performing devices, which results in output feature maps predictions. In the second process a compressed version of the large model is trained over the dataset (sampled form + ground truth), which results in output feature maps predictions, which is then combined with the output feature maps of larger model thus providing knowledge (distilled) from larger model to the compressed one by still marinating accuracy and net loss. Some approaches involves \cite{sarfraz2021knowledge} direct correspondence between layer of large and smaller model sometimes also referred as utilising the soft probabilities from larger network to train smaller network rather than the ground truth, as this information not only contains the output feature maps but also the activation maps thus making the smaller network learning faster. This approach has shown potential for transferring the large models from high performance devices to edge devices or embedded processors, but to achieve high model compression ratio with soft probabilities or direct correspondence is still a challenge. As the other approaches such as pruning and quantization is capable of balancing a trade-off between accuracy and compression ratio. Some approaches \cite{kim2021pqk, okuno2021lossless, li2021mixmix} also involves using combination of multiple compression techniques: knowledge distillation, pruning, and quantization to achieve better accuracy and compression ratio.

\subsubsection{Role of Edge AI}
This section discusses the influence of edge computing and related applications on autonomous driving. As the volume of data keeps on growing with the number of sensors, a research direction is focused on processing data near the sensing device. Cloud computing, cloud centralized intelligence \cite{EE3, s21} was initially proposed as solution for fully connected autonomous driving, however the latency requirement for time sensitive applications and the expected bandwidth (Table~\ref{t:edge-cloud} shows comparison of Edge and Cloud intelligence) for data transmission became a challenge. To address this challenge Edge Intelligence has been proposed as a suitable solution, which allows processing of data closer to the edge device rather than in a centralized cloud. 

\begin{table}[!ht]
\caption{Edge Intelligence \& Cloud Intelligence parameters comparison for Self-driving vehicles}
\label{t:edge-cloud}
\resizebox{\linewidth}{!}{
\begin{tabular}{|c|c|c|}
\hline
{\color[HTML]{000000} \textbf{Parameters}}                                  & {\color[HTML]{036400} \textbf{\begin{tabular}[c]{@{}c@{}}Vehicular Edge\\ Intelligence\end{tabular}}} & {\color[HTML]{00009B} \textbf{\begin{tabular}[c]{@{}c@{}}Cloud\\ Intelligence\end{tabular}}} \\ \hline
\textbf{Architecture}                                                       & \begin{tabular}[c]{@{}c@{}}Heterogeneous ASIC \\ Accelerator\end{tabular}                              & \begin{tabular}[c]{@{}c@{}}CPU, GPU, \\ TPU, FPGA\end{tabular}                               \\ \hline
\textbf{\begin{tabular}[c]{@{}c@{}}Computing\\ Performance\end{tabular}}    & Medium                                                                                                & High                                                                                         \\ \hline
\textbf{Storage}                                                            & Limited                                                                                               & Highly Scalable                                                                              \\ \hline
\textbf{Power Consumption}                                                  & Low                                                                                                   & High                                                                                         \\ \hline
\textbf{\begin{tabular}[c]{@{}c@{}}Context-Aware \\ Computing\end{tabular}} & Applicable                                                                                            & Not Applicable                                                                               \\ \hline
\textbf{Architecture Topology}                                              & Distributed                                                                                           & Centralized                                                                                  \\ \hline
\textbf{Deployment Cost}                                                    & Low                                                                                                   & High                                                                                         \\ \hline
\textbf{Reliability}                                                        & High                                                                                                  & High                                                                                         \\ \hline
\textbf{Security}                                                           & High                                                                                                  & Limited                                                                                      \\ \hline
\textbf{Communication}                                                      & Wireless                                                                                              & Wireless + Optical                                                                           \\ \hline
\textbf{Computation}                                                        & Locally                                                                                               & Central Server                                                                               \\ \hline
\textbf{Bandwidth Requirement}                                              & Low transmission rate                                                                                 & High transmission rate                                                                       \\ \hline
\textbf{Latency}                                                            & Low                                                                                                   & High                                                                                         \\ \hline
\end{tabular}}
\end{table}

In \cite{r11} the authors presented in detail about the motivation and benefits of using edge intelligence where the primary concepts highlighted and can be linked with autonomous vehicles are: the volume of data generated by vehicle senors at the edge device need machine and deep learning approaches for processing and decision making process thus proposing the concept of AI at the Edge. The concept has been proposed in several stages where the primary focus is on  transmission of sensed data to the server or cloud for processing and decision making. The first stage contains the parameters of cloud intelligence shown in Table ~\ref{t:edge-cloud}, thus allowing training and inference via a centralized cloud. The second stage comprises of edge-server joint training and inference. In this stage depending upon the requirement and processing ability the model can be jointly  trained at the edge and server or at the server and inference occurs at both using distributed learning and computing methods.


The last stage of edge intelligence allows the training and inference occurrence on the device itself or near the device (edge) through data offloading and real-time compressed sensing approaches \cite{ECS6}. For autonomous driving applications Pi-Edge \cite{PI-Edge} and AVe \cite{AVE} are the two initial proposed framework consisting of driving services with data offloading and resource allocation techniques. Later proposed edge AI framework for autonomous driving \cite{Lopecs}, is also influenced by Pi-Edge and proposed data offloading and resource allocation scheme, thus allowing edge-server joint inference using hybrid communication architecture. However the framework misses energy saving mechanism and the assumptions on trade-off which data offloading and compression brings on the end-to-end accuracy of the model. In \cite{santa2019surrogates, santa2020migrate} the authors propose intelligent edge architecture for autonomous driving vehicles with OpenStack and ETSI open-source MANO. Using the architecture the allocated and resources of edge devices can be visualized at the server or cloud and also allows managing of mutli-access edge and mobile computing, thus allowing to free edge device memory from raw data using offloading.

In \cite{ibn2021edge} the authors proposes an edge architecture with low latency communication and resource allocation scheme for compute intensive tasks. Using the reference architecture the authors designed an advanced autonomous driving communication protocol to enhance and facilitate communication between edge device, servers, data centers and the centralized cloud. Here the cloud contains legacy or ground truth data contributed from the vehicle sensors, servers, infrastructure sensors and the vehicular surrounding. For the decision making process a deep reinforcement learning approach is used for training and inference. The edge frameworks, offloading schemes and approximations are comprehensively covered in section IV and V.

\begin{table*}[!t]
\caption{Publicly available Dataset for Autonomous Driving\\ The table is arranged according to the timeline of release, URL's were last accessed on 15-February-2023.}
\label{t:Datasets}
\begin{center}
\begin{tabular}{|c|c|c|c|c|c|c|c|c|}
\hline
\multirow{2}{*}{\textbf{Year}} & \multirow{2}{*}{\textbf{Dataset}} & \multicolumn{7}{c|}{\textbf{Sensors Included}}\\ 
\cline{3-9} &   & \textbf{Camera} & \textbf{LiDAR} & \textbf{Radar} & \textbf{GPS/GNSS} & \textbf{IMU} & \textbf{HD MAP} & \textbf{URL} \\ \hline
2012 - 2015 & KITTI \cite{DB4} &Y  &Y  &N  &N  &Y  &N &\href{http://www.cvlibs.net/datasets/kitti/eval_object.php?obj_benchmark=3d}{KITTI} \\ \hline

2015 - 2019 & KAIST Dataset \cite{DB9} &Y  &Y  &N  &Y  &Y  &N &\href{https://irap.kaist.ac.kr/dataset/}{KAIST}\\ \hline

2016 & HD1K \cite{DB14} &Y  &Y  &N  &N  &N  &N &\href{http://hci-benchmark.iwr.uni-heidelberg.de/}{HD1K}\\ \hline
2016 & CVC-14 \cite{DB15} &Y  &N  &N  &N  &N  &N &\href{http://adas.cvc.uab.es/elektra/enigma-portfolio/cvc-14-visible-fir-day-night-pedestrian-sequence-dataset/}{CVC-14}\\ \hline
2016 & Brain4Cars \cite{DB17} &Y  &N  &Y  &Y  &N  &N &\href{http://brain4cars.com/}{Brain4Cars}\\ \hline
2016 & JAAD \cite{DB18}  &Y  &N  &N  &N  &N  &N &\href{https://github.com/ykotseruba/JAAD/tree/master}{JAAD}\\ \hline
2016 & Cityscapes \cite{DB19} &Y  &N  &N  &Y  &Y  &N &\href{https://www.cityscapes-dataset.com/}{CITYSCAPES}\\ \hline
2016 & Udacity &Y  &N  &N  &N  &N  &N &\href{https://github.com/udacity/self-driving-car/tree/master/datasets}{UdaCity}\\ \hline
2016 - 2019 & comma.ai driving dataset \cite{DB20} &Y  &N  &Y  &Y  &Y  &N &\href{https://github.com/commaai/comma2k19}{Comma datasets}\\ \hline

2017 & TRoM \cite{DB21} &Y  &N  &N  &N  &N  &N &\href{http://www.tromai.icoc.me/}{TRoM}\\ \hline
2017 & DDD17 \cite{DB22} &Y  &N  &N  &N  &Y  &N &\href{http://sensors.ini.uzh.ch/news_page/DDD17.html}{DDD17}\\ \hline
2017 & Raincouver \cite{DB23} &Y  &N  &N  &N  &N  &N &\href{https://www.cs.ubc.ca/~ftung/raincouver/index.html}{Raincouver}\\ \hline
2017 & VPGNet \cite{3d22}&Y  &N  &N  &Y  &N  &N &\href{https://github.com/SeokjuLee/VPGNet}{VPGNet}\\ \hline
2017 & TuSimple  &Y  &N  &Y  &N  &N  &N &\href{https://github.com/TuSimple/tusimple-benchmark/issues/3}{TuSimple}\\ \hline
2017 & TorontoCity \cite{DB25} &Y  &Y  &N  &N  &N  &N &\href{https://arxiv.org/abs/1612.00423}{TorontoCity}\\ \hline
2017 & CityPersons  &Y  &N  &N  &N  &Y  &N &\href{https://github.com/CharlesShang/Detectron-PYTORCH/tree/master/data/citypersons}{CityPersons}\\ \hline
2017 & Mapillary Vistas \cite{DB26} &Y  &N  &N  &N  &N  &N &\href{https://www.mapillary.com/dataset/vistas}{Mapillary Vistas}\\ \hline
2017 & Multi-spectral (Univ of Tokyo) \cite{DB27} &Y  &N  &Y  &N  &N  &N &\href{https://www.mi.t.u-tokyo.ac.jp/static/projects/mil_multispectral/}{Multi-spectral}\\ \hline

2018 & CULane \cite{3d25} &Y  &N  &N  &Y  &Y  &N &\href{https://xingangpan.github.io/projects/CULane.html}{CULane}\\ \hline
2018 & DBNet \cite{DB29} &Y  &Y  &Y  &Y  &Y  &N &\href{http://www.dbehavior.net/}{DBNet}\\ \hline
2018 & IDD \cite{DB30} &Y  &N  &N  &N  &N  &N &\href{https://idd.insaan.iiit.ac.in/}{IDD}\\ \hline
2018 & MVSEC (U Penn) \cite{DB31} &Y  &Y  &N  &N  &N  &N &\href{https://daniilidis-group.github.io/mvsec/}{MVSEC}\\ \hline
2018 & NightOwls \cite{DB32} &Y  &N  &N  &N  &N  &N &\href{https://www.nightowls-dataset.org/}{NightOwls}\\ \hline

2018 & Road Damage \cite{DB33} &Y  &N  &N  &N  &N  &N &\href{https://github.com/sekilab/RoadDamageDetector/}{Road Damage}\\ \hline

2018 & Wilddash \cite{DB34} &Y  &N  &N  &N  &N  &N &\href{https://wilddash.cc/}{wildDash}\\ \hline
2018 - 2020 & BDD-100K \cite{DB35} &Y  &Y  &N  &Y  &Y  &N &\href{https://bdd-data.berkeley.edu/}{Berkeley}\\ \hline
2018 - 2020 & ApolloScape \cite{apolloscape} &Y  &Y  &N  &Y  &Y  &N &\href{http://apolloscape.auto/index.html}{Apollo}\\ \hline
2018 - 2020 & Honda Driving \cite{DB39}  &Y  &Y  &N  &Y  &Y  &N &\href{https://usa.honda-ri.com/H3D}{HDD}\\ \hline

2019 & \bcl{lb}{Argoverse} \cite{DB40} &Y  &Y  &N  &N  &N   &Y &\href{https://www.argoverse.org/index.html#download-link}{Argo}\\ \hline

2019 & Astyx HiRes \cite{DB41} &Y  &Y  &N  &N  &N   &N &\href{https://www.astyx.com/development/astyx-hires2019-dataset.html}{Astyx}\\ \hline

2019 & BLVD \cite{DB42} &Y  &Y  &N  &N  &N  &N &\href{https://github.com/VCCIV/BLVD}{BLVD}\\ \hline
2019 & Boxy Driving \cite{DB43}  &Y  &N  &N  &N  &N  &N &\href{https://boxy-dataset.com/boxy/}{BOSCH}\\ \hline
2019 & EuroCity \cite{DB44} &Y  &N  &N  &N  &N  &N &\href{https://eurocity-dataset.tudelft.nl/}{Eurocity Persons}\\ \hline
2019 & \bcl{lb}{EU Long-term Dataset} \cite{DB45} &Y  &Y  &Y  &Y  &Y  &N &\href{https://epan-utbm.github.io/utbm_robocar_dataset/}{EU Dataset}\\ \hline
2019 & IceVisionSet \cite{DB46} &Y  &Y  &N  &Y  &N  &N &\href{http://oscar.skoltech.ru/}{IceVision}\\ \hline
2019 & StreetLearn \cite{DB47} &Y  &N  &N  &N  &N  &N &\href{https://sites.google.com/view/streetlearn/dataset}{Street Learn}\\ \hline
2019 & PandaSet &Y  &Y  &N  &Y  &N  &N &\href{https://scale.com/resources/download/pandaset}{PandaSet}\\ \hline
2019 & WoodScape \cite{DB64} &Y  &Y  &N  &Y  &Y  &N &\href{https://woodscape.valeo.com/dataset}{WoodScape}\\ \hline
2019 & Unsupervised Llamas - Bosch \cite{DB48} &Y  &Y  &N  &Y  &N  &N &\href{https://unsupervised-llamas.com/llamas/}{Bosch}\\ \hline

2020 &$4-$Seasons \cite{wenzel2020fourseasons} &Y  &N  &N  &Y  &Y  &N &\href{https://www.4seasons-dataset.com/dataset}{4-Seasons}\\ \hline
2020 & A*3D \cite{DB49} &Y  &Y  &N  &N  &N  &N &\href{https://github.com/I2RDL2/ASTAR-3D}{ASTAR-3D}\\ \hline
2020 & \bcl{lb}{nuScenes} \cite{DB53}  &Y &Y  &Y  &Y  &Y  &Y &\href{https://www.nuscenes.org/nuscenes}{nuscenes}\\ \hline
2020 & POSS \cite{DB63} &Y  &Y  &N  &N  &N  &N &\href{http://www.poss.pku.edu.cn/semanticposs.html}{POSS}\\ \hline
2020 & DDD20 \cite{DB50}  &Y  &N  &N  &Y  &Y  &N &\href{https://sites.google.com/view/davis-driving-dataset-2020/home}{DDD20}\\ \hline
2020 & Highway Driving \cite{DB51} &Y  &N  &N  &N  &N  &N &\href{https://sites.google.com/site/highwaydrivingdataset/}{Kaist}\\ \hline
2020 & Lyft Level 5 \cite{DB52} &Y  &Y  &N  &N  &N  &Y &\href{https://self-driving.lyft.com/level5/data/}{lyft}\\ \hline
2020 & \bcl{lb}{Brno Urban Dataset}  &Y  &Y  &Y  &Y  &Y  &N &\href{https://github.com/Robotics-BUT/Brno-Urban-Dataset}{BRNO}\\ \hline
2020 & \bcl{lb}{Ford Multi AV} \cite{DB55} &Y  &Y  &N  &Y  &Y  &Y &\href{https://avdata.ford.com/home/default.aspx}{Ford Seasonal}\\ \hline
2020 & A2D2 \cite{DB56} &Y  &Y  &N  &N  &N  &N &\href{https://www.a2d2.audi/a2d2/en/dataset.html}{Audi}\\ \hline
2020 & LIBRE \cite{DB57} &Y  &Y  &Y  &Y  &Y  &N &\href{https://sites.google.com/g.sp.m.is.nagoya-u.ac.jp/libre-dataset}{LIBRE}\\ \hline
2020 & Toronto-3D &Y  &Y  &N  &Y  &Y  &N &\href{https://github.com/WeikaiTan/Toronto-3D#download}{Toronto-3D}\\ \hline
2021 & \bcl{lb}{NEOLIX} \cite{DB61} &Y  &Y  &Y  &Y  &Y  &N &\href{https://gas.graviti.cn/dataset/graviti-open-dataset/NeolixOD}{Neolix} \\\hline
2021 & CADC \cite{DB58} &Y  &Y  &N  &Y  &Y  &N &\href{http://cadcd.uwaterloo.ca/}{CADC}\\ \hline
2021 & RadarScenes \cite{DB59} &Y  &N  &Y  &Y  &Y  &N &\href{https://radar-scenes.com/}{RadarScenes}\\ \hline

2021 & CARRADA \cite{DB60} &Y  &N  &Y  &N  &N  &N &\href{https://arthurouaknine.github.io/codeanddata/carrada}{CARRADA}\\ \hline
2021 & \bcl{lb}{Waymo} \cite{DB54} &Y  &Y  &N  &N  &N  &Y &\href{https://waymo.com/open/}{Waymo Open}\\ \hline
2021 & SODA10M \cite{DB65} &Y  &N  &N  &N  &N  &N &\href{https://soda-2d.github.io/index.html}{SODA10M}\\ \hline
2021 & \bcl{lb}{PixSet:LeddarTech} \cite{DB62} &Y  &Y  &Y  &Y  &Y  &N &\href{https://leddartech.com/solutions/leddar-pixset-dataset/}{PixSet} \\\hline
2021 & ONCE \cite{DB6} &Y  &Y  &N  &N  &N  &N &\href{https://once-for-auto-driving.github.io/download.html}{ONCE}\\ \hline
2021 & \bcl{lb}{Deep Route AI} &Y  &Y  &Y  &Y  &Y  &Y &\href{https://gas.graviti.cn/dataset/graviti-open-dataset/DeepRoute}{Deep Route} \\\hline
2021 & {DurLAR}\cite{li21durlar} &Y  &Y  &N  &Y  &Y  &N &\href{https://collections.durham.ac.uk/collections/r2gq67jr192}{DurLAR} \\\hline
2022 & {MUAD}\cite{franchi2022muad} &Y  &N  &N  &N  &N  &N &\href{https://muad-dataset.github.io/}{MUAD} \\\hline
2022 & {SHIFT} &Y  &Y  &N  &N  &Y  &N &\href{https://github.com/SysCV/shift-dev}{SHIFT} \\\hline
2022 & {Rope3D}\cite{Ye_2022_CVPR} &Y  &Y  &N  &Y  &N  &N &\href{https://thudair.baai.ac.cn/rope}{Rope3D} \\\hline
2022 & {CODA}\cite{li2022coda} &Y  &Y  &N  &Y  &N  &N &\href{https://coda-dataset.github.io/}{CODA} \\\hline
2022 & \bcl{lb}{View-of-Delft} \cite{apalffy2022} &Y  &Y  &Y  &Y  &Y  &N &\href{https://tudelft-iv.github.io/view-of-delft-dataset/}{Delft-View} \\\hline
\end{tabular}
\end{center}
\end{table*}

\subsection{An overview of Dataset for Autonomous Driving}
\begin{figure}[!ht]
  \centering
  \includegraphics[width=\linewidth]{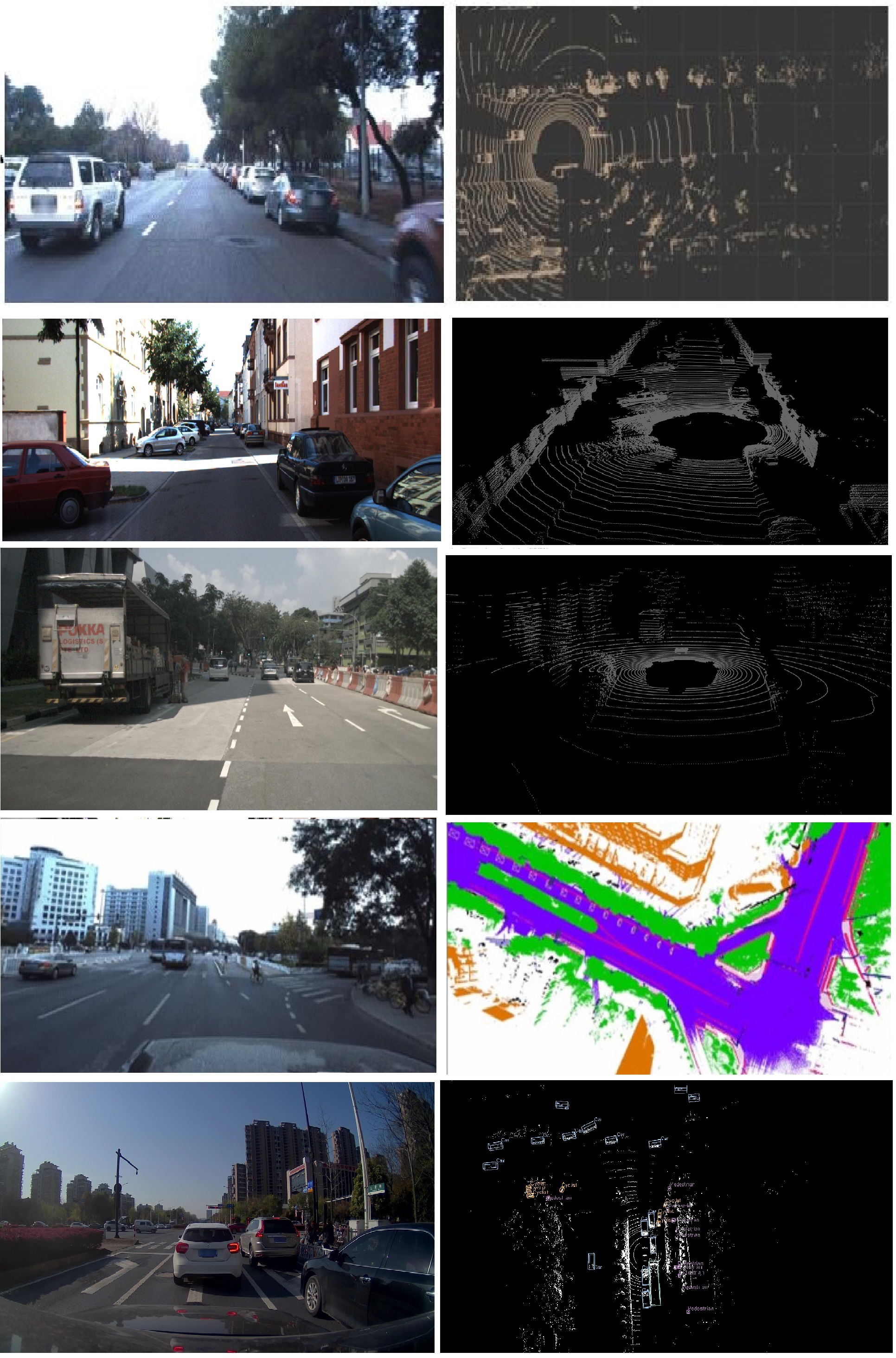}
  \caption{Frames and point clouds from popular datasets. Images are from Lyft, KITTI, nuScenes, ApolloScape, and ONCE dataset\cite{DB52, DB4, DB53, apolloscape, DB6}, respectively.}
  \label{data-sample}
\end{figure}

An important requirement to develop machine/deep learning based autonomous driving services or tasks is dependent dataset. Several datasets has been made available by the universities research groups, and the automotive companies in the last decade. In this article an attempt to categorise these datasets has been made on the basis of sensors and the driving application which can be derived as a result. Based on convolutional neural network, one of the most researched topic is object detection containing several classes such as pedestrians, traffic signs, lane, vehicles (cars, truck, ambulance, school bus). The advancement in minor features recognition from the image or video frames also resulted in development of applications such as: vehicle model detection, license plate classifier, and other cooperative applications. Some of the commonly used datasets are KITTI \cite{DB4}, Cityscapes \cite{DB19} and PASCAL VOC \cite{PascalVoc}. After 2017 high quality data comprising of multi sensors primarily camera and LiDAR has been collected and released for development of advanced applications targeting level 5 autonomy \cite{DB52, DB54} also shown in Figure~\ref{data-sample}.

To prevent developing biased AI models, the traffic scenes or data were also combined from multiple continents, countries and cities. The EU Long-term dataset \cite{DB45} is collected in several location within europe, nuscenes \cite{DB53} collected in Singapore and USA, comprises of multi-sensor suite. Argoverse \cite{DB40} dataset collected by Ford is one of the unique dataset which also provides functionality to try and test the high definition map applications based on LiDAR and camera sensors. As the sensor/data fusion approach is being researched for low powered embedded devices, the driving tasks, such as adaptive cruise control, path planning, and SLAM has involved usage of radar sensor values with the LiDAR point clouds and the camera frames. Radarscenes \cite{DB59}, Astyx HiRes \cite{DB41}, Ford multi av\cite{DB55}, Neolix \cite{DB61}, Pixset \cite{DB62}, are some datasets which provides the annotations on data based on these three sensors. Similarly another high quality dataset also comprising of HD Map annotation has been made publicly available by the Deep Route AI targeting the level 4+ Full-stack self-driving system. Table~\ref{t:Datasets} shows list of open-sourced datasets available for the AI model development and testing. 

\begin{tcolorbox}[title=\textbf{Lessons Learned}, breakable, skin=enhanced jigsaw]

\begin{enumerate}

    \item Adversity: Popular datasets do not include unexpected or undesirable uncertainties, as it is difficult to estimate a ground truth for them. Similarly, there is a limited representation of different weather and light conditions in the training and testing datasets. An AI model trained/validated on such a dataset might not be generalisable.
    
    \item Biases: The majority of the datasets are collected from urban driving conditions. This does improve the accuracy and development of an AI model for urban driving scenarios but also brings significant challenges to the model's adaptability to diverse and dynamic conditions such as highway driving or severe weather conditions. 
    
    \item Disparity: A form of bias can be inherited in AI models due to the disparity of annotated classes. Popular driving datasets generally discuss the number of scenes, annotations, and bounding boxes covered for training-testing. However, they lack a discussion on diversity and the distribution of classes covered. For example, the annotations of vehicles, and traffic signs are much higher represented as compared to cyclists, motorcyclists, or pedestrians.
    
    \item Data fusion and Collection format: Statistical models are developed and adapted as per the format of datasets. Current datasets vary in logging approaches which brings challenges to model or cross-data transformation which can also create a bias on the developed AI algorithm.
    
\end{enumerate}
\end{tcolorbox}

\section{Edge AI with Autonomous Driving}
\begin{figure}[!ht]
  \centering
  \includegraphics[width=\linewidth]{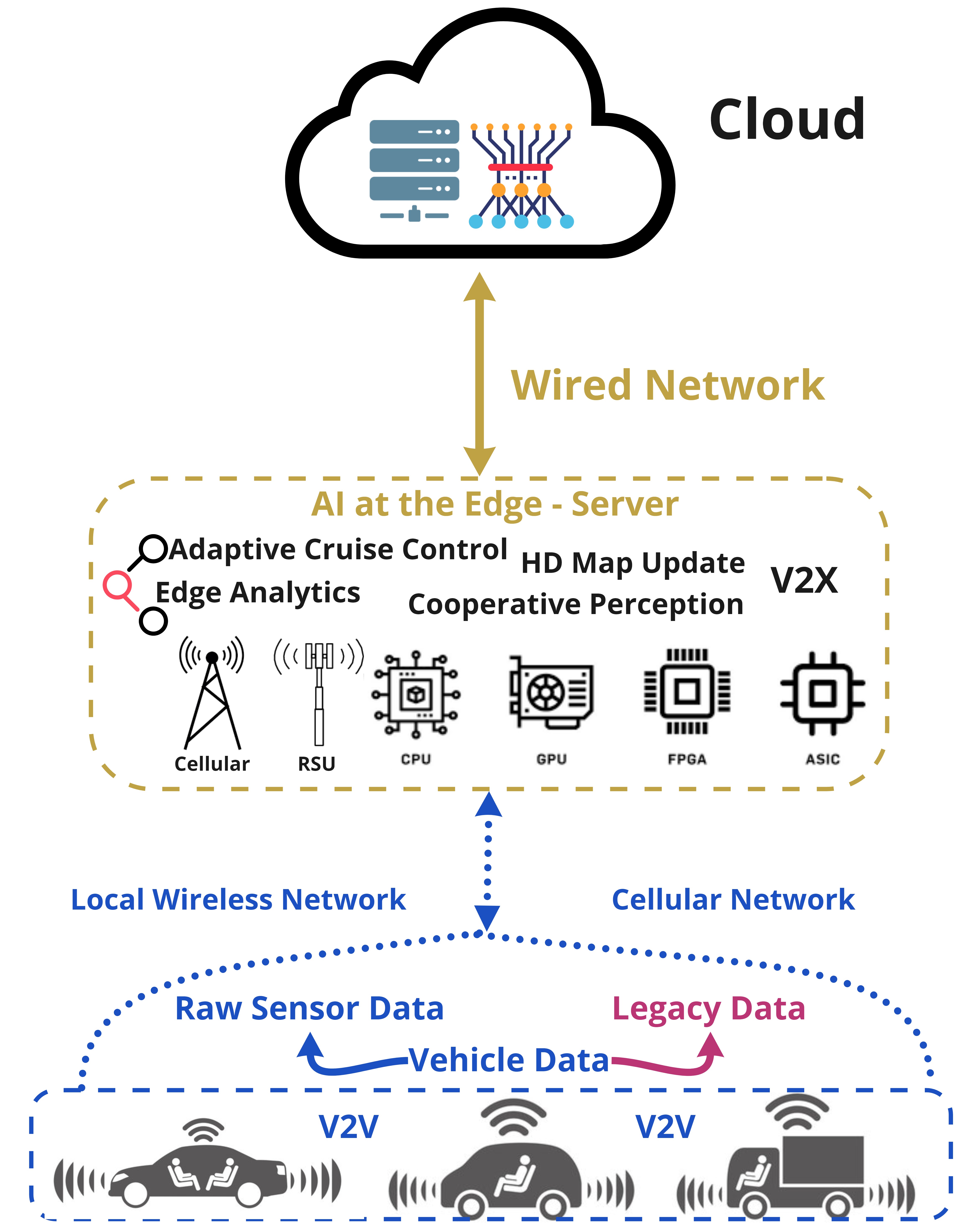}
  \caption{Edge AI layers for connected vehicles}
  \label{Edge-Ad}
\end{figure}

Edge computing systems have already been used and tested IoT use-cases or applications, which require relatively less computation, and power \cite {PE3, PE4}. Hardware manufacturers such as Nvidia, IBM, Intel, Qualcomm, NXP has developed and released edge computing hardware with respect to the dedicated tasks such as speech recognition and vision based applications. For autonomous driving the edge intelligence demands data processing pipeline which should be capable of data management, analysis and data storage. Popularly used vehicle edge computing devices include Nvidia's Jetson and Xavier Platform. These devices are largely used in combination with on-board sensors such as: cameras, LiDAR, radar, IMU, GNSS and V2X module or router for communication with other devices and server. As per current description the subsystems required to enable fully connected autonomous vehicle comprises of: the autonomous vehicle containing cellular or edge connectivity, the roadside units connected with the infrastructure, Edge server, the micro data centers, and lastly the cloud or main server having connectivity with all the mentioned subsystem and the autonomous vehicle, a description and layers are shown in Figure~\ref{Edge-Ad}. It is important to note that the introduction of vehicular edge computing and intelligence \cite{HD1}, have further strengthened the scope and area of vehicle-to-everything communication (V2X) \cite{vc14, vc25}. The key components for enabling edge artificial intelligence for autonomous driving includes edge training, inference, caching, optimization, and communication. Vehicular communication has already been covered in the previous section, however distributed approaches such as federated learning remains, therefore this section first discusses Edge training and inference, Edge computing-based applications for autonomous driving, and recently proposed federated learning approaches, cooperative and collaborative autonomous driving.

\subsection{Edge Computing and Intelligence}
The future of autonomy in vehicle has been previously proposed with centralized cloud \cite{s21} and machine/deep learning algorithms deployed at cloud \cite{EE3}, however transmitting the large volume data from the vehicle to cloud and receiving the model weights from cloud to vehicle brings latency issues for the time critical applications such as SLAM. This technical challenges leads to bringing artificial intelligence closer to the edge using distributed learning, in this context edge device (present in vehicle) and edge-server (present in vehicle surrounding), corresponding abstraction of Edge AI layer is shown in Figure~\ref{Edge-Ad}. Some of the proposed collaborative applications and approaches includes perception \cite{CEAD5}, SLAM \cite{HD10, HD11, HD13}, HD map \cite{HD15}, collision warning systems\cite{CEAD7, CEAD11} and path planning \cite{AutoC2X}. 

In cooperative perception applications at edge, F-cooper \cite{CEAD5} provides collaborative object detection using high level fusion from multiple vehicles LiDAR point clouds. For object detection authors used voxel feature fusion (as shown in Figure~\ref{3d-det}), and spatial feature fusion approach. The object detection methods were lightweight and allows the transmission and sharing over dedicated short range communication. The presented approach is deployed in the edge device and the method was tested using real-world data. Similar approach is presented in \cite{ArnoldPerception}, here the authors proposes an early fusion scheme and late fusion scheme. The early fusion scheme is used for detecting the objects and the late fusion scheme is used to propose the bounding box on the detected objects. For testing the proposed approach the authors used the synthetic dataset over a T-junction and roundabout vehicle environment. For evaluation of the proposed schemes the precision, communication cost and on-board computational latency has been compared. An approach based on value-anticipating networking is proposed in \cite{CEAD6}, here the vehicle based on previous learning decides about transmitting the sensed information to other vehicle. Another cooperative perception \cite{CEAD9} is proposed using deep reinforcement learning for connected autonomous vehicles. The proposed model uses scheme to select sensed data for transmission amongst the connected vehicles. The authors further develops a cooperative vehicle simulation platform  for object detection and communication.

Similar to perception, collaborative SLAM using edge-server\cite{HD11} has been proposed for highly automated vehicles. As mentioned previously SLAM suffers with high computational demand and low latency requirement. To overcome computational requirement cloud-based SLAM has been proposed \cite{schmuck2019ccm}, however some drawbacks in centralized approach are the extreme low latency requirement and the current uplink bandwidth. Edge assisted SLAM \cite{HD10, HD11, HD13} approaches includes efficient computation, task scheduling algorithms, data offloading and sharing strategies. The backbone used in \cite{HD11, HD13} is ORB-SLAM \cite{ORB-SLAM} and ORB-SLAM2 \cite{mur2017orb} which provides the algorithm centimeter level localization accuracy. The approach uses distribution of SLAM block from ORB-SLAM2, across the edge-device and server thus overcoming the edge-device(on-board) computational complexity and processing the computation at the edge-server. To further improve the results and high precision, approaches involving crowd-source semantic mapping or fusing the results with HD map \cite{HD6} can be proposed.

\subsection{Edge Training and Optimization}
In collaborative learning setting for autonomous driving, training or retraining a model will be common practice as edge devices present in vehicles collaborate to train, a deep neural network model with the help of server acting as mode of parameter or weight updates for edge devices. For autonomous driving the edge training and optimization model should consist of model that needs to be trained, training acceleration methods, optimization parameters and model uncertainty estimation. Inspired from this, an edge training and optimization process consist of training dataset present as either raw-sensed values or as the legacy data, and the tunable parameters. For edge devices training can be organized for an individual edge device or for group of edge devices \cite{z21}. While training a model on single edge device no inputs or parameters exchange occurs, however in group training the participating edge devices communicates and share the model weights and parameters as per the set iterations.

The computational demand and memory requirements for individual training is much higher, therefore using distributed and collaborative learning approach, attention has been given to group training \cite{ye2020federated} to address the computational demand. In the group training of devices an attention is also given to communication-efficient approaches to better energy-efficiency, improve the communication round and decrease the training time. In \cite{tao2018esgd}, authors proposed a stochastic gradient descent method for improving the convolutional neural network training on the edge devices. The approach consist of sparse methods to improve the convergence rate and overall performance parameters of the model. To implement compression the gradient sparsification methods are used, which reduces the communication cost by identifying the gradients needed to share. To counter the convergence rate, which can be caused by the frequent sparse updates, a momentum residual is proposed. For evaluation, a model training using MNIST dataset was implemented.

\subsection{Edge Inference}
\begin{figure}[!ht]
  \centering
  \includegraphics[width=\linewidth]{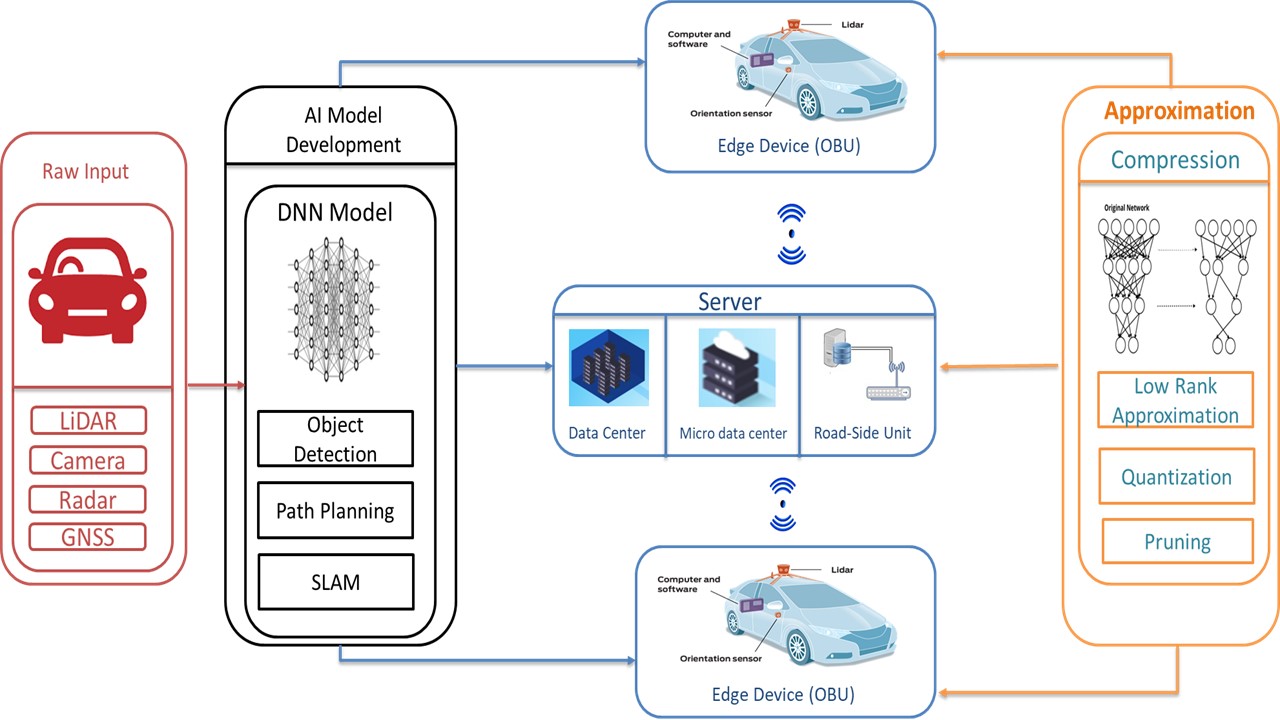}
  \caption{Edge Assisted Autonomous Driving Inference}
  \label{edge-inf-ad}
\end{figure}

Edge inference is the process of converting raw sensed data into decision making task by processing them over the AI models deployed on edge device. As mentioned previously the approach is already being used for perception, SLAM, HD map and video analytics applications. Data flow and process of edge inference is shown in Figure~\ref{edge-inf-ad}. As covered in Section II, most of the existing AI models for perception and SLAM are developed on the devices/machines which are powerful and consist of high-end graphic processing units and excessive memory. Therefore to make the AI model deployment possible on resource constrained embedded/edge device \cite{Shiqiang2018FL}, compression and software approximation approaches are implemented on the pre-trained models \cite{Tran2019FL}. 

Current Edge Inference practices for autonomous vehicles can be classified into three categories: local Inference on the edge device (vehicle), inference at Server, and joint-inference at the vehicle and server \cite{r11}. In the case of local inference, the sensing and decision-making process is performed on-board, this approach is currently in practice and requires large memory space and expensive computation devices \cite{ibn2021edge}. Local inference is very useful for lightweight applications such as on-board speech recognition. However, for heavy computational tasks, this approach suffers from computational complexity, data storage, and energy consumption problems. In server based inference, the sensing takes place on the vehicle or infrastructure sensors, and the data is uploaded to Server using wireless communication. The server is deployed with heterogeneous computing devices, processing the received data on the deep learning model, which are responsible for decision making process \cite{ECS13}. An example of analytics oriented applications are presented in \cite{ECS6, ECS7}, which contains of edge framework deploying edge intelligence based on a hierarchical manner. The approach is very useful to bring down the on-board computational cost and energy consumption, however, this practice brings challenges based on latency for time-critical applications, privacy, and security of data and model which is being shared over a wireless channel. Also, communication delay can be encountered from a corresponding server if it is responsible for the processing of data from too many vehicles at the same time. Edge-Server joint inference for connected vehicular applications is proposed in \cite{Lopecs, MobileEdge}. In these proposed approaches, the sensing takes place on-board, and based on the available on-board computational resources, part of the computation and decision-making process takes place on-board, which contains a lightweight or compressed AI model, and the remaining takes place at the server, which contains the global or dense model. After the model weights are generated individually, using an aggregation approach the model weights are combined and the decision process takes place. Edge-assisted SLAM, perception, HD map updates are some practiced and proposed methods. Some of the frameworks and approaches proposed in this category are \cite{chen2020joint, chen2021communication}. In these approaches, the common practice is to split and partition the deep neural network amongst the participating devices and server. Resource allocation scheme \cite{Nishio2019, ma2021adaptive}, communication-efficient algorithms \cite{Jeong-non-iid-data, communication-efficient2020}, task scheduler \cite{Umair2019FL, EE4}, early-exit models\cite{AC5, AC8} and heterogeneity-aware layer\cite{PIRT, EE1} are proposed in Edge-Server joint inference to take advantage of on-board and server resource to implement energy-efficient approaches. For further optimization of joint inference methods, a hardware acceleration approach such as parallel computation using  heterogeneous architecture device \cite{zeng2021energy, EE13} is proposed. In similar category, software acceleration approaches \cite{seewald2021coarse, Nishio2019} involve resource management, Edge AI pipeline design, approximating compilers, and compression of models.

\begin{tcolorbox}[title=\textbf{Lessons Learned from Subsections A, B and C:}, breakable, skin=enhanced jigsaw]
\begin{enumerate}
    \item Latency: For functional-safety applications, latency is a key parameter. Applications such as obstacle detection, adaptive cruise control, emergency braking, localization requires strict latency. This property can be considered as one of reason for shift from vehicular cloud intelligence to edge computing and Edge AI applications.

    \item Heterogeneous Networks: Connectivity within the ecosystem can be separated from short-range to long-range communication. Within the dynamic operational environment, proposed communication and delivery schemes should be capable of adapting to the diverse distributed network (Base stations, V2X, Cellular (4G/5G/6G), road-side units, edge-servers, cloud etc.).
    
    \item Resource Management: Similar to a heterogeneous network, computing devices within the vehicle-edge ecosystem is also expected to be distributed. Devices may consist of distributed CPU, GPU, FPGA, TPU, and accelerators. Resource allocation and management schemes at the edge-server are required to process the sensed and transmitted data. Deployed resource allocation and management schemes can also counter other challenges such as excessive energy consumption from computing, data filtration, pseudo labeling, re-training approaches and update for the global AI model.
    
    \item Joint-Inference: Strict latency, network bandwidth constraint and high volume data in connected vehicular applications provide an opportunity to focus on approaches that allows computation distributions at the vehicular and edge-server level. Early-exit DNN, federated learning, data aggregation and model partition approaches are the potential solutions when combined with communication-efficient AI approaches and mechanisms.
\end{enumerate}
\end{tcolorbox}

\subsection{Federated Learning and Autonomous Driving}
Concept and applications of federated learning were initially proposed in \cite{konevcny2016federated, McMahan}, with the aim of training a large machine learning model in a distributed manner across several devices to accelerate the process. In recent years exploration and scope of federated learning have been further extended to reducing the communication costs \cite{chen2021communication, Jeong-non-iid-data}, enabling privacy preserving methods and  enhancing security of the model and data \cite{Dinh2021FL, Nishio2019, Bonawitz2019FL, ye2020federated}, and resource allocation/management schemes for the participating devices \cite{Xiao2021FL, Tian2020FL, Nishio2019}. For connected and autonomous driving applications federated learning have also been proposed with edge computing to jointly utilize the computation power of edge servers, and to take advantage training the model with dynamically distributed data over the edge devices, by further encouraging privacy preserving methods at the edge node or system level. Based on communication and computation approaches, the research topic covered below are further categorized as: ``Communication efficient algorithms" \cite{diao2021semifl, konevcny2016federated, chen2020joint, chen2021communication, Sattler2019FL, zhao2021fedpage}, ``Resource constrained devices" \cite{shen2021communication, Shiqiang2018FL, ma2021adaptive, Xiao2021FL, wang2019edge}, ``Heterogeneity aware" \cite{Nishio2019, Tian2020FL, Dinh2021FL, Tran2019FL, chen2020joint}, ``Energy efficient approaches" \cite{Umair2019FL, Amiri2021FL, Shi2020FL, Ren2021FL}.

\textbf{Resource constrained:} Edge device-server joint inference and optimization \cite{Xiao2021FL, wang2019edge, shen2021communication}, involving edge device computation capability and associated local model accuracy with minimum cost. The resource in this context is computation, power capability and communication overhead between edge device and server. Joint optimization is prioritized using vehicle parameters such as position and velocity to ensure a round of communication and parameter update with local edge server. The system \cite{Xiao2021FL} comprises of connected autonomous vehicles where edge device handles the initial computation requiring less resources and offloads the heavy computational tasks to the distributed edge servers in the urban driving scenario, with local model training, selected model aggregation \cite{ye2020federated}, computation complexity and weights transmission as primary matrices. For computation optimization a self-adaptive global best harmony search (SGHS) algorithm is used. For on-device resource allocation combination of SGHS and on-board computing and transmission power optimization algorithm is used to enhance the local model accuracy.

\textbf{Heterogeneity Aware:} In collaborative driving the data obtained from multiple sources such as infrastructure sensors, legacy data available in server or from other vehicle sensors is of heterogeneous form \cite{Jeong-non-iid-data}. This basis and requirement bring heterogeneity aware distributed learning as a primary criterion for fully connected autonomous driving. Federated learning by choosing edge devices is addressed by \cite{Nishio2019, Tran2019FL, chen2020joint} to counter the computational capability and communication bandwidth. In the approach edge server randomly chooses the client for model aggregation and requests for current communication and computation resource available for processing, based on the received information the edge server distributes the model parameters to the edge devices with high available resources for the model aggregation and which uses batch normalization approach for updating the global model. Another distributed approach is studied in \cite{Dinh2021FL} where the heterogeneous data is combined in subsets to minimize the aggregation loss from edge devices and improve the convergence, combination of these approach is also followed in \cite{Tian2020FL}, where low latency communication is ensured through quadratic convex functions.

\textbf{Communication efficient:} A  semi-supervised federated learning (SSFL) is proposed in \cite{diao2021semifl}, to alternatively train the statistical model at the edge server with unlabeled data using semi-supervised fixmatch \cite{jeong2020federated, zhang2020improving} and mixmatch learning method \cite{berthelot2019mixmatch}. For acceleration and better convergence of local model, static batch normalization technique is used which is adaptation of batch normalization \cite{jeong2020federated} and group normalization\cite{zhang2020improving}. In alternative training the local model at edge server is aggregated by retraining with the ground truth or legacy data to enhance the model accuracy at each round of training and in the next round of communication between the node and server the aggregated model weights are transmitted to update the global model and legacy data. Similar joint learning method is proposed in \cite{chen2020joint, chen2021communication}, where the local model is re-trained over edge devices and is transmitted over cellular network to the base stations for global model aggregation. To minimize the model learning loss and to collectively use the communication bandwidth, the base station categorically select the edge device using greedy approach by proposing a resource allocation and power allocation schemes at base station and edge device respectively. For the power allocation scheme at the edge devices two primary criteria: retraining of local model and power needed for model or weights transmission is considered. Other proposed method includes sparsification of data and gradient, quantization for minimizing communication bandwidth, which has been discussed below.

\subsubsection{Sparsification}  For collaborative or federated learning the commonly used approaches for sparsification is to compress the gradient and/or the data. Edge computing or processing near the edge is being adopted as a popular approach for an autonomous vehicle. Instead of transmitting the data or raw data, the model weights processed at the edge is transmitted to the devices participating in communication. Reducing the transmission time \cite{CE23} or using efficient delivery scheme, such as REMD is also proposed as communication-efficient approach in FL setting \cite{ismael2021reliable}. Another approach \cite{CE3, zhou2018addressing} proposed in FL use-case is to use of a lower-limit value in which the gradients with certain magnitude and greater than the predefined lower-limit are sent from the edge to the server and the left-over gradients are not used to weight or model aggregation. Using this approach the compression on the up-link and down-link communication can be implemented. However, the challenge is to choose the favorable lower-limit value, as similar to soft-filter pruning, the quantization and selection of the wrong lower-limit value can directly impact the overall model aggregation, which may provide an overall reduced model size but decreases the accuracy.

To overcome the previous challenge, stochastic gradient descent with k-sparsification is proposed in \cite{stich2018sparsified}, by reducing the data and model size and also improving convergence through error compensation for the transmission taking place between edge and server. A similar approach is used in \cite{aji2017sparse}, the method proposes to fix the sparsity rate. The communication or transmission of the gradient is only enabled for a fraction of the gradient with the highest magnitudes and keeping the unused gradient in the container. The sparsity rate used by the authors is p = 0.001, and this approach has relatively less impact the learned model overall accuracy and performance. To further overcome this performance gap, authors in \cite{lin2017deep} proposed modifications to the existing approach through deep gradient compression. Deep gradient compression uses approaches such as: momentum correction, local gradient clipping, for the convolutional neural network and recurrent neural network. Results show that gradients are compressed by ratio of 270-660 following a hierarchical approach, without slowing down the model convergence. Sparsification methods were initially proposed with the function of improving and promoting distributed and parallel training among the cloud and data-centers. However, these methods lacked model convergence and aggregation as a scope which is currently a most essential metrics for the federated and distributed machine learning. Similarly,  attention should be given to the number of edge devices participating in the transmission and the server participating in collaborative training. As the study in \cite{lin2017deep} shows the communication between the edge and server will not be compressed and reduced if the number of devices participating in training is less than the chosen sparsity value.  

\subsubsection{Quantization} Along with the usage for compression of deep neural network, the approach is also used in communication-efficient algorithms, with the goal of minimizing the communication bandwidth between the edge device and server. Quantization in communication applications with a federated learning setting, can approximate the weight updates on edge devices by limiting the update to a certain set of values. One such implemented approach on independent and identical distributed data is signSGD \cite{bernstein2018signsgd}. In the proposed method authors quantized each gradient update to the allocated binary sign and reduced the bit size, with a value of 32. It is important to note that signSGD also implements compression at the server by approximating the gradient received from edge devices and further contains investigation and theoretical analysis of algorithm in distributed machine-learning setting. In this approach the participating devices transmits the information of the associated gradient to the local server which transmits back the updated and aggregated gradient sign to the participating devices for the local model aggregation. The analysis shows that this approach achieve a similar variance score in comparison to other contemporary methods and has a better convergence rate to a stationary point of a general non-convex function. Similar approaches of scalar quantization through stochastic methods are proposed in PowerSGD\cite{vogels2019powersgd}, ATOMO \cite{wang2018atomo}, TernGrad\cite{wen2017terngrad}, QSGD \cite{alistarh2017qsgd, alistarh2018convergence}.

ATOMO \cite{wang2018atomo} and QSGD \cite{alistarh2017qsgd} propose to quantize the gradients with a better convergence rate allowing faster distributed training of neural networks, which is highly suitable for enabling collaborative learning within the vehicle-edge environment. However, the performance analysis in the vehicle-edge surrounding should consider trade-off such as accuracy-efficiency-reliability for safety-critical and real-time applications while accuracy-energy for the latency tolerable applications. While deploying such methods focus can be also given to compression ratio and convergence rate, as for communication and federated learning within autonomous vehicles it is necessary to consider compression in uplink and downlink transmission and communication. In \cite{alistarh2017qsgd, alistarh2018convergence} authors theoretically analyse the quantized stochastic gradient descent to balance the trade-off with federated learning parameter: convergence and communication cost. In this approach, the edge devices are allowed to adjust the number of bits transmitted in each iteration of communication according to the variance. As shown in \cite{alistarh2017qsgd} the device in a federated setting can transmit around 2.8n+32 bits in one communication round (here n is the number of parameters in model). This setting leads to 5x approximate bandwidth saving. Similarly, to speed up the training amongst participating devices an approach is presented by \cite{seide} to perform gradient quantization using one bit, which can make the distributed training to be 10x faster. For evaluation in \cite{seide}, authors used neural network with speech recognition which is highly anticipated use-case in autonomous driving \cite{Lopecs, PI-Edge}. 

Dedicated uplink compression has been explored in \cite{CE9} by using the quantization theory. In this work authors explores the transmission of trained model by identifying the available channel bandwidth through quantization scheme. The authors further propose an encoding-decoding approach consisting of partitioning, dithering, quantization and entropy coding at the encoding function and entropy decoding, dither subtraction, collecting and model recover at the decoding function. The evaluation of proposed quantization system is demonstrated through numerical study which shows error is mitigated through federated averaging and high federated learning performance gains. Contrary approach to scalar quantization methods, for the uplink and donwlink compression is vector quantization method \cite{CE7}. As compare to scalar methods, vector quantization offers dimension reduction along with the quantization scheme in federated learning setting. In the vector quantization method \cite{CE7}, numerical studies similar to \cite{CE9} were conducted. The method comprises of encoding strategy similar to \cite{CE9} and analysis using probabilistic quantization. However, a different decoding step of dither subtraction is applied to reduce the distortion and minimize the error. The approach also involves using of lossless source coding scheme in entropy coding and entropy decoding to generate non-uniform distribution of the quantized outputs.

\begin{tcolorbox}[title=\textbf{Lessons Learned:}, breakable, skin=enhanced jigsaw]

\begin{enumerate}
    \item FL using Edge: Collaborative or joint-learning applications, such as Edge computing and Edge AI, complements federated learning. The advantages of using these techniques in conjunction with each other allow a reduction in communication bandwidth to the cloud and also promote privacy by not sharing/transmitting sensitive data.
    \item Compression: It is extremely challenging to implement traditional federated learning techniques within conventional edge devices. Model compression approaches have been explored to accelerate the training/inference by reducing the computational complexity and requirement.
    \item Re-Training: AI models deployed for connected vehicle applications can often encounter unseen data. Property of FL to retrain the model and update the weights through convergence benefits use-cases, such as HD map update.  
    \item Communication Reduction: Current federated learning approaches focus on reducing the communication overhead through compression by overlooking the exploration of protocols that are lightweight in nature.
        \end{enumerate}
\end{tcolorbox}

\subsubsection{Overcoming Communication Overhead} An open challenge for autonomous vehicles in federated or distributed learning environment is overcoming the computational complexity and communication overhead. Federated averaging \cite{McMahan} proposes methods to reduce the communication frequency to overcome communication delay by not initiating communication between device and server after every iteration. Rather the federated averaging method computes the weight for every participating device using multiple iterations of stochastic gradient descent. Implementing the approach on convolutional neural network and recurrent neural network, the analysis shows that communication between participating devices can be delayed upto 100 iterations by still maintaining the convergence rate. A key requirement for this convergence rate is that the data should be independent and identically distributed between the participating devices. The communication round can be further increased with a higher delay, but as a trade-off it increases the computational cost on participating devices. As shown in above subsections, the work to overcome communication overhead combines the use of sparsification and gradient quantization \cite{Bonawitz2019FL, Tran2019FL, Tian2020FL}. These methods however do not have a better convergence rate.

A ternary quantization-based federated learning approach is proposed in \cite{shen2021communication} to overcome the communication overhead in uplink and downlink communication. The quantization method is implemented on the participating devices and the server thus implementing local training and global model update through weights. This approach also reduces the model complexity for the edge and server devices. For evaluation authors performed simulation considering the battery powered vehicle with connected autonomous driving capability to achieve fast inference and low communication overhead thus making inference possible on resource-constrained embedded and edge devices \cite{diao2021semifl, ma2021adaptive}.

\begin{tcolorbox}[title=\textbf{Challenges for vehicular services:}, breakable, skin=enhanced jigsaw]
Distributed learning has been a popular approach to tackle computation and communication challenges. Federated learning has provided alternative methods to re-train and deploy AI models with low communication and computation cost in dynamically distributed heterogeneous settings. Deployment of connected autonomous driving services (e.g. OTA update, traffic monitoring, and forecast) using federated learning approaches will enhance the privacy of data used for training and can also prevent attacks on the AI model. However, for real-time applications such as vehicle localization, and mapping, challenges exist in terms of computational resource requirement, latency, and communication bandwidth. A typical SLAM application in the vehicular application is deployed using large sensed data from a camera, LiDAR, and radar. The data size is approximately in gigabytes and should be processed by the AI model in less than 5ms, which also makes it challenging to transmit it to a nearest participating device for computation. 

Deployment of FL using Edge AI for vehicles can be considered as an optimization problem. The complexity further increases when energy-efficiency is considered as a direct parameter. A major challenge currently encountered for optimizing such efficient applications with FL context is the unavailability of the real-world large-scale dataset. As the problem has to be tackled by considering the communication and computing cost.
\end{tcolorbox}

\section{Enabling Frameworks for Autonomous Driving Services}
Due to the limited computation, storage, and communication resources of edge nodes, as well as the privacy, security, low-latency, and reliability requirements of AI applications, a variety of autonomous driving oriented edge AI system architectures have been proposed and investigated for efficient training and inference. This section gives a comprehensive survey of different Edge AI frameworks and related architecture. It starts with a general discussion on different architectures and categorically comparison.

\begin{figure*}[ht]
  \centering
  \includegraphics[scale = 0.088]{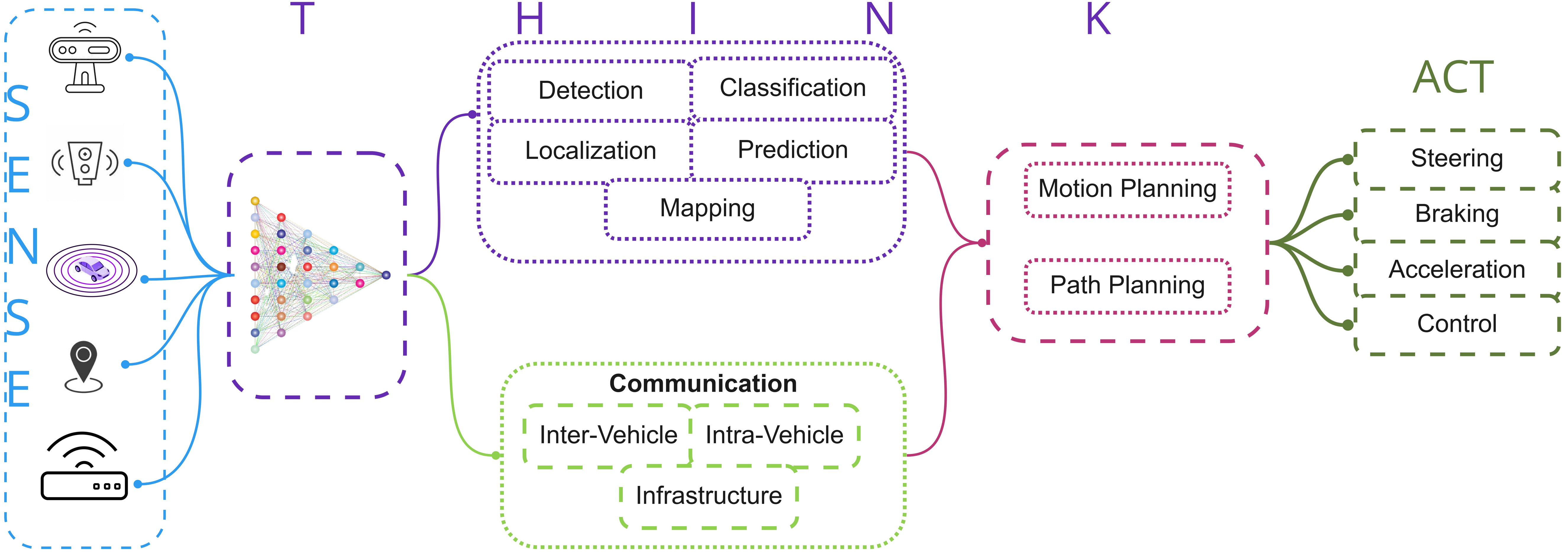}
  \caption{Sense-Think-Act model, which has been used as a backbone for autonomous driving frameworks \cite{Autoware, xu2021opencda}.}
  \label{fig-sta}
\end{figure*}

\subsection{Autonomous Driving Framework}
Since the development of deep neural network supporting perception and SLAM applications, researchers have focused on the design and development of simulators, software often referred to as a framework. Nvidia Drive \cite{e2e2016-nvidia}, Waymo \cite{ackerman2021full}, ApolloAuto \cite{apolloauto2021} are some commercially released driving frameworks supporting vehicular applications. Autoware \cite{Autoware} based on ROS is developed for an embedded platform that was released in 2018. OpenCDA \cite{xu2021opencda}, is one of the recently released and most complete open-sourced driving frameworks consisting of communication modules, real-time feedback and a simulation environment, thus providing a platform for cooperative driving applications. Following section details the architecture and components of these frameworks.

\subsubsection{Autoware}
Autoware \cite{Autoware} is ROS \cite{2009ros} based framework. It is developed on the concept of the sense-think-act model, also shown in Figure~\ref{fig-sta}. It is primarily designed for vehicles driving in urban areas. Autoware is dependent upon perception-based sensor suites such as cameras and LiDAR for enabling object detection, tracking, and localization using deep neural networks. The sensed information is fused from both sensors to also create 3D maps around the vehicle, which helps in precise localization by combining it with SLAM algorithms and sensors such as GNSS and IMU. The other major components are planning and control, which is based on probabilistic robotics utilizing deep neural networks. The software can be installed on the autonomous embedded platform containing Ubuntu operating system by using ros packages and dependencies to enable self-driving functionality in urban scenarios. Additional software module development and sensors integration such as radar is in progress which is required for the highway and related driving scenarios.

\subsubsection{Apollo Software Platform}
Apollo software platform has seen multiple revisions since its release, the currently available version integrates processing components: localization, perception, prediction, planning, control, and communication (V2X). At present, the platform incorporates deep learning models to perform major tasks through a dedicated computing unit comprising of CPU and GPU. One of the unique components of this platform is HD Map which can be also be tracked on the generic display monitor to perform and visualize accurate localization. The platform can be easily integrated with autonomous embedded platforms running UNIX operating systems. However, one of the important to calibrate with respect to the sensors and computing hardware installed on-board. The components in the apollo framework \cite{apolloauto2021}:

\textbf{Perception:} The perception module majorly focuses on obstacle detection, traffic lights and lanes. The perception module is mostly performing 3D object detection and is implemented using a deep neural network focusing on the region of interest on the high precision map. The output from the object detection module comprises 3D bounding boxes around the object based on the class, height, width and probability of the detected object. In the background a detection to track algorithm is used in order to identify the individual objects with respect to the timestamps, this timestamp is logged in the system and later serve as feedback to improve the accuracy for the similar detected objects. The perception module utilises the data fusion strategy using the Kalman filter.

\textbf{Localization:} In the platform, multisensory fusion localization is used which is based on GPS, IMU, LiDAR, radar, and HD maps. The localization module is based on the fusion approach of the Kalman filter comprising of two-step prediction update cycle. It comprises of two major blocks, the GNSS localization which provides the position and velocity information and the LiDAR localization which provides the position and heading information. Finally, the inertial navigation solution is used for the prediction step of the Kalman filter, while the GNSS and LiDAR localization is used to update the measurement step of the Kalman filter.

\textbf{HD Map:} The high definition map \cite{HD12} component in apollo comprises legacy data collected by sensors containing information related to road definitions, intersections, lanes, traffic signals. It is used to reduce the computational demand of the hardware by integrating the existing information of the street or lane the vehicle is currently driving on. In the apollo platform, it is also used as a safety feature providing centimetre level accuracy in localization of the vehicle. The steps involved in the development and publication of HD Maps include sensor data sourcing, processing, object detection and manual verification. In case of road or lane change, the existing platform utilises updates of HD maps in data centres through crowd sourcing which can involve data collected by other autonomous vehicles, smartphones and other sensors on the intelligent map production platform.

\textbf{Simulation:} Along with the on-device implementation, apollo platforms also provides the function to virtually create the driving scenarios by choosing the above-mentioned modules, dedicated deep neural networks and test driving scenarios, validate, and optimise the existing models. The simulation results of the driving scenario can be logged which can be further utilised as feedback for the development of algorithms and tackling the false-positive scenarios. 

\subsubsection{OpenCDA}
OpenCDA \cite{xu2021opencda} is one of the driving frameworks designed for cooperative driving with simulation and prototyping capability, it contains three major components which are: cooperative driving system, co-simulation tools and scenario manager. In the background the cooperative driving system is also based on the sense, think, act model and comprises of perception, communication, planning and control as the fundamental blocks to enable individual as well as cooperative driving. There is an application layer also present which is responsible for enabling cooperative perception, cooperative localization, platooning, and cooperative merge. For the second component i. e. simulation part, this framework utilises CARLA \cite{dosovitskiy2017carla} for autonomous driving simulation and SUMO \cite{krajzewicz2010traffic} for traffic simulations, and with combined integration of these two, the traffic scenes and simulation can be created for example vehicle platooning, traffic merge. The simulation tools exchanges information with the sensor and processed data, it continuously provides the HD map data to the system and receives control commands. The third component which is scenario manager exchanges information with simulation tools and cooperative driving system, to evaluate the cooperative driving states, and trigger special event and provide it to the simulation tools. The framework is developed in python and is also scalable for the 64-bit OS UNIX system.

\subsubsection{Openpilot}
This is another framework in the category of conditional or partial automation. The framework is developed by \url{http://comma. ai/} \cite{comma} and was released in 2017, and with revisions and additions of new features from 2017-2021, it is primarily dependent upon vision sensors and provides assistance to the driver with the driving services such as adaptive cruise control (ACC), forward collision warning (FCW), lane departure warning (LDW), and automated lane centring. The framework is dependent upon the services or components which can be divided as: Sensors and actuators, Neural network runners, Localization and calibration, Control, and System Logging \& miscellaneous services. The versions of the framework can be integrated into embedded devices supporting the android or UNIX operating system.

\subsubsection{Autopilot}
Autopilot \cite{krok2020tesla} provides assistance to the driver by sensing the environment around the vehicle through high definition automotive cameras and ultrasonic sensors. The software stack comprises of assistance and safety features such as automotive emergency braking, collision warning (front, rear and side), obstacle detection and also include smart navigation systems thus providing actuation and control. The framework on the backend uses a deep neural network performing object detection, semantic segmentation, and depth estimation to further provide the feedback and output for motion and path planning algorithm which suggests optimal route and actuate according to the destination set in the navigation. The software framework was initially designed to support the driver for highway driving scenarios and is also being tested for urban driving conditions.

\subsubsection{CARMA}
This framework \cite{bujanovic2021carma} falls in the category of cooperative driving by enabling connected vehicles. The software stack is programmed in C++ programming languages and is configured using the ROS environment for the Ubuntu operating system. The framework utilises the Autoware cite{Autoware} for enabling level 3 automation capability and additionally contains a communication module in the sensing layer which includes DSRC, V2X and cellular connectivity, thus initiating communication and exchange of information with other vehicles, infrastructure and the cloud. The cooperative feature of this platform consists of four levels of planning for the vehicle which includes route planning, maneuver planning, trajectory planning and command planning.

\subsubsection{AutoC2X}
AutoC2X \cite{AutoC2X} is a cooperative driving framework that is a combination of two software: Autoware cite{Autoware} and OpenC2X cite{OpenC2X} developed for cooperative driving applications. OpenC2X is cooperative intelligent transport system software that is open source and is helpful for prototyping solutions such as traffic management, and platooning. AutoC2X setup comprises of pair of devices which is a computing unit and router, installed with AutoC2X-AW and AutoC2X-OC software at the car and infrastructure respectively. The flow of information can be from car to infrastructure or from infrastructure to car. For the test experiment, the authors enabled cooperative driving services such as perception, coordinate transformation, localization, path planning through a proxy cooperative awareness V2X messages. The results from the experiment show that cooperative perception messages using AutoC2X were delivered within 100 ms.

\begin{tcolorbox}[title=\textbf{Lessons Learned:}, breakable, skin=enhanced jigsaw]
\begin{enumerate}
    \item Stack: The discussed autonomous driving framework incorporates popular deep-learning algorithms to perform perception, localization, mapping and path-planning tasks.
    \item Resource: These frameworks require an onboard high-performance computing device with extensive memory capacity to process large-volume data and deploy intelligent algorithms such as CNN, DNN, or RNN.
    \item Energy: The presence of extreme resources and computing devices results in high on-board energy consumption, which has been overlooked.
    \item Communication: Initially proposed driving frameworks lacked the presence and usage of a communication unit/module, which is highly important to enable collaborative driving and fully autonomous vehicle.
\end{enumerate}
\end{tcolorbox}

\subsection{Application oriented Frameworks}
In autonomous driving frameworks, the other proposed approaches are tasks oriented and are strongly influenced by distributed or collaborative learning approaches. Popular research directions for an energy-efficient edge in these categories are data partition, model partition, Offloading, and communication. In the data partition method\cite{s21}, the collaborative compressed sensing approaches are used, which allows the distribution of data amongst participating devices, thus leveraging repetitive computational load on an individual device. Model partition approaches\cite{RC9} utilize resource allocation schemes\cite{ECS6}, which are based on the availability of computing resources at the participating devices. A large DNN model is split into smaller forms for collaborative training and inference. Using the server as the central or primary mode of communication in edge-server joint inference applications computation offloading-based edge inference systems \cite{Off1, Off2, Off3} has been proposed. The approach involves offloading data or offloading a part of the inference load or the entire task to the edge server in the surrounding. In this context, communication and resource-aware techniques are also implemented, which decides on choosing a server amongst the available server based on latency.

\begin{tcolorbox}[title=\textbf{Lessons Learned:}, breakable, skin=enhanced jigsaw]
\begin{enumerate}
    \item The approach proposed in these application-oriented frameworks for connected vehicles considers either data reduction or model reduction, which can result in energy-saving mechanisms from either computation delay or communication perspective. However, for energy-efficient connected vehicles, both metrics need joint optimization and acceleration.
    \item  The communication approach proposed in these use-cases generally considers ideal conditions in communication. However, the communication in the vehicular ecosystem is often dynamic and heterogeneous, which consists of several low, and mid-range protocols with minor differences in distances. Therefore, another limitation of these frameworks is the inability to work in dynamic network conditions.    
    \item Similar shortcomings can be seen in computation as well. Edge in the vehicular ecosystem is constructed from heterogeneous devices with different computing abilities. AI models proposed in these application-oriented frameworks does not account for computing heterogeneity which may lead to miscellaneous cost.
\end{enumerate}
\end{tcolorbox}

\subsection{Energy-Efficient Edge Frameworks}

\subsubsection{OpenVDAP}
Open vehicular data analytic platform (OpenVDAP) \cite{OpenVDAP} is a data analysis framework developed for connected autonomous vehicles (CAV) with the design requirements of edge computing. The services included in OpenVDAP are real-time diagnostics, advanced driver assistance systems, infotainment, and other quality-of-experience services. The platform is developed to deal with low latency applications in autonomous driving by collaborating with the other edge nodes (other vehicles), base stations, local servers, and the cloud in the driving environment. With respect to the application, the platform consists of on-board heterogeneous computing, a communication unit, an edge-based vehicle operating system (EdgeOS$_v$), a driving data integrator, and edge computing aware libraries for vehicular data processing. The primary purpose of using these components is to intelligently allocate the on-board computing resource to the algorithms for the data processing, implement the data offloading strategies and also enable communication between the vehicle and infrastructure.

\subsubsection{CAVBench}
The benchmark suite \cite{CAVBench} was proposed to evaluate the performance of edge computing frameworks and software in connected autonomous driving services. Applications or services included in the CAVBench are object detection, tracking, SLAM, battery diagnostic, edge video analytics, and speech recognition, which are similar to the components included in OpenVDAP \cite{OpenVDAP}. The services and deep learning algorithm associated are evaluated based on latency (on-device processing), and power consumed as these can help in the development of an end-to-end autonomous driving application. For the evaluation purpose, the state-of-art algorithms such SSD \cite{SSD2016}, ORB-SLAM \cite{ORB-SLAM} were implemented and resulted in observations such that the priority is to be given real-time applications with the latency demands for instances the demand for localization and processing is greater than the tracking. Therefore, the system demands a processing layer or container to execute the driving data and tasks in a hierarchical manner. The observation also shows end-to-end deep learning applications can decrease the processing latency of computing units with heterogeneous structures. Therefore, distributed algorithms can perform better than the baseline for some of the autonomous driving services. 

\subsubsection{$\pi$-Edge}
To enable the computational intensive tasks simultaneously on resource-constrained embedded systems, $\pi$-Edge \cite{PI-Edge} is proposed which enables edge intelligence on the low powered embedded devices using the operating system $\pi$-OS. As the present embedded devices contain heterogeneous computing structure \cite{PE4, PE5}, the authors proposed a heterogeneity aware run-time and scheduling layer to execute the tasks by targeting the on-board energy efficiency. The framework also contains a component that enables the communication between edge-node and server and also performs the data offloading tasks to save the on-board power consumption. For offloading experiments, authors used applications and data from object detection and speech recognition, as their latency demand (requires approx 100 ms) is more compared to SLAM applications (should be performed within $4-5$ ms). The offloading algorithm is implemented through collaboration between edge-node(vehicle) and the server where it categorically searches for edge-node where data can be offloaded and estimate a time required for this application along with the needed computational resources. If the server is not capable of offloading the data the information is shared over the network with the purpose of executing the offloading task on the next available local server.  The results were demonstrated by integrating the framework on Nvidia Jetson devices which consume 11 W of power.

\subsubsection{MobileEdge}
As connected autonomous vehicles are processing and integrating multiple driving services at the same time, the vehicle computing unit can face significant load because of computational complexity. To address these issues several distributed computing approaches in the vehicular ecosystem has been proposed. MobileEdge \cite{MobileEdge} is one such edge computing framework that utilises the main vehicle computing units and the other resource-constrained edge-node or devices such as raspberry pi or Hikey970, present in the vehicular ecosystem. The architecture of the MobileEdge framework consists of two processes one which is executed on the vehicle computing unit and the second process which occurs on the random edge-node. The vehicle computing unit further consists of a management system and device resource monitor, the on-board task scheduler and the task execution process. while the edge-node consists of resource monitor, task receiver and task execution process. The communication between the vehicle computing unit and edge-node is initiated over the local wireless network. The resource monitor on both devices is responsible to track the system usage and being aware of the power consumed. The task scheduler manages the incoming raw data from the sensors and passes them for execution or to offload it to free resources. The task executor process the driving services associated such as video analytics or speech recognition. Task receiver module which is present on the edge-node receives offloaded data from the vehicle-computing unit and pass it to task execution module of edge-node, by implementing the distributed computing application.

\subsubsection{LoPECS}
LoPECS \cite{Lopecs} is another low power edge computing system for real-time autonomous driving. It has addressed the challenges of implementing computational intensive tasks on resource-constrained embedded devices and can be considered as an extension of $\pi$-Edge as it replaces the $\pi$-OS with the real-time OS which is lightweight as compared to traditional used ROS. The architecture of LoPECS contains four major layers: services classification, runtime layer, heterogeneous aware layer and edge-server coordinator. The services classification layer helps in the identification of tasks and features which needs real-time execution and associated power consumption. The second layer is runtime which contains the real-time OS, architecture-aware scheduler and API. The architecture-aware scheduler can be further categorized into the inter-core scheduler and inner-core scheduler. This scheduler helps in processing the incoming data and acts as a data pipeline to the systems GPU, CPU, video and audio accelerator. The last layer is the edge-server coordinator and it performs the data and algorithm management strategies by enabling communication in the vehicular environment. This layer is also responsible to implement data offloading strategies. For the evaluation purpose, the framework combining SLAM, object detection and speech recognition is implemented on Nvidia Jetson TX1 (15 W capacity) with consuming 3.5 W on GPU, and 4.2 W on CPU from these tasks and still allows resource and memory for implementing other driving tasks.

\subsubsection{AC4AV}
AC4AV \cite{AC4AV} framework is designed for connected autonomous vehicles and proposes the access control techniques for the autonomous vehicle. The framework also utilises a data processing and abstraction method in which the sensed data from the sensors is identified and applied for access related applications. The primary purpose is to protect the sensed data from phishing attacks or being maligned from the vehicle environment. The architecture of AC4AV comprises of three-layer to prevent the raw sensor data from unauthorized access which are: access control engine, action control, and lastly a logger database. The access control engine provides dynamic authentication to access the data and also incorporates a data processing layer that identifies the type of data and its relevant use in the autonomous driving services, as the vehicle is sensing from several sensors and the same data can be used for multiple algorithms. The action control service layer is responsible for two tasks which are action capturing and responding. The last layer is the logger database which captures and records the actions. The information from the logger database can be used as an audit for future actions as it can help in improving latency for targeted applications. The implementation is based on publishing and subscribing, a classic approach for message and communication within an embedded environment. A similar framework autonomous vehicular edge \cite{AVE}, is based on ant colony optimization, which includes offloading and task scheduling strategies with a decentralized approach. In this paper, the task scheduling strategies use a generalization assignment problem and is categorized according to the driving priority and latency demand. The computational complexity using a greedy algorithm and ant colony optimization were analysed in which the computational power is measured along with the latency and ant colony optimization results in latency less than 1 ms.

\begin{tcolorbox}[title=\textbf{Key Takeaways and Lessons Learned:}, breakable, skin=enhanced jigsaw]
\begin{enumerate}
    
    \item  OS: Traditional autonomous driving frameworks used ROS or similar open-source systems integrated with Unix to deploy CAV. In contrast to the on-board vehicular frameworks, the discussed edge frameworks are integrated using a custom lightweight OS to reduce computational delay for computing-intensive applications on resource-constrained devices.
    \item Scheduler: As the vehicular services are hierarchy-oriented and require execution within a short timeframe. These edge frameworks focused on integrating a scheduling algorithm also sometimes referred to as the runtime layer to optimize the data processing for vehicular services. 
    \item Communication: The discussed edge frameworks mostly used the combination of OBU and RSU to exchange vehicle data and model weights. A few frameworks also used local wireless networks (802.11b) installed customarily at the road intersection to initiate communication. However, the frameworks lacked testing the communication  heterogeneity using the combination of edges such as base stations, RSU, cellular stations, and embedded devices integrated with wireless modules. Strict latency transmission of information to The communication approach proposed in these use-cases generally considers ideal conditions in communication. However, the communication in the vehicular ecosystem is very dynamic and heterogeneous, which consists of several low, and mid-range protocols with minor differences in distances. Therefore, another limitation of these frameworks is the inability to work in dynamic network conditions.    
    \item Data: A shortcoming in the edge frameworks is the inability to handle high-volume data from the vehicle sensors in case of collaborative inference between multiple vehicles. These frameworks do not propose any modules to offload or aggregate the sensed data at the edge. This may result in flood of data at the edge and repetitive computation for the redundant data.
\end{enumerate}
\end{tcolorbox}

\section{Research Outlook and Open Problems}

This survey studies a comprehensive and categorized review of approximation techniques and energy-efficient methods for autonomous driving services. The perspective and basis on selection of topics is based on previously and recently proposed AI and Edge Computing approaches for the driving services considering model size and real-time deployment for the low powered embedded devices, and the relevant conclusive factor of these approaches is based on the heavy computation complexity which results into high energy consumption on embedded devices. The main question asked in this survey is, What are the current approaches and trends which can promote the concept of Level 5 self-driving by enabling the Artificial Intelligence at the Edge Devices with an energy-efficient approach. During the process, some of the secondary questions related to development of model, Optimization and Inference approaches such as Federated learning were explored. However there are some research gaps and open problems which needs to be considered such as: Data management and process techniques on the Edge devices, Categorization for autonomous driving use-cases for real-time use-cases, autonomous driving tasks hierarchical categorization and energy implications of them. These topics are covered in the following subsections.

\subsection{Connected Vehicle Service and Case-Study}

\subsubsection{HD-Map}
Vehicle drivers has been regularly using 2-D map (for example: Google Maps, Apple Maps) with the cellular technologies to have a precise and short duration travel within or between the cities. For Self-driving vehicle this is been replaced by High Definition maps or 3D maps which are a result of mapping the roads and infrastructure using high definition cameras and LiDAR sensors to localize the vehicle precisely in the 3D environment and by saving the information over the data centers or cloud services. The average roads or dynamic scenes in a developed country changes only 5\% - 13\% \cite{HD12} over the year, due to construction or any other dynamic events. Therefore an approach can be implemented along with SLAM technique to update the previous captured HD Map in the cloud based on change in the scenarios. Lately, research approaches \cite{HD15} has been proposed to have a DNN model to update the HD map data available in the cloud from the crowd-sourced data.

\subsubsection{Vehicular Networks and Communication}
For Edge-Assisted autonomous driving learning a cooperative approach needs to be implemented and practiced for collaborative decision making. Federated Learning has been proposed as potential solution for this problem, however open directions remains on the topics including common framework and deployment for heterogeneous vehicular networks, resource allocation using Federated Learning,  communication, computing, and caching strategies for FL, data privacy and model security, collaborative intelligence.

\subsection{Enablers for Edge Application in Autonomous Driving}

\subsubsection{Data Management for Edge-Assisted Services}
The current autonomous driving practices involves individual implementation of tasks such as Classification, Detection or Localization. One of the reason associated with individual processing is non-availability of data management techniques and practices for the edge devices. If data management techniques can be proposed a heterogeneity aware layer can be integrated to serve as a data flow between the Sensor and DNN algorithm. Having Data Management techniques for the Edge-devices can simultaneously enhance the collaborative driving functionality and also improve the offloading strategy thus enabling each vehicle to make independent decisions and also share the output for cooperative driving use-case. Real-time compression of streaming data (from IoT/camera) and to be stored on the Edge for tracking or monitoring.

\subsubsection{Collaborative Edge Intelligence}
The limited data bandwidth over wireless communication may lead to failure with decision making process in an autonomous cars as in case of cooperative driving the autonomous vehicle should continuously transmit data between the vehicle and the cloud. Implementing AI at the Edge on large scale can enable autonomous cars to efficiently process data and also enabling communication between vehicles, to overcome the network and communication related issues, distributed edge computing and federated learning approaches can be implemented which can enable the data processing and computation close or near to the vehicle as compare to the approaches in cloud computing where the processing and computation takes place in the centralized cloud. With the computation occurring close to the vehicle challenges and critical requirement such as accuracy, low-latency, reliability,  power, and energy consumption, of autonomous vehicles \cite{lin2018architectural} can be achieved. However, bringing services near the vehicles’ network where connectivity of the cars and their data is increasing at a tremendous rate often becomes highly crucial due to scalability issues in terms of functionality, administration, and load. Moreover, the connectivity among a large number of devices results in a flood of data production that can hinder the edge node to perform analytic on such a large-scale data by meeting strict latency requirements of autonomous cars. An adequate consideration must be given to resolve the edge-related issues for enabling successful deployment of autonomous cars.

\subsubsection{Training and Inference at the Edge}
As covered in this survey, the volume of data from the sensors and the quality of data is rapidly changing and increasing depending upon the change in dynamic layer. To ensure the adaptability of Edge AI algorithm for a new or different data from the autonomous driving services environment, it becomes necessary to perform and implement AI model training and inference at the edge. As this will ensure the real-time update of legacy or ground-truth data available near Edge and will also ensure the timely update of global model by exchanging binary weights with the backend cloud. The training and inference approach at the edge device can counter two major challenges: Inference latency which can be caused when the model is trained over other device or system (for example cloud) and Secondly the privacy as on device training will prevent the data from being shared over cloud.

\subsubsection{Common Edge Framework}
The implementation of approach such as Federated Learning, in autonomous driving demands a common Edge AI framework to be implemented across entities involved. A common edge framework across Vehicles, Edge Server, Infrastructure Sensors and Centralized cloud needs to be deployed to increase the efficiency and accuracy of applications. A common edge framework can bring the performance of individual devices to optimum level with need-basis collaboration from the vehicles and infrastructure sensor, Also it is important for privacy and security features.

\begin{figure*}[!ht]
\centering
  \includegraphics[width=\linewidth]{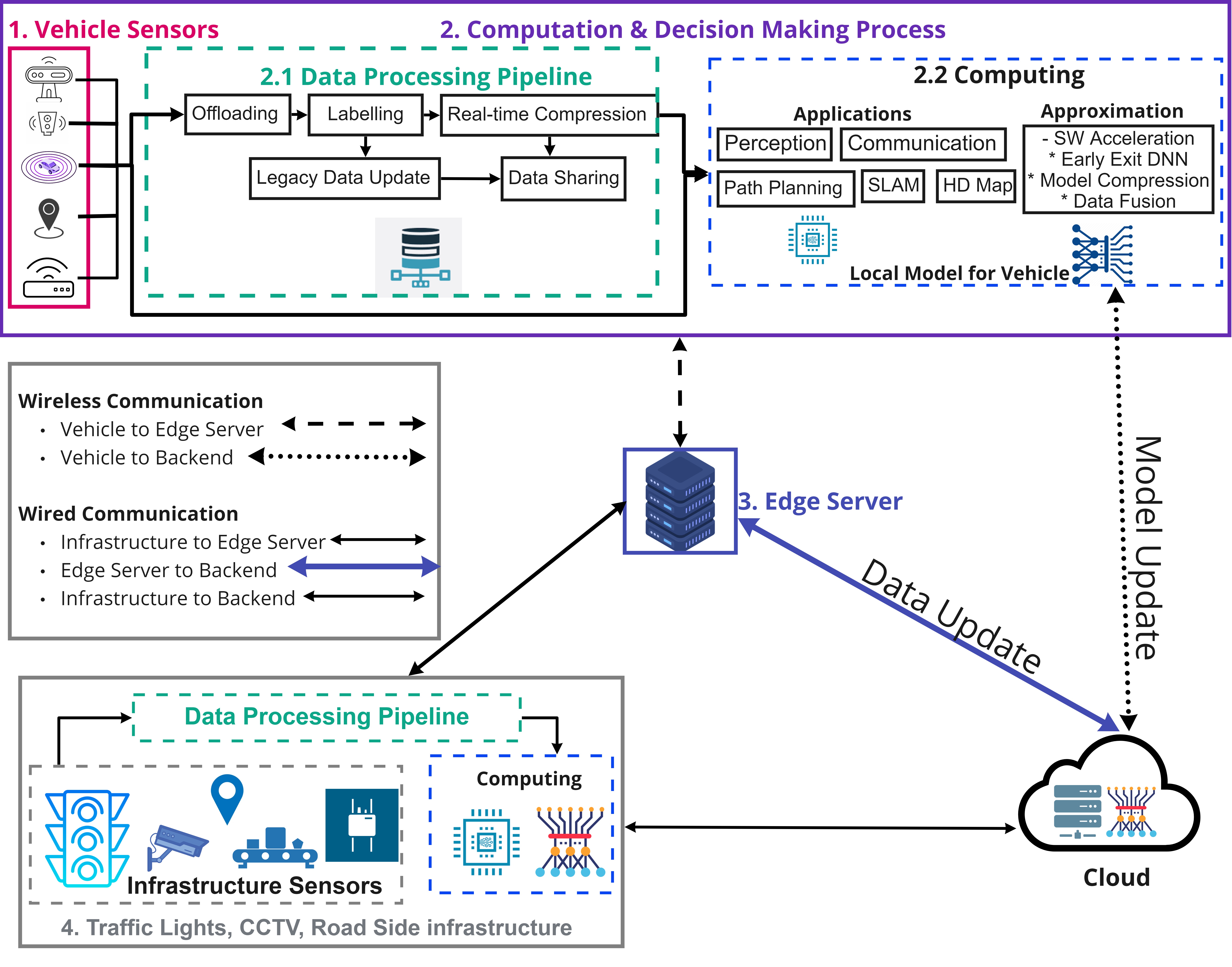}
  \caption{Edge assisted autonomous driving. The pipeline consists of on-board vehicle sensors in the car, the computation and decision-making process, Edge-server, infrastructure sensors {\&} devices, and the remote cloud.}
  \label{Edge-Ai-Pipeline}
\end{figure*}

\subsection{Energy Efficiency Evaluation of DNN Implementation on embedded devices}

\textbf{Resource Constrained Devices:}
Deep neural networks have delivered competitive accuracy for detection, segmentation, mapping and localization-related tasks for autonomous driving and with the advancement, in libraries and frameworks, they have also been deployed on resource-constrained devices such as smartphones, FPGA. However, there are several drawbacks which cannot be overlooked. The best-in-class accuracy from the state-of-the-art DNN is delivered at the extreme computational cost caused during training and inference \cite{CE12} which significantly increases the overall energy consumption in the autonomous driving ecosystem. Literature covered in this survey shows several methods that have been proposed to improve the accuracy and speed of DNN processing by optimizing metrics involved, for example optimizing the binary weights and operations involved in complex layer such as convolutional, Fire modules. These approaches do not necessarily make a significant improvement on the embedded device deployment and applications. Therefore there is an open requirement to propose an efficient DNN model for autonomous driving training and inference applications which simultaneously tackle the problem of low latency applications by overcoming the challenge of data and the energy consumed. 

Real-time applications such as SLAM or vision related tasks requires low latency and high precision by the embedded devices.  The relevant literature covered in this survey mostly exploits high-end GPU which is cost-intensive for large scale deployment. To enable these tasks on edge embedded devices a combined software and hardware acceleration approaches can be proposed which integrates data offloading strategies and energy or power saving techniques by simultaneously enhancing the accuracy and performance of these resource-constrained devices.

\subsection{Outlook of Edge AI Pipeline}
Takeaways and lessons learned from this survey highlight the need for an Edge AI processing pipeline that can process large volumes of data to carry out decision making processes. Figure~\ref{Edge-Ai-Pipeline} shows an overview of the Edge AI processing pipeline envisioned for future connected autonomous driving services, where the design of this pipeline corresponds to the joint processing of data at the vehicle on-board computing unit and at the Edge-server. In the proposed scenario, the AI processing pipeline consists of four major components. The first component comprises of the sensing unit present in the vehicle (camera, LiDAR, radar, GPS, and the communication unit (on-board unit + cellular connectivity), which is capturing data from the vehicles surrounding. 

The second component consists of computation and decision-making process, it involves an edge device placed in the vehicle processing the data through a deep neural network thus enabling driving services such as perception, SLAM and communications. The computation and decision-making process is a complex task while incorporating energy-efficient autonomous driving service through edge intelligence. Therefore, it is necessary to highlight the process which consumes a significant amount of on-board energy. Further, the computation and decision-making process is divided into data processing pipeline and computing respectively. The data processing pipeline is assigned with tasks, such as offloading, labelling, real-time compression, legacy data update and sharing the refined data with other entities involved in the surrounding, such as other vehicles, or edge servers. The processes carried out in the data processing pipeline can solve the primary concern of memory and power for resource-constrained edge embedded devices. The computing part involves processing the refined data over a deep neural network to generate the weights for driving applications. With the possibility of optimizing deep neural networks further acceleration and approximation techniques such as deep neural network model compression, data fusion or approaches such as early exit deep neural networks can be used. It is important to note that tasks such as SLAM, object-tracking, obstacle detection has low-latency and high bandwidth requirements, which makes it necessary and practical to process sensed data at the vehicle's on-board computing unit for these tasks instead of processing at the edge or remote cloud. Therefore, one of the inputs from the vehicle sensors bypasses the data processing pipeline and is directly used for computational purpose. 

The third component of the proposed edge AI processing pipeline consists of an edge server that is responsible for the processing of large-volume data and enabling communication in the vehicular ecosystem. The communication here can be categorized as: vehicle to edge server (for sharing of raw data), Edge server to a vehicle (for sharing of DNN model weights and refined or processed data), Edge server to infrastructure, and lastly edge server to backend cloud. To reduce the extensive on-board energy consumption in an autonomous vehicle, it is important to process the computationally intensive tasks over the edge-server, which implements lossless compression, optimization, and software approximation approach, which can help in achieving overall end-to-end energy efficiency.

The fourth component consists of roadside infrastructure which includes a sensor suite (CCTV, traffic lights, LiDAR, communication unit, GPS) similar to the vehicle and helps in tasks and applications such as smart traffic flow, traffic monitoring, map update etc. As illustrated in Figure~\ref{Edge-Ai-Pipeline} the component also comprises of similar data processing pipeline executing tasks such as offloading, labeling, real-time data compression and data or model sharing over wired communication with the edge server and backend cloud. The backend cloud is communicating with the vehicle, server and infrastructure sensors in case of DNN model update, or legacy data update. To improve the accuracy and enable collaborative driving, the model weights and data update should be shared between the backend cloud, vehicle and edge server over wireless and wired networks respectively.

\section{Conclusion}
This paper has explored and reviewed autonomous driving applications of perception, SLAM, HD map, vehicular communications, and inference approaches deployed on autonomous embedded platforms and edge devices. Attention has been given to exploring the currently available datasets and autonomous driving frameworks. Focusing on the impact of computational complexity and energy-efficiency on resource-constrained devices, we highlight the communication efficient approaches and software approximation techniques, including low-rank approximation, pruning, quantization and sparsification, which aim at reducing the statistical model parameters for inference. In addition, we also covered the energy-efficient deployment of AI applications on resource-constrained devices using allocation schemes, heterogeneity-aware mechanisms and federated learning. Our purpose is to provide a dedicated review of energy-efficient approaches for connected autonomous driving, ranging from vehicular communication, edge computing, approximation techniques to novel software-hardware frameworks. Besides identifying research gaps, we highlight the existing challenges and open problems that deserve further research investigations from the community. Finally, based on the identified gaps, we envision an Edge AI processing pipeline to share our outlook on potential development of energy-efficient applications for level 4 and beyond edge-assisted autonomous driving applications.


%



\section*{Acknowledgment}
The authors gratefully acknowledge funding from European Union's Horizon $2020$ Research and Innovation programme under the Marie Sk\l{}odowska Curie grant agreement No. $956090$ \href{http://www.apropos-itn.eu/}{(APROPOS: Approximate Computing for Power and Energy Optimisation)} and grant agreement No. $101021808$.

\ifCLASSOPTIONcaptionsoff
  \newpage
\fi

\bibliographystyle{plain}
\bibliography{bare_jrnl.bib}

\end{document}